%% file: main.tex
\documentclass[accepted]{uai2023} % after acceptance, for a revised
\usepackage[american]{babel}
\usepackage{natbib} % has a nice set of citation styles and commands
\usepackage{mathtools} % amsmath with fixes and additions
\usepackage{booktabs} % commands to create good-looking tables
\usepackage{tikz} % nice language for creating drawings and diagrams
\usepackage{amsmath}
\usepackage{amssymb}
\usepackage{mathtools}
\usepackage{amsthm}
\usepackage{color}
\usepackage{soul}
\usepackage{dsfont}
\usepackage{nccmath}
\usepackage{caption}
\usepackage{stfloats}
\usepackage{anyfontsize}
\input{math_commands.tex}

\title{Adaptive Conditional Quantile Neural Processes}

\author[1]{\href{mailto:<peiman.mohseni@tamu.edu>?Subject=Adaptive Conditional Quantile Neural Processes}{Peiman Mohseni}{}}
\author[2]{Nick Duffield}
\author[3]{Bani Mallick}
\author[4]{Arman Hasanzadeh}

\affil[1]{%
    Computer Science and Engineering Department\\
    Texas A\&M University
}
\affil[2]{%
    Electrical and Computer Engineering Department\\
    Texas A\&M University
}
\affil[3]{%
    Statistics Department\\
    Texas A\&M University
}

\affil[4]{%
    Google Cloud
}
  
\begin{document}
\maketitle

\begin{abstract}\label{abstract}
Neural processes are a family of probabilistic models that inherit the flexibility of neural networks to parameterize stochastic processes. Despite providing well-calibrated predictions, especially in regression problems, and quick adaptation to new tasks, the Gaussian assumption that is commonly used to represent the predictive likelihood fails to capture more complicated distributions such as multimodal ones. To overcome this limitation, we propose Conditional Quantile Neural Processes (CQNPs), a new member of the neural processes family, which exploits the attractive properties of quantile regression in modeling the distributions irrespective of their form. By introducing an extension of quantile regression where the model learns to focus on estimating \emph{informative} quantiles, we show that the sampling efficiency and prediction accuracy can be further enhanced. Our experiments with real and synthetic datasets demonstrate substantial improvements in predictive performance compared to the baselines, and better modeling of heterogeneous distributions' characteristics such as multimodality.
\end{abstract}

\section{Introduction}\label{sec: intro}
Conventionally, regression problems are approached by modeling the relation between inputs and outputs with a deterministic function where the parameters of this function are optimized with respect to a loss function. Non-parametric statistical methods, however, choose a different perspective by viewing the regression function itself as a random object. This allows for fitting a family of functions that are coherent with the data instead of a single one. (Conditional) Neural Processes (C/NPs) \citep{garnelo2018conditional, garnelo2018neural} are a class of such models that inherit the computational efficiency of neural networks and integrate it with desirable properties of Gaussian Processes (GPs), specifically uncertainty quantification and rapid adaptation to new observations \citep{garnelo2018conditional, garnelo2018neural}. In fact, \citet{rudner2018connection} showed that under certain conditions, NPs recover GPs with deep kernels. 

NPs can be viewed as the composition of an encoder and a decoder where the encoder embeds a finite collection of observations $\mathcal{E}_{\mathrm{context}}=\{(x_i, y_i)\}_{i=1}^{N}$, also known as context set, into a latent space. Subsequently, the decoder takes in a new target location $x^*$ together with the latent representation of context data to parameterize the conditional distribution $p(\ry^*\,|\,x^*, \mathcal{E}_{\mathrm{context}})$ of the corresponding target output $y^*$. Several variants of NPs have been introduced. \citet{kim2019attentive, nguyen2022transformer, kim2022neural, guo2023versatile, feng2023latent} incorporate attention mechanisms to make NPs less prone to under-fitting. \citet{gordon2019convolutional, foong2020meta} build translation equivariance into NPs by introducing convolutional deep sets. \citet{holderrieth2021equivariant} further extend the translation equivariance to more complicated transformations such as rotations and reflections. \citet{volpp2020bayesian} improve context aggregation by casting it as a Bayesian inference problem. \citet{lee2020bootstrapping} use the bootstrap technique for inducing the functional uncertainty. \citet{bruinsma2021gaussian, markou2022practical} propose Gaussian Neural Processes to model predictive correlations between different target locations. \citet{wang2020doubly} propose Doubly Stochastic Variational Neural Processes to capture target-specific local variation by adding local latent variables in a hierarchical encoder.

Despite their higher expressive power compared to vanilla C/NP, they fail to model multimodality that may be exhibited by the predictive distribution in real-world problems such as those in transportation science, economics, astronomy, and meteorology \citep{chen2016nonparametric}. 
This is due to the fact that existing models have a common Gaussian likelihood assumption.
To circumvent this issue, we propose to employ an infinite mixture of asymmetric Laplace distributions as the likelihood for C/NPs. This likelihood, which can be seen as an extension to the well-known quantile regression \citep{koenker1978regression, koenker2005quantile}, has shown to be an effective approach for modeling heterogeneous distributions \citep{dabney2018implicit, brando2019modelling}.
To further improve the expressive power of the model, we propose an adaptive extension of our approach where instead of fixing the quantile levels, the model learns to predict the quantiles that contribute more to the predictive likelihood which we will refer to as \emph{informative} quantiles.
We integrate our model with CNPs resulting in (Adaptive) Conditional Quantile Neural Processes (A/CQNPs) and conduct several experiments on both synthetic and real-world datasets to illustrate the performance enhancements achieved by our method. 
While, in this work, our focus is on CNPs, we emphasize that the simple, yet generic nature of the proposed approach allows for quick adaptation to other members of the NPs family.

\section{Preliminaries}\label{sec: prelim}
\subsection{Quantile regression}\label{subsec: quantile-regression}
In regression analysis, given a set of observations $\train = \{(x_i, y_i)\}_{i=1}^{N}$ from a pair of random variables $(\rvx, \ry)$, the objective is to learn a function $f$ that maps the inputs $x_i$ to the outputs $y_i$. 
We further assume that $\rvx$ and $\ry$ are vector and scalar random variables, respectively. In the case where $\rvy$ is a vector, we consider its scalar elements independently.
In decision theory, the optimality of an estimator is measured through its risk function $\mathbb{E}_{\train}[\mathcal{L}(y_i,\, f(x_i)]$ where $\mathcal{L}(\cdot\,,\,\cdot)$ is a loss function and an estimator with lower risk is preferred.
A common choice for the loss function is the mean squared error. It is well-known that for this loss, the estimator with minimum risk is $f(x) = \mathbb{E}[\ry \,|\, \rvx=x]$ which relates the inputs and outputs through the conditional mean \citep{casella2021statistical}.
However, in more complicated instances where $p(\ry \,|\, \rvx=x)$ is asymmetric, or multimodal, the conditional mean is not a sufficient statistic to summarize the distribution characteristics. This can be alleviated by using more robust statistics such as \emph{quantiles}. For $\tau \in (0, 1)$, the $\tau$-th conditional quantile $\mu_{\tau}(x) = \inf\, \left\{ \mu \,\,|\,\, p(\ry\leq \mu \,|\, \rvx=x)\geq \tau \right\}$ is obtained by minimizing the asymmetric absolute loss $\mathcal{L}_{\tau}(y_i, x_i) = \rho_{\tau}(y_i - \mu_{\tau}(x_i))$ where $\rho_{\tau}(y) = \max\, \left\{y\tau,\, y(\tau - 1)\right\}$. 
It is straightforward to show that minimizing $\gL_{\tau}$ is equivalent to maximizing the log-likelihood of an Asymmetric Laplace ($\gA L$) distribution \citep{yu2001bayesian} with a constant scale parameter. The density function of $\gA L$ distribution is defined as the following:
\begin{multline}\label{eq: al-distribution}
    \gA L\left(y\,|\, q_{\tau},\sigma_{\tau} ,\tau \right) = \\ \frac{\tau (1-\tau)}{\sigma_{\tau}}\times\exp{\left(-\frac{1}{\sigma_{\tau}}\rho_{\tau}(y - q_{\tau})\right)},
\end{multline}
where $q_{\tau} \in \mathbb{R}$, $\sigma_{\tau} \in \mathbb{R}_{>0}$, and $\tau \in (0, 1)$ are the location, scale, and skew parameters \citep{yu2005three}. Note that in vanilla quantile regression $\tau$ and $\sigma_{\tau}$ are fixed.

Like mean, a single quantile may not be sufficient to model heterogeneous distributions. To address this, several works have proposed to predict a set of quantiles instead of a single one \citep{liu2009stepwise, liu2011simultaneous, sangnier2016joint, dabney2018implicit, brando2019modelling, brando2022deep}. Among them, \citet{brando2019modelling} use an uncountable mixture of $\mathcal{A}L$ distributions to approximate the predictive distribution. More specifically, the predictive distribution is parameterized as follows:
\begin{multline}\label{eq: al-mixture}
    p(\ry \,|\, x) = \\ \mathbb{E}_{\tau \,\sim\, \mathcal{U}(0,1)}\left[\alpha_{\tau}(x) \, \mathcal{A}L\left(\ry \,|\, \mu_{\tau}(x), \sigma_{\tau}(x), \tau\right)\right],
\end{multline}
where $\mathcal{U}(0, 1)$ is the uniform distribution and $\alpha_{\tau}(x) \geq 0$ is the mixture coefficient such that $\mathbb{E}_{\tau \,\sim\, \mathcal{U}(0, 1)}[\alpha_{\tau}(x)] = 1$.
The parameters of this mixture distribution are estimated using deep neural networks.

\subsection{Conditional neural processes}\label{subsec: CNP}
Let $\mathcal{D} = \{(x_i, y_i) \in \sX \times \sY \}_{i=1}^N$ be a set of training observations corresponding to a realization of the following stochastic process; let $p(\rf)$ be a probability distribution over functions $f: \sX \rightarrow \sY$, then for $f \sim p(\rf)$, set $y_i = f(x_i)$ \citep{garnelo2018conditional}.
CNP is a conditional stochastic process $q_\upsilon$ (where $\upsilon$ is the set of parameters of $q$) that approximates the distribution over functions $p(\rf)$. In vanilla CNP, first, each pair $(x_i, y_i)$ of training observations (or a subset of them, i.e. \emph{context} set) is embedded into a latent space. Next, these embeddings are aggregated to form a representation, which is used with target inputs to predict the target outputs. We note that the aggregated representation is invariant to the ordering of context points. More specifically, given an encoder $\varphi_{\mathrm{enc}} \colon \sX \times \sY \rightarrow \sR^{d_e}$ and a decoder $\varphi_{\mathrm{dec}} \colon \sX \times \sR^{d_e} \rightarrow \Theta$, where $\sR^{d_e}$ is the embedding space and $\Theta$ is the set of parameters of the predictive distribution, CNP formulates the predictive distribution of $f(x^*)$ for a given target $x^*$ as:
\begin{equation}\label{eq: CNP-general-likelihood}
    p_\theta(f(x^*)\,|\, x^*, \mathcal{D}) = p_\theta\left(f(x^*) \,|\, \varphi_{\mathrm{dec}}\big(x^*, \varphi_{\mathrm{enc}} \left(x^*, \mathcal{D}\right)\right)\big).
\end{equation}
As mentioned earlier, different members of the CNPs family, have different encoder and decoder architectures.
In the vast majority of CNP variants, the predictive distribution is chosen to be a simple Gaussian distribution resulting in:
\begin{equation}\label{eq: CNP-Gaussian-likelihood}
    \begin{split}
        p_\theta(f(x^*) \,|\, x^*, \mathcal{D})
        &= \mathcal{N}(f(x^*)\,|\,\varphi(x^*, \mathcal{D})),
    \end{split}
\end{equation}
where $\varphi(x^*, \mathcal{D}) = \{\varphi_{\mathrm{dec}}^{\mu}(x^*, \varphi_{\mathrm{enc}} (x^*, \mathcal{D})),\, \varphi_{\mathrm{dec}}^{\sigma}(x^*,$ $ \varphi_{\mathrm{enc}} (x^*, \mathcal{D}))\}$ is the set of functions mapping the embeddings and $x^*$ to the mean and standard deviation
of $y^*$. To the best of our knowledge, none of the existing CNPs are capable of modeling heterogeneous distributions such as multi-modal ones.

\section{Method}\label{sec: method}

\subsection{Adaptive quantile regression} \label{subsec: adaptive-quantile-regression}
Although the uncountable mixture of $\gA L$s in \eqref{eq: al-mixture} gives us a comprehensive picture of the conditional distribution by fitting the full quantile function, it might lead to some practical inefficiencies that will be discussed below. For the rest of this discussion, we use $\mathcal{B}(c, \epsilon)$ to denote an open interval of length $2\epsilon>0$ and centered at $c$, i.e. $\mathcal{B}(c, \epsilon) = (c-\epsilon,\, c+\epsilon)$. 

Let's consider a case where there are two equally probable outcomes $y_1$ and $y_2$ ($y_1 \neq y_2$) for an input variable $x$. It is easy to verify that fitting the mixture distribution in \eqref{eq: al-mixture} will force $\gA L$ components with $\tau \in \mathcal{B}(0.5, \epsilon)$ ($0<\epsilon < |y_1 - y_2|/4 $) to settle around the median which is $\mu_{\tau}(x) = (y_1 + y_2)/2$. However, in this scenario, the median is not of high interest as we like our model to concentrate the probability density around $y_1$ and $y_2$. This example indicates that depending on the problem at hand, not all quantiles are equally important. 
To compensate for this, the mixture weights of $\gA L$ components corresponding to non-informative quantiles are expected to shrink to zero.
Theoretically, where we have infinite samples of $\tau$, this is not an issue, but in practice, the expectation in \eqref{eq: al-mixture} is approximated by a finite number of Monte Carlo samples. In other words, we can approximate the integral up to a certain precision which depends on the number of samples of $\tau$. Hence, it would be more efficient to avoid drawing samples of $\tau$ that correspond to non-informative quantiles. In our example, we like to sample $\tau$ such that $\mu_{\tau} \in \mathcal{B}(y_1, \epsilon) \cup \mathcal{B}(y_2, \epsilon)$. This will yield a more accurate approximation of $p(\ry\,|\,x)$ around $y_1$ and $y_2$ and prevent wasting computing resources.

% The second 
Another possible issue is regarding making point estimations. In most applications, the final stage involves reporting a set of values predicted by the model for a given input. In the case of using a symmetric unimodal distribution like Gaussian for representing $p(\ry\,|\,x)$, we usually report the distribution mean as it coincides with the mode. 
However, finding the modes of the uncountable mixture in \eqref{eq: al-mixture} is not always straightforward. A naive way to address this is by considering a set of uniformly sampled $\tau$ values, calculating their corresponding quantities $\{ \alpha_{\tau},\, \mu_{\tau},\, \sigma_{\tau}\}$, and finally selecting the most probable quantiles by comparing their mixture weights (or likelihoods). Similar to the previous case, it is quite likely that a small subset of quantiles is of interest. In case of having knowledge of these quantiles, we will be able to select better candidates as our final predictions with less effort which comes as a result of reducing the search space. 

Motivated by the above discussion, we propose using an adaptive set of quantiles $\mathcal{T}_x$ for each $x$ where the model learns to approximate the quantiles that are more significant in modeling $p(\ry\,|\,x)$. A simple approach for finding such a set incorporates replacing the non-informative uniform distribution $\mathcal{U}(0, 1)$ in \eqref{eq: al-mixture} with $q(\tau \,|\,x)$ (such that $\tau \in (0, 1)$) whose density is mostly concentrated around values corresponding to informative quantiles of $p(\ry\,|\,x)$. Note that the dependence of $q(\tau \,|\,x)$ on $x$ can be arbitrarily complex. Therefore, $\mathcal{T}_x$ can be viewed as samples from $q(\tau \,|\,x)$ and the problem of finding $\mathcal{T}_x$ changes to estimation of $q(\tau \,|\,x)$. Notice that conditioning on $x$ is crucial as the conditional distribution $p(\ry\,|\,x)$, and, hence, its quantiles (presumably) change at different inputs. Rewriting \eqref{eq: al-mixture} yields:
\begin{multline}\label{eq: al-mixture-with-posterior-of-tau}
    p(\ry \,|\, x) = \\ \mathbb{E}_{\tau \,\sim\, q(\tau \,|\,x)}\left[\alpha_{\tau}(x) \, \mathcal{A}L\left(\ry \,|\, \mu_{\tau}(x), \sigma_{\tau}(x), \tau\right)\right].
\end{multline}
The Monte Carlo Approximation of this expectation only requires samples from $q(\tau \,|\,x)$. Hence, having an analytic probability density function for $q(\tau \,|\,x)$ is not necessary as far as samples can be drawn. Various density estimation techniques can be deployed to find $q(\tau \,|\,x)$. Considering the complicated nature of $q(\tau \,|\,x)$, we propose to approximate it with a reparameterizable implicit distribution \citep{Diggle84implicit, mohamed2016learning}. This means that to sample from $q(\tau \,|\,x)$, we can first draw an auxiliary variable $u \sim \mathcal{U}(0, 1)$ and then set $\tau$ as a deterministic function $\psi \colon \sX \times (0,1) \rightarrow (0,1)$ of $x$ and $u$:
\begin{equation}\label{eq: implicit-distribution-def}
    u \sim \mathcal{U}(0, 1),\, \tau = \psi(x, u) \quad \equiv \quad \tau \sim q(\tau \,|\,x)
\end{equation}
When $\psi(x, u)$ is invertible w.r.t. $u$ and for a fixed $x$, $q(\tau \,|\,x)$ can be calculated by a simple application of the change of variable formula: 
\begin{equation*}\label{eq: change-of-variable-formula}
    q(\tau \,|\,x) = \mathds{1}_{(0, 1)}(g_x^{-1}(\tau))\, \frac{d}{d\tau}(g_x^{-1}(\tau)),
\end{equation*}
where $\tau=g_x(u)=\psi(x, u)$. However, this is generally not the case, and hence $q(\tau \,|\,x)$ is implicit. Using equation \ref{eq: implicit-distribution-def}, we can approximate the conditional distribution in equation \ref{eq: al-mixture-with-posterior-of-tau} as follows:
\begin{multline}\label{eq: al-mixture-with-adaptive-quantiles}
    p(\ry \,|\, x) \approx \mathbb{E}_{u \,\sim\, \mathcal{U}(0, 1)}\big[\alpha_{\psi(x, u)}(x)\\ \qquad\quad\mathcal{A}L(\ry\,|\,\mu_{\psi(x, u)}(x), \sigma_{\psi(x, u)}(x), \psi(x, u))\big],
\end{multline}
where $\psi$ is a fully-connected neural network. The high expressive power of neural networks allows $q(\tau \,|\,x)$ to be highly flexible and capture the dependencies between the elements of $x$ and $u$.

\subsection{Conditional Quantile Neural Processes} \label{subsec: CQNP}
Despite the attractive properties of likelihood-based models, their expressive power is highly impacted by the form of conditional distribution. Inherently, CNPs with Gaussian likelihood struggle to model more complicated distributions. We remedy this by adapting the predictive distribution in \eqref{eq: CNP-general-likelihood} to the compound distribution in \eqref{eq: al-mixture-with-adaptive-quantiles}. This requires augmenting the domain of $\varphi_{dec}$ and $\psi$ as demonstrated below:
\begin{equation*}
    \begin{split}
        \psi \colon \sX \times (0,1) \rightarrow (0,1) \quad &\text{to} \quad \psi \colon \sX \times \sR^{d_e} \times (0,1) \rightarrow \mathcal{T}_{\sX} \\
        \varphi_{\text{dec}} \colon \sX \times \sR^{d_e} \rightarrow \Theta \quad &\text{to} \quad \varphi_{\text{dec}} \colon \sX \times \sR^{d_e} \times \mathcal{T}_{\sX} \rightarrow \Theta
    \end{split}
\end{equation*}
Putting all pieces together, the predictive distribution of $f(x^*)$ for a given target input location $x^*$ would be:
\begin{multline}\label{eq: ACQNP-liklihood}
    p(f(x^*)\,|\,x^*, \mathcal{D}) =
    \mathbb{E}_{u \sim \mathcal{U}(0, 1)}[\alpha_{\tau}(x^*, \mathcal{D})\\ \times \mathcal{A}L (f(x^*)\,|\,\mu_{\tau}(x^*, \mathcal{D}), \sigma_{\tau}(x^*, \mathcal{D}), \tau)]
\end{multline}
where
\begin{align*}
    \tau=\psi(x^*, \varphi_{enc}(x^*, \mathcal{D}), u) \,\,\,\, & \\
    \{\alpha_{\tau} (x^*, \mathcal{D}) ,\, \mu_{\tau} (x^*, \mathcal{D}) ,\, \sigma_{\tau} &(x^*, \mathcal{D})\} = \\ &\varphi_{dec}(x^*, \varphi_{enc}(x^*, \mathcal{D}), \tau)
\end{align*}
We refer to this model as Adaptive Conditional Quantile Neural Process (ACQNP). In the case where $\tau = u$, the resulting model is Conditional Quantile Neural Process (CQNP). The expectation in \eqref{eq: ACQNP-liklihood} is approximated by drawing $N_{\tau}$ Monte Carlo samples from $\mathcal{U}(0, 1)$. Unlike \cite{brando2019modelling}, we avoid posing a uniform distribution over the mixing weights $\alpha_{\tau}$. Instead, we normalize them using a SoftMax function to have a valid convex combination of $\gA L$ distributions. The final likelihood can be expressed as follows:
\begin{multline}\label{eq: ACQNP-likelihood-approximation}
    p(f(x^*)\,|\,x^*, \mathcal{D}) \approx 
    \sum_{k=1}^{N_{\tau}} \big[\frac{e^{\alpha_{\tau_k}(x^*, \mathcal{D})}}{\sum_{k=1}^{N_{\tau}} e^{\alpha_{\tau_k}(x^*, \mathcal{D})}}\, \\ \times \mathcal{A}L (f(x^*)\,|\,\mu_{\tau_k}(x^*, \mathcal{D}), \sigma_{\tau_k}(x^*, \mathcal{D}), \tau_k)\big]
\end{multline}
Since CNPs and A/CQNPs use the same architectural design for computing the context representation, the computational complexity imposed by the encoder, i.e. $\mathcal{O}(\varphi_{enc})$, remains unchanged. However, the computational complexity of the decoder which comes from estimating $N_\tau$ quantiles at each input location is raised from $\mathcal{O}(\varphi_{dec})$ to $\mathcal{O}(N_{\tau}\varphi_{dec})$ resulting in the overall complexity of $\mathcal{O}(\varphi_{enc} + N_{\tau}\varphi_{dec})$. This extra computation can be done in parallel as different quantiles are estimated independently and mixed in the final stage.

\begin{figure*}[!t]
	\centering
        \scalebox{0.9}{
            \begingroup
            \setlength{\tabcolsep}{-0.5pt}
            \begin{tabular}{c@{\hskip -0.2pt}ccccc}
                \raisebox{4.8\normalbaselineskip}[0pt][0pt]{\rotatebox[origin=c]{90}{Double Sine}} &
                \includegraphics[width=0.22\textwidth]{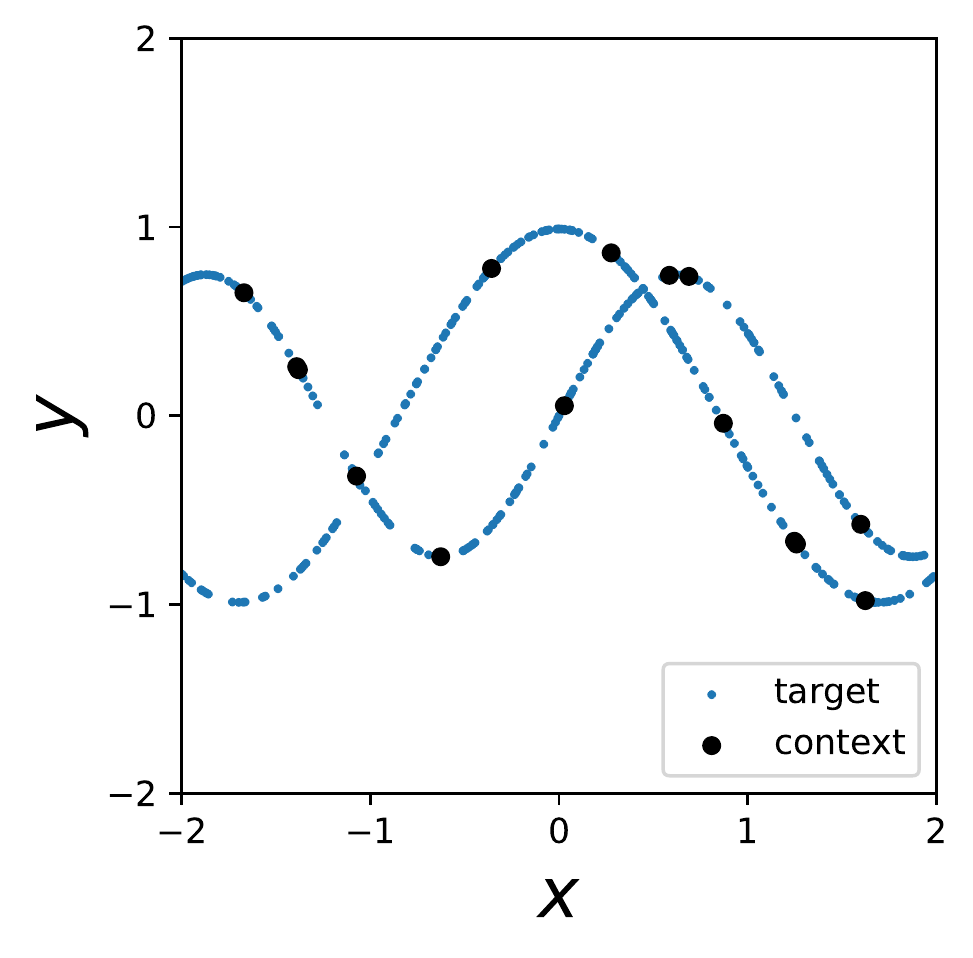} & 
	          \includegraphics[width=0.22\textwidth]{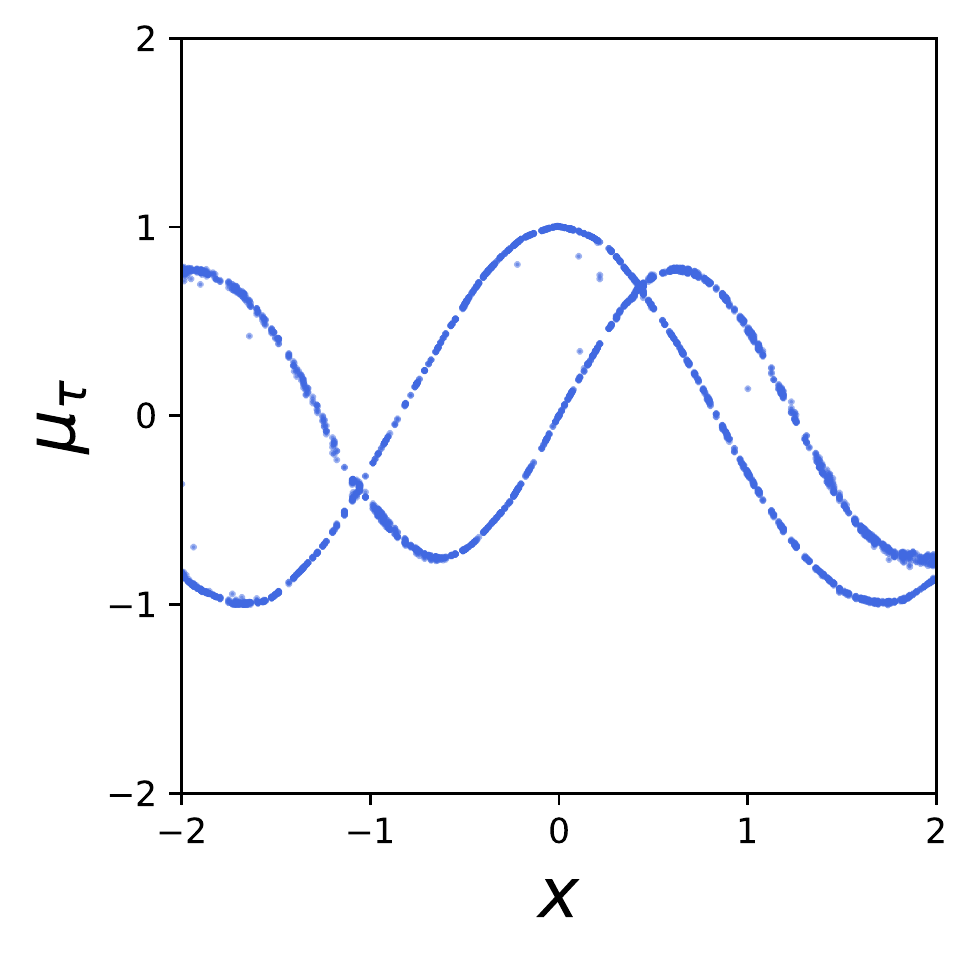} &
                \includegraphics[width=0.22\textwidth]{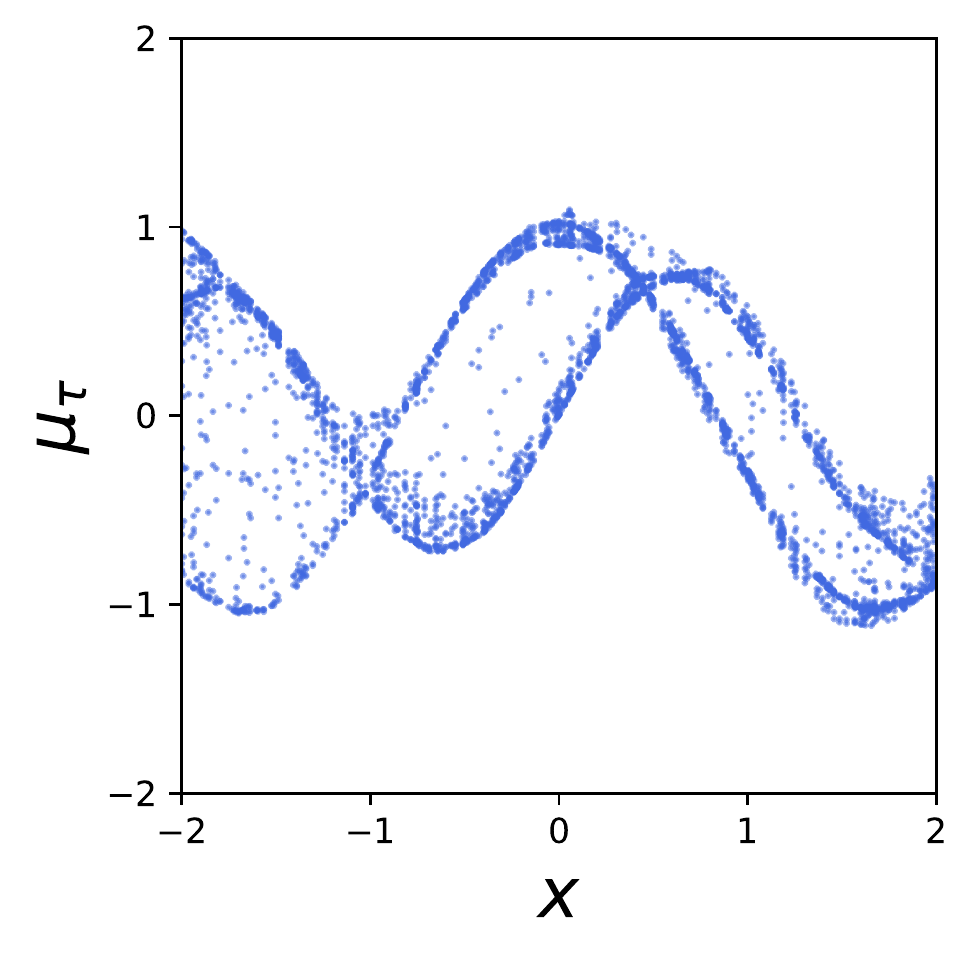} &
                \includegraphics[width=0.22\textwidth]{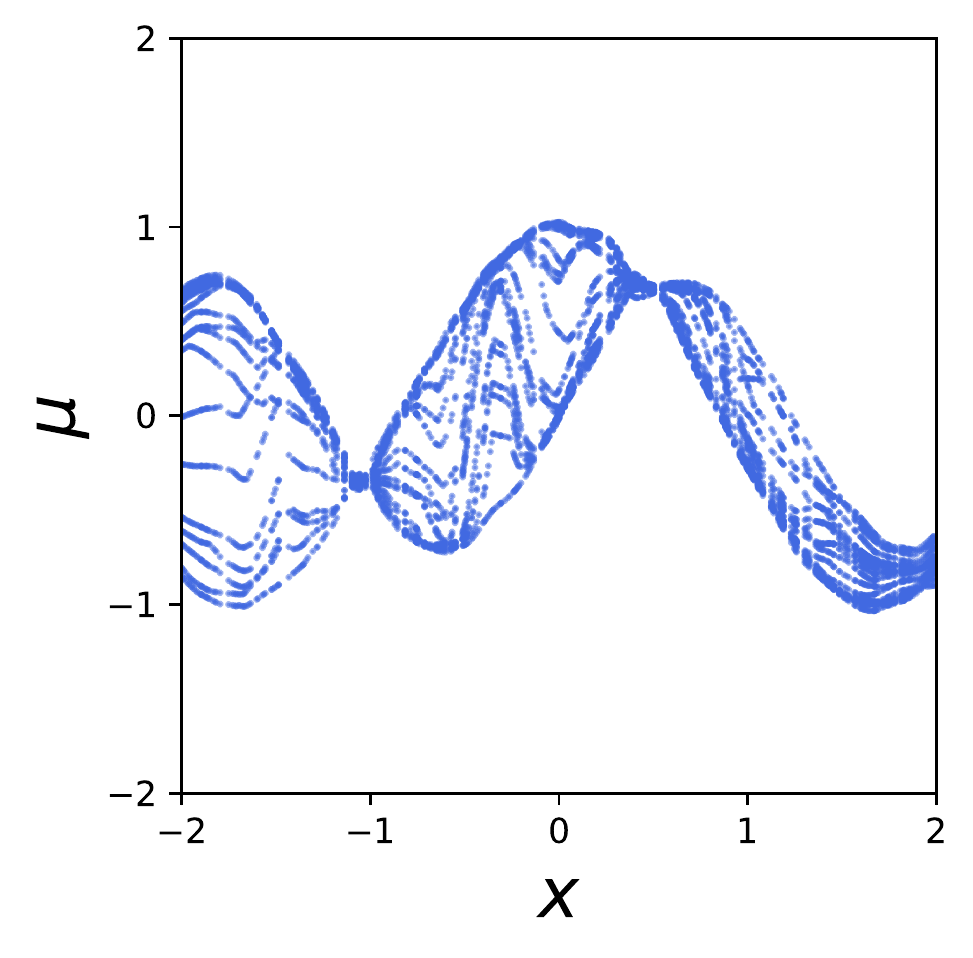} &
                \includegraphics[width=0.22\textwidth]{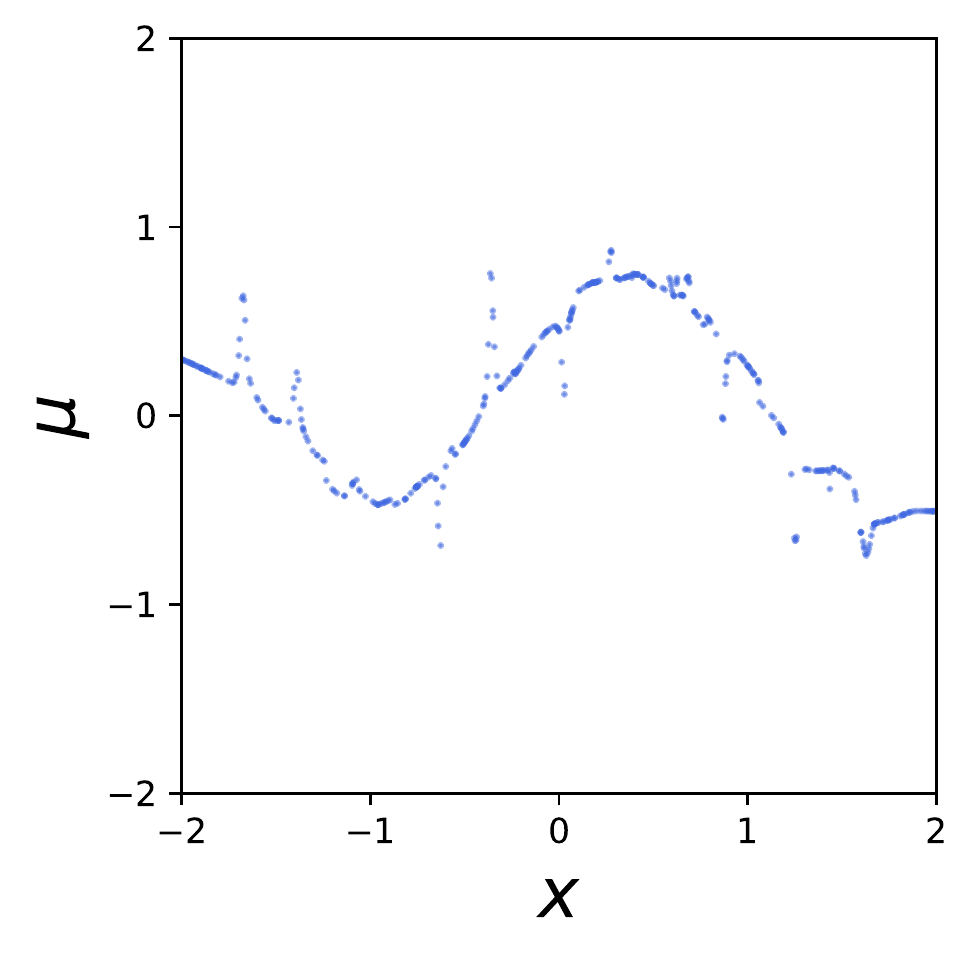} 
                \\
                \raisebox{4.6\normalbaselineskip}[0pt][0pt]{\rotatebox[origin=c]{90}{Circle}} &
                \includegraphics[width=0.22\textwidth]{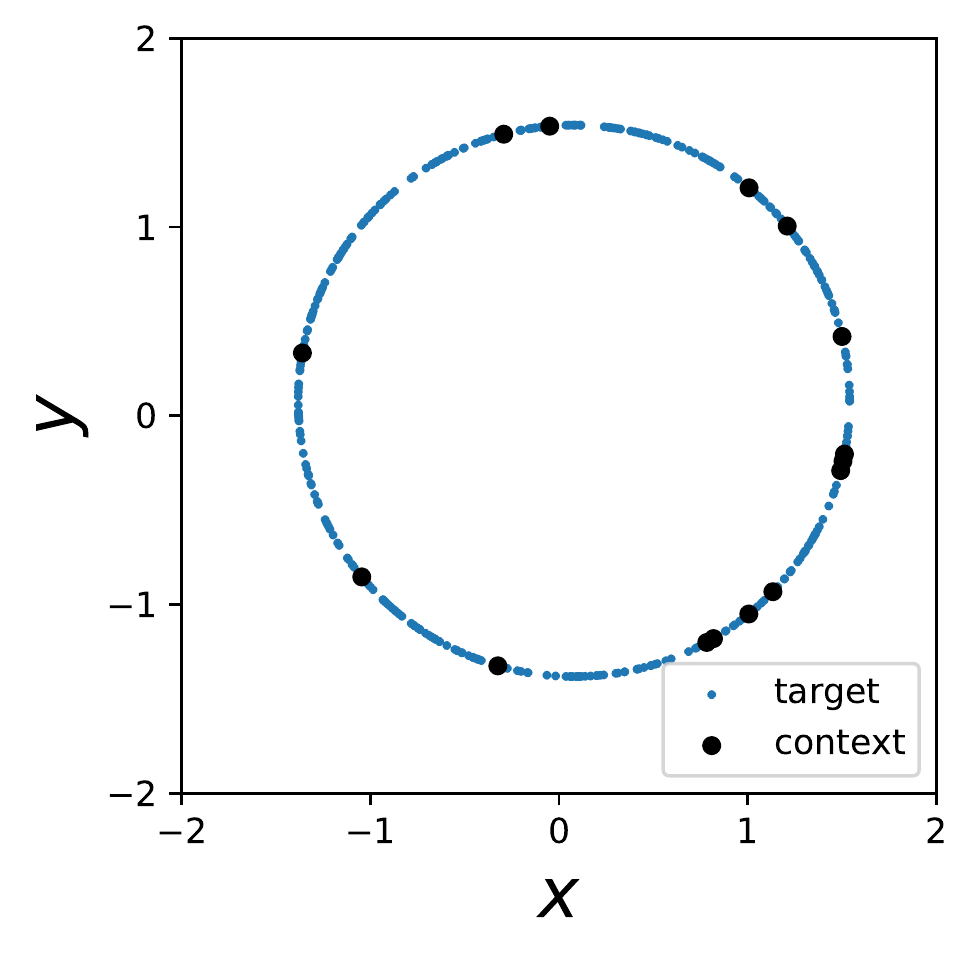} & 
	          \includegraphics[width=0.22\textwidth]{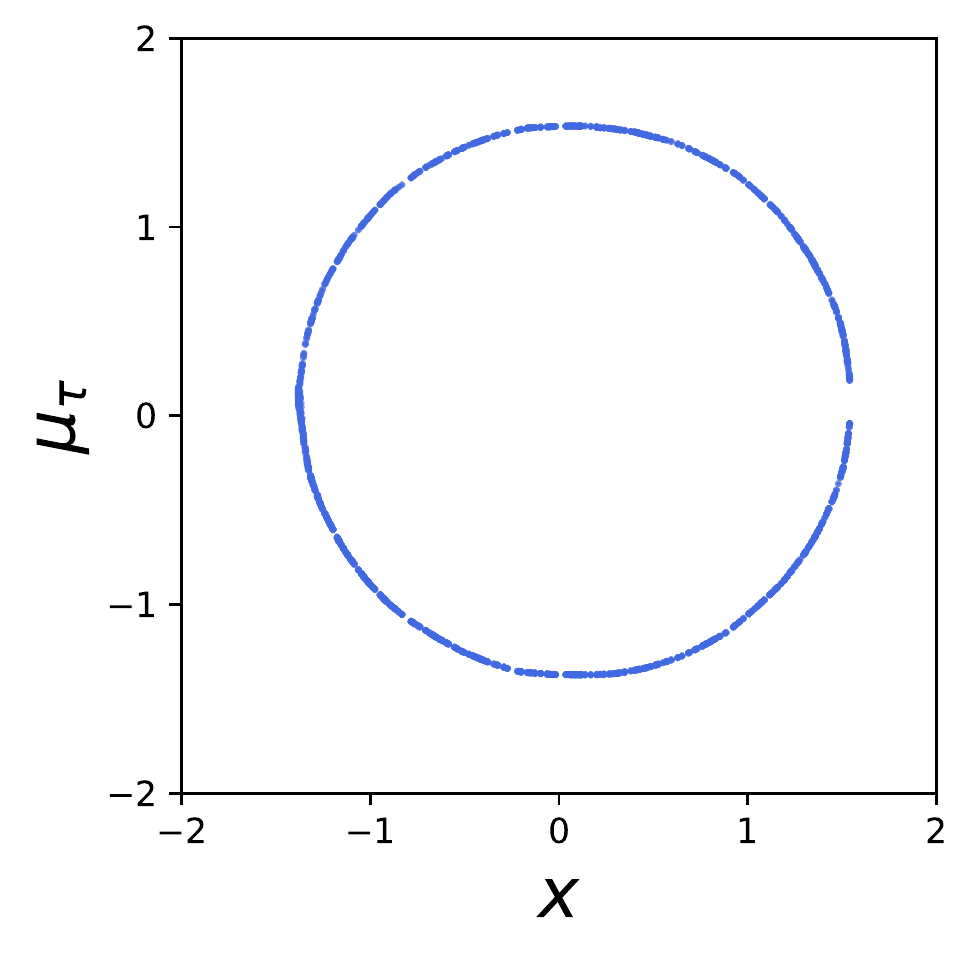} &
                \includegraphics[width=0.22\textwidth]{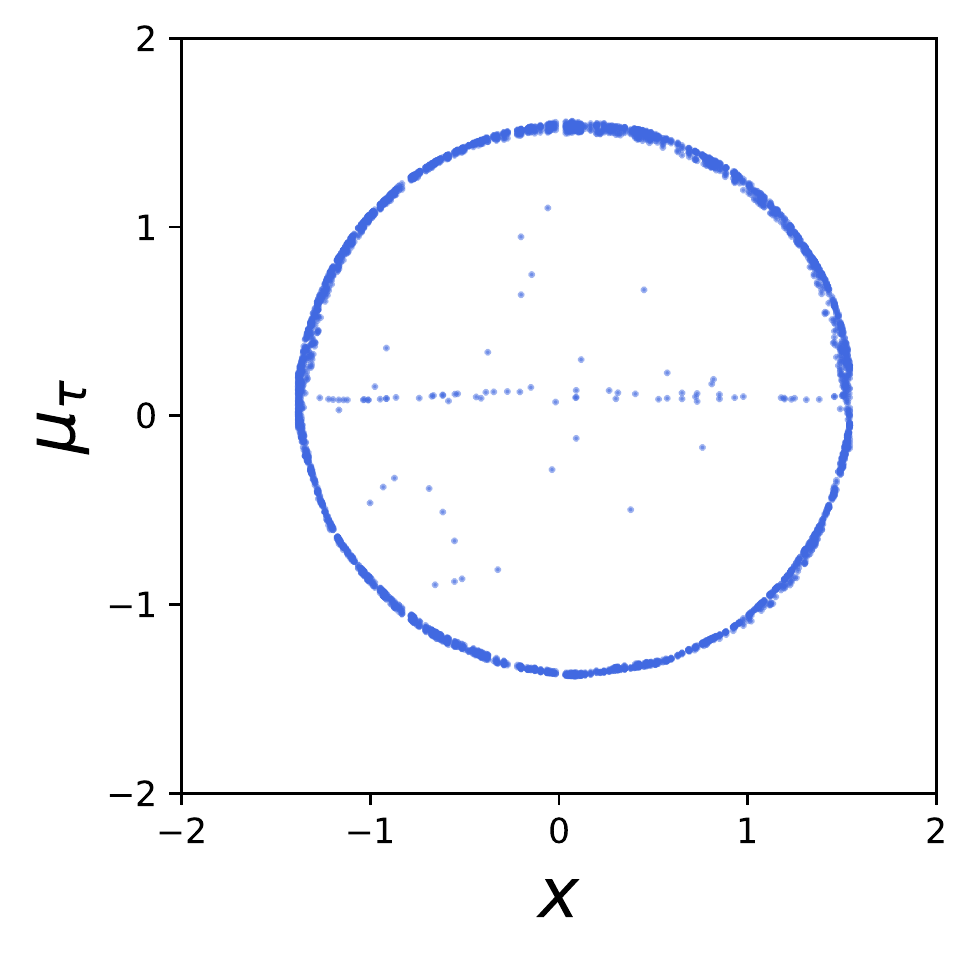} &
                \includegraphics[width=0.22\textwidth]{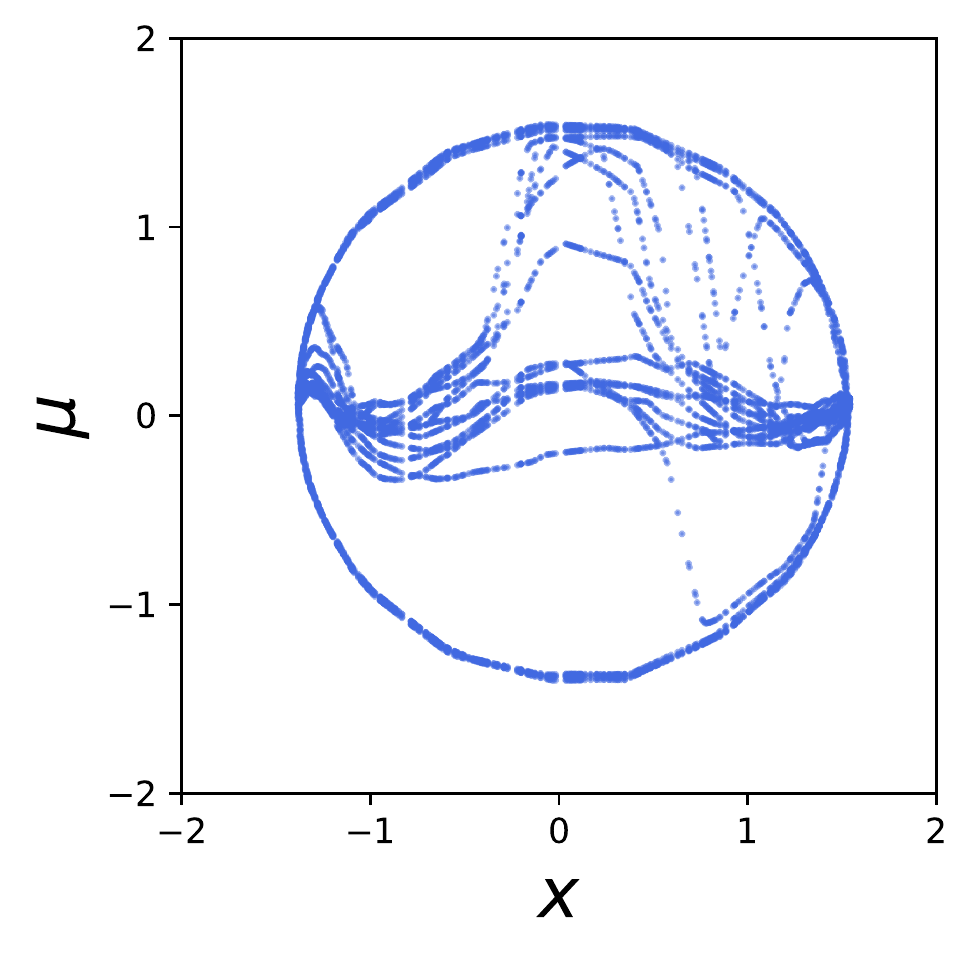} &
                \includegraphics[width=0.22\textwidth]{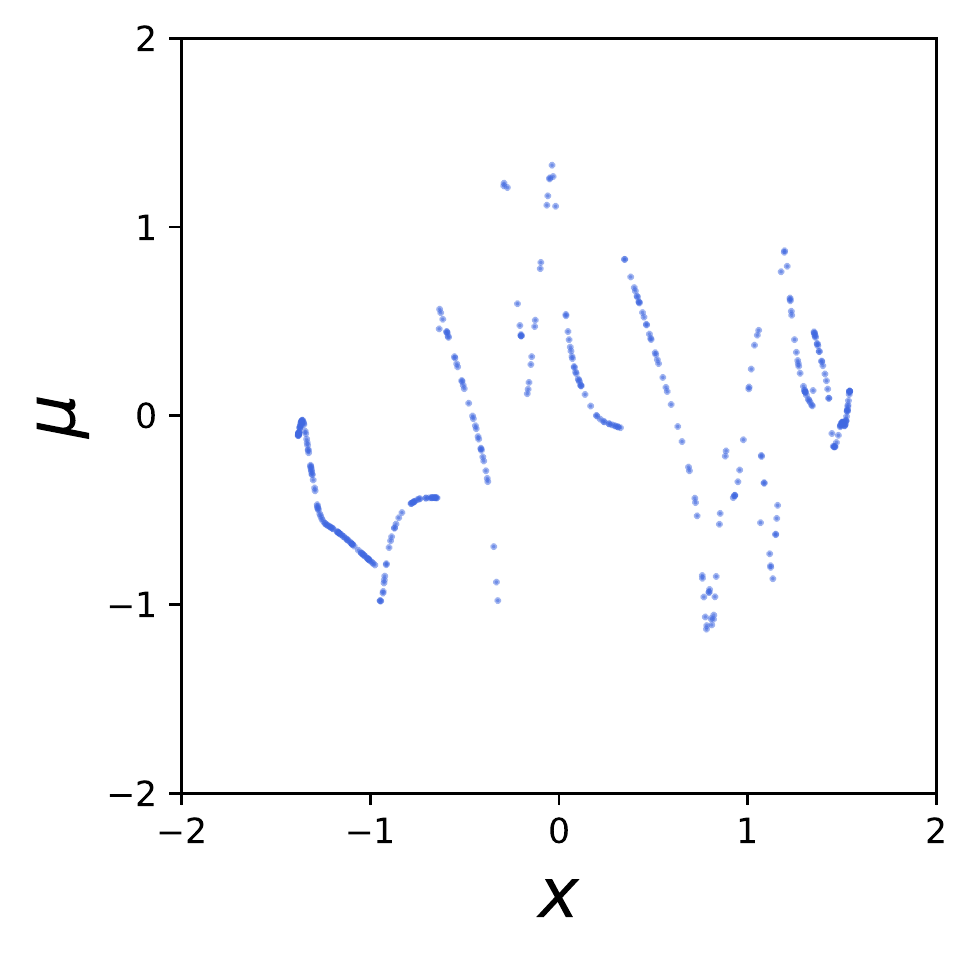}
                \\
                \raisebox{4.8\normalbaselineskip}[0pt][0pt]{\rotatebox[origin=c]{90}{Lissajous}} &
                \includegraphics[width=0.22\textwidth]{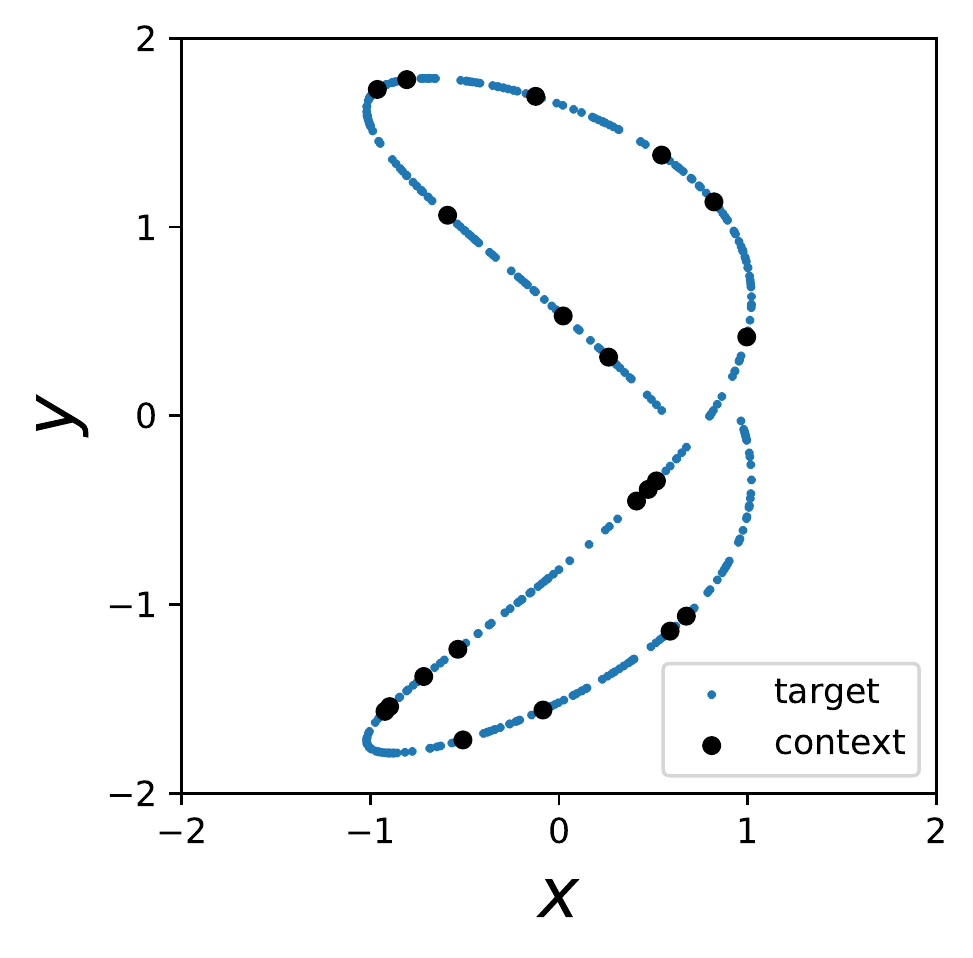} & 
	          \includegraphics[width=0.22\textwidth]{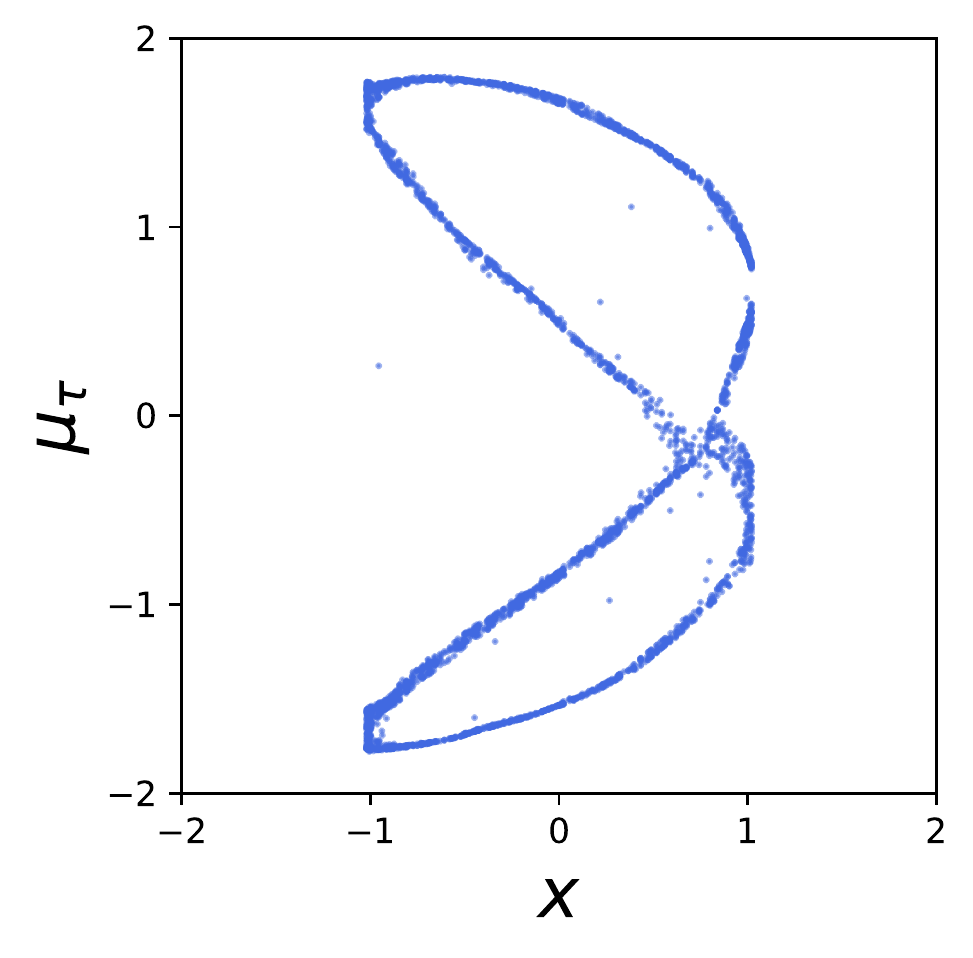} &
                \includegraphics[width=0.22\textwidth]{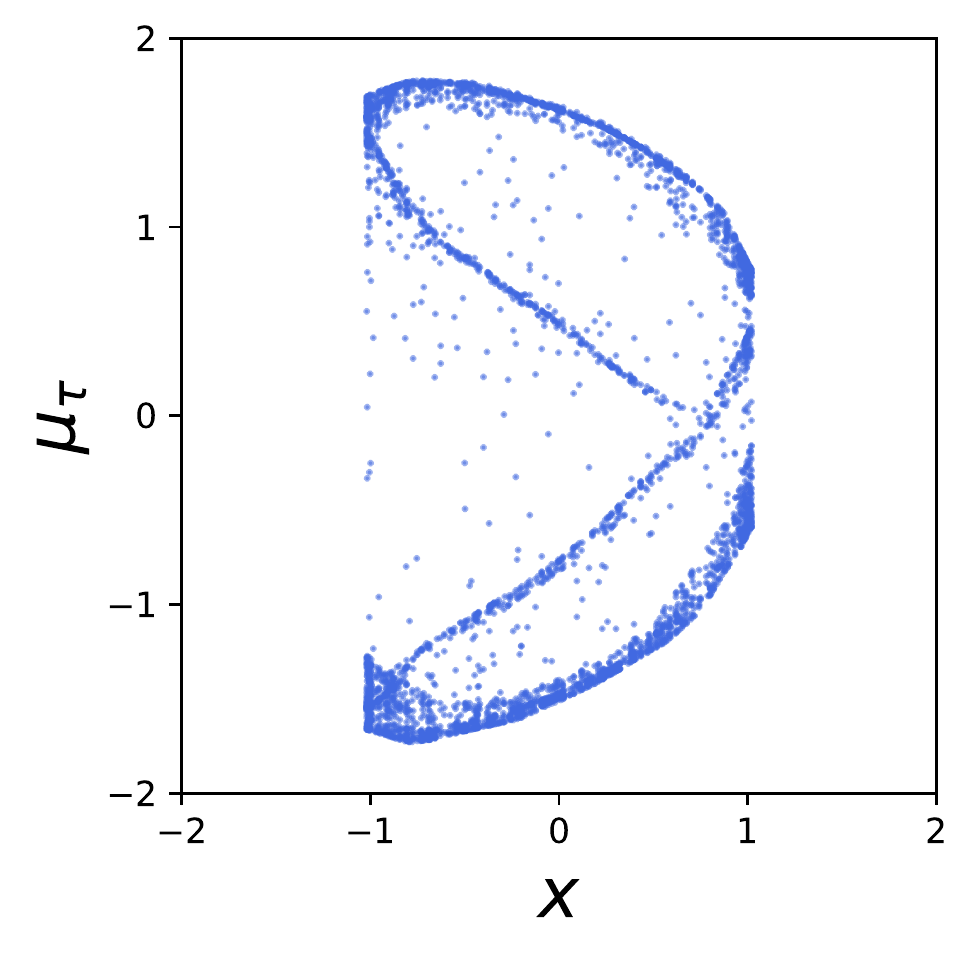} &
                \includegraphics[width=0.22\textwidth]{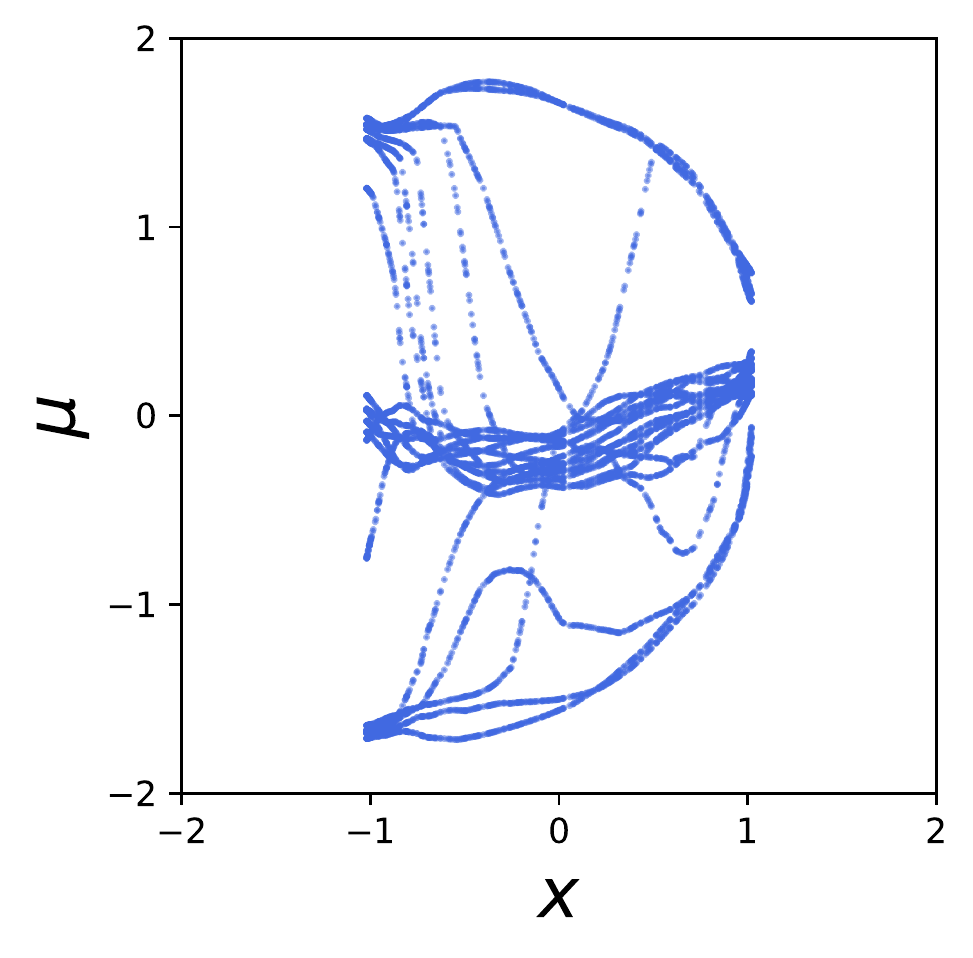} &
                \includegraphics[width=0.22\textwidth]{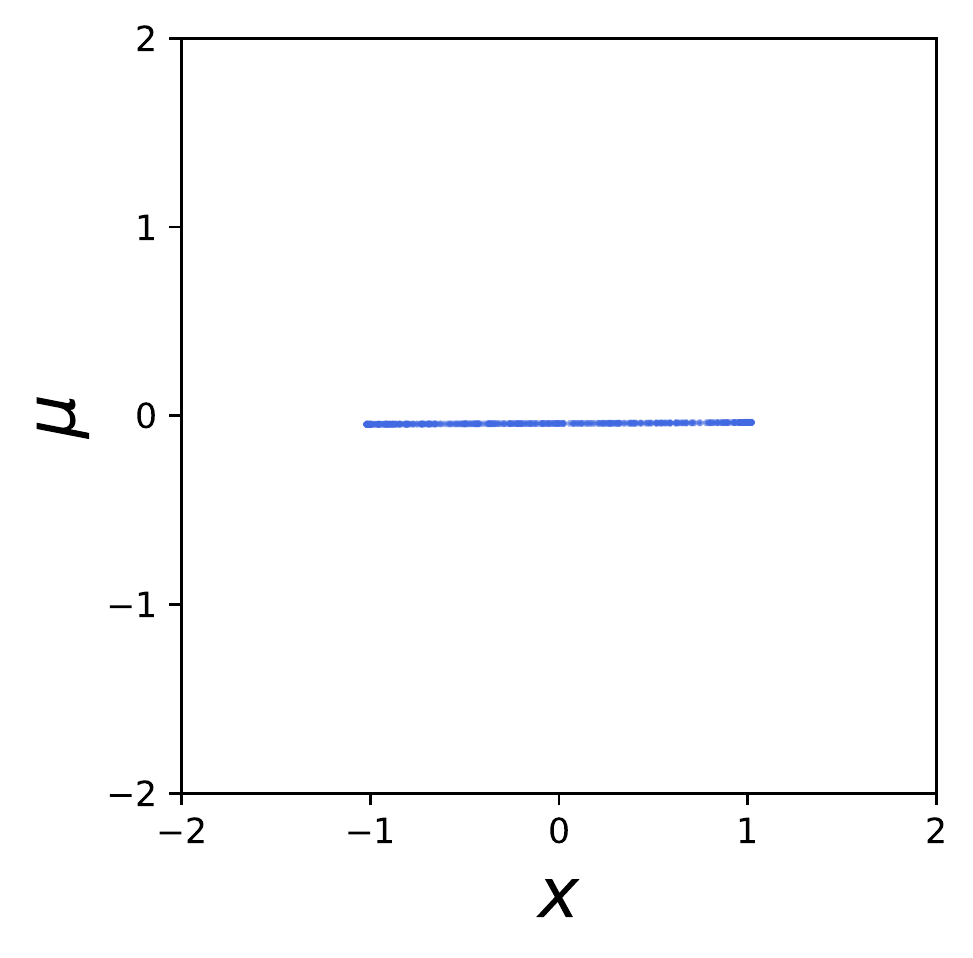} \\
                &a) Data & b) ACQNP & c) CQNP & d) BNP & e) CANP
            \end{tabular}
            \endgroup
        }
	\caption{Examples of predictions made by different methods on synthetic datasets. For A/CQNP, 10 randomly chosen conditional quantiles of $p(\ry \,|\,x)$ at each input location $x$ are plotted. For BNP, we plot the conditional means of the Gaussian predictive distributions obtained from 20 different sets of bootstrap contexts. Similarly, we plot the mean of CANP's conditional distribution as its predictions.}
        \label{fig: 1d-synthetic-benchmark}
\end{figure*}

\begin{table*}[!bp]
	% \vspace*{5pt}
	\centering
        \caption{Synthetic processes used in multimodal 1D regression experiments.}
        \label{table: 1d-benchmark-generative-processes}
        \scalebox{0.9}{
	{\setlength{\tabcolsep}{1.2pt}
	\begin{tabular}{lc} \toprule
		Process          & $g(s)=(g_x(s),\, g_y(s))$ \\ \midrule
		Double Sine   & $\quad g_x(s)=s\, , \, g_{y}(s) = \alpha_{1} \sin{(\omega_{1} s)} \, \mathds{1}_{(0, 0.5)}(p(s))+ \alpha_{2} \cos{(\omega_{2} s)}\,\mathds{1}_{[0.5, 1)}(p(s))$ \vspace*{5pt} \\ 
		Circle        & $\quad g_x(s)=\alpha \cos(s)+\delta \, , \, g_{y}(s) = \alpha\sin(s)+\delta$ \vspace*{5pt} \\
		Lissajous     & $\quad g_x(s)=\alpha_{1} \sin(\omega s + \delta) \, , \, g_{y}(s) = \alpha_{2} \sin(s)$ \vspace*{1pt}\\ \bottomrule
	\end{tabular}}
 }
\end{table*}

\section{Experiments}\label{sec: experiments}
We evaluate our proposed framework on one and two-dimensional regression tasks. Note that unlike other members of the NPs family that focus on building more expressive encoder-decoder blocks, our work is primarily concerned with the form of conditional likelihood and its effect on the model's performance. Nonetheless, we compare our A/CQNP with Conditional Attentive Neural Process (CANP) \citep{kim2019attentive}, and Bootstrapping Neural Process (BNP, \citet{lee2020bootstrapping}) as our baselines to provide a better overview of performance gains obtained by modifying different components of vanilla CNP.
Note that, although NPs share fundamental properties with GPs, they usually are not compared directly because of the different training regimes \citep{kim2019attentive}. While NPs are trained on different functions sampled from the underlying generative process, GPs are fit to observations corresponding to one realization of the process.
In order to compare the goodness of fit across different methods, we report the  log-likelihood on context and target data separately. Methods with higher context log-likelihood offer better reconstructions of context points and hence, are less prone to under-fitting, while higher target log-likelihood indicates more accurate predictions \citep{kim2019attentive, lee2020bootstrapping}. 
Detailed information on model architectures, training, and testing procedures are included in the supplementary materials.
\footnote{Code at {\url{https://github.com/peiman-m/ACQNP}}}

\subsection{Synthetic Data}\label{subsec: synthetic-data-results}
\begin{figure*}[!tp]
	\centering
        \scalebox{0.85}{
            \begingroup
            \setlength{\tabcolsep}{-0.5pt}
            \begin{tabular}{ccc} 
	          \includegraphics[width=0.33\textwidth]{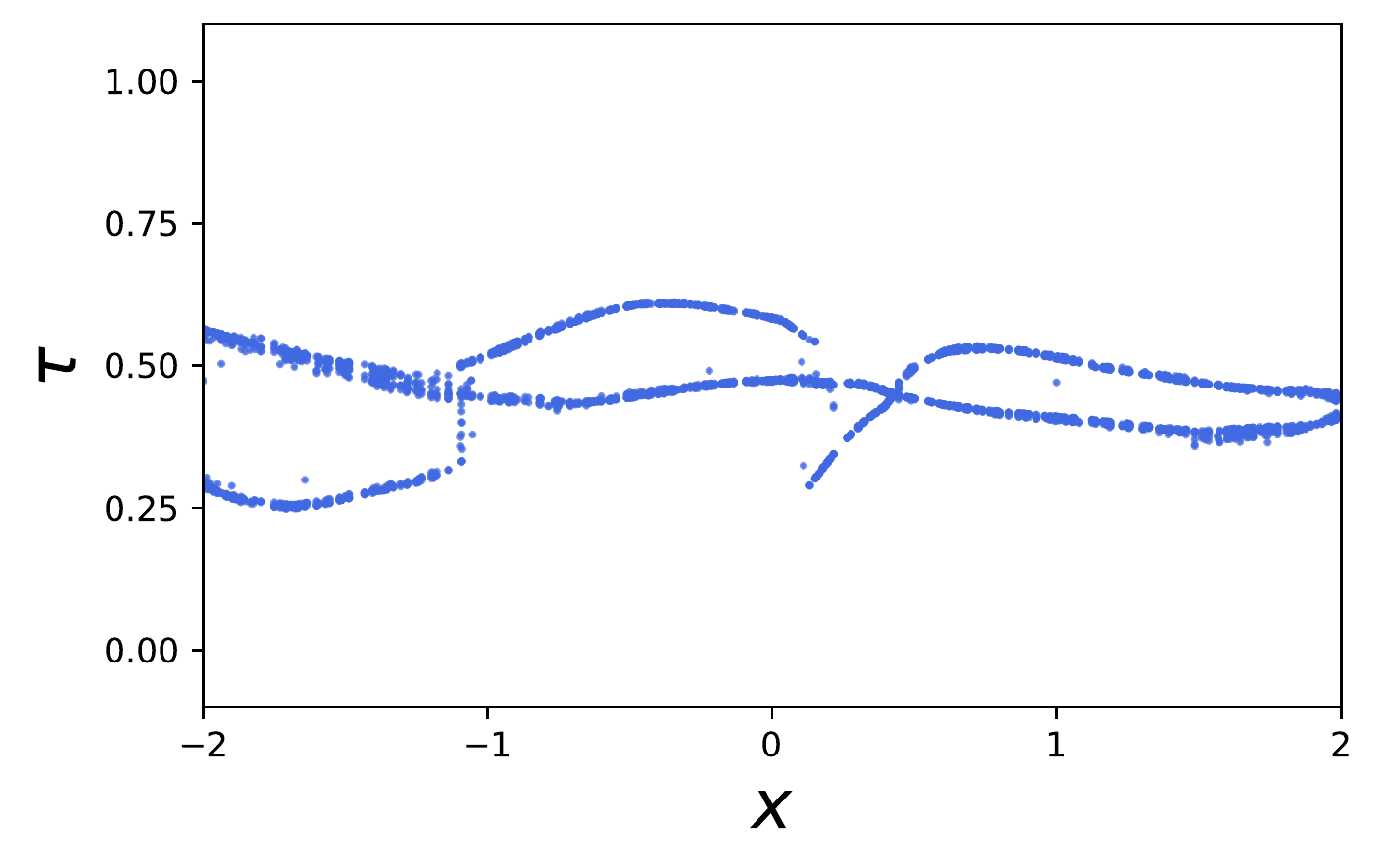} & 
	          \includegraphics[width=0.33\textwidth]{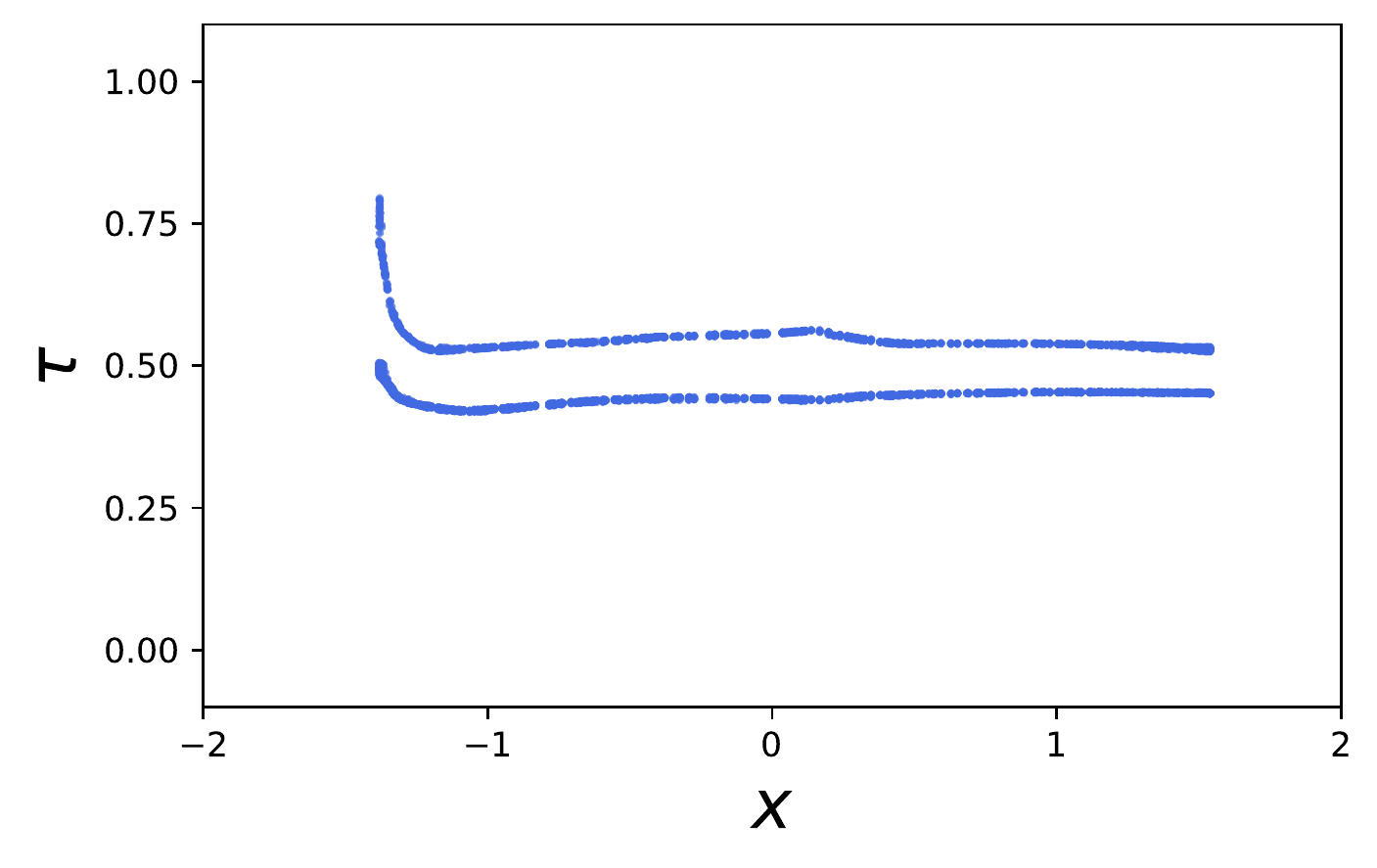} & 
	          \includegraphics[width=0.33\textwidth]{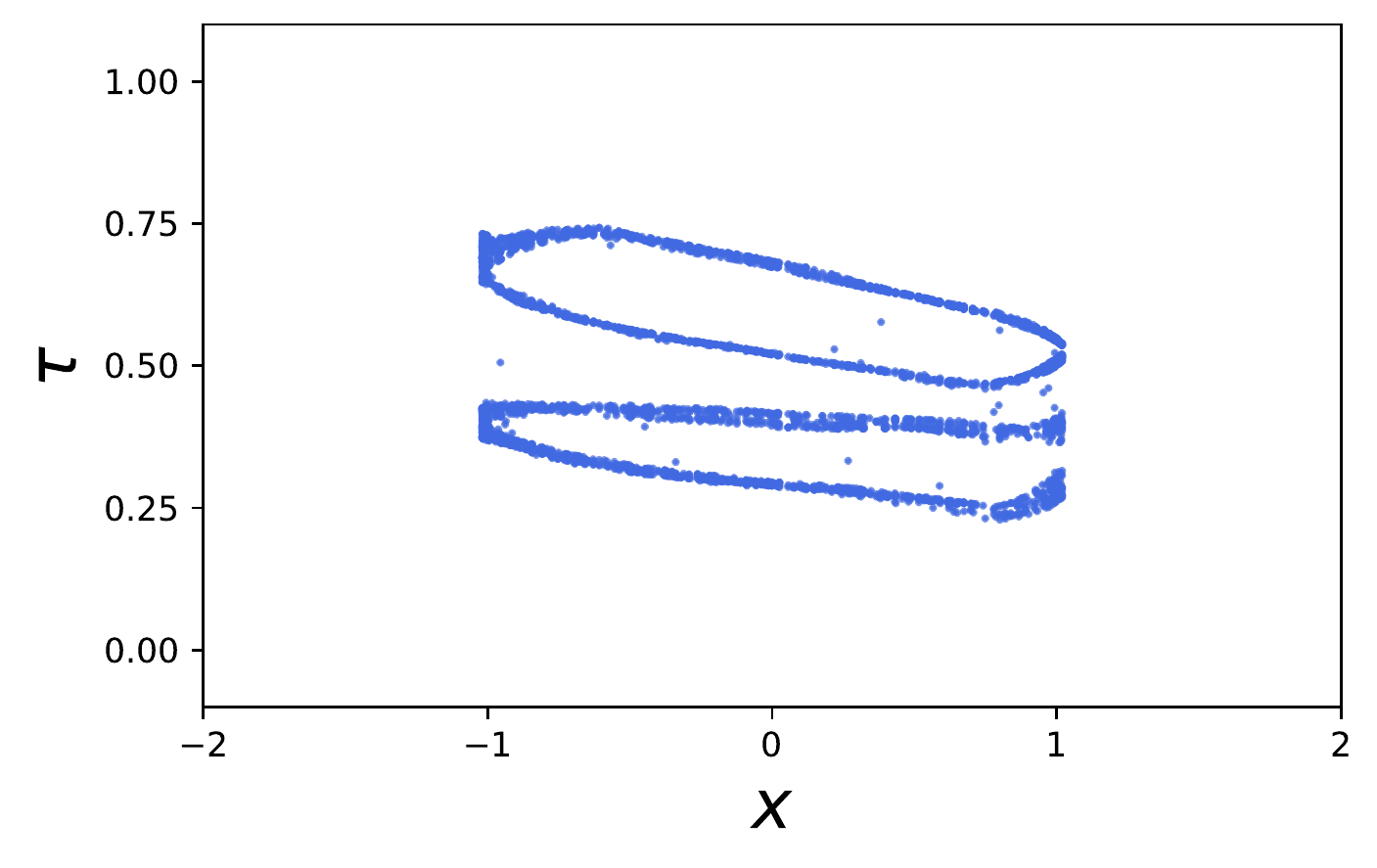} \\
                a) Double Sine  & b) Circle  & c) Lissajous
            \end{tabular}
            \endgroup
        }
	\caption{Distribution of the adapted $\tau$ levels corresponding to the estimated quantiles in figure \ref{fig: 1d-synthetic-benchmark}b after applying the transformation $\psi$ in equation \ref{eq: ACQNP-liklihood}.}
        \label{fig: 1d-benchmark-tau-distribution}
\end{figure*}

We start our study by examination of each model over several synthetically generated datasets. Each collection consists of a handful number of functions sampled from a known stochastic process. In each iteration, a batch of $n_b$ functions $\mathcal{G} = \{g_k\}_{k=1}^{n_b}$ are sampled from a stochastic process such that $g_k: \mathbb{R} \rightarrow \sX \times \sY$ and $g_k(s) = (g_{k, x}(s),\, g_{k, y}(s))$. For each $g_k$, a set $\mathcal{S}_k$ of $N_{\mathrm{total}}$ random input locations is chosen where $\mathcal{S}_k = \{ s_{k,l} \}_{l=1}^{N_{\mathrm{total}}}$ and $s_{k, l} \sim \mathcal{U}[I_\mathrm{min},\, I_\mathrm{max}]$ ($I_\mathrm{min}$ and $I_\mathrm{max}$ are fixed constants). Applying $g_k$ to the corresponding $\mathcal{S}_k$ will yield a collection of pairs $\mathcal{E}_k = \{(x_{k, l},\, y_{k, l})\}_{l=1}^{N_{\mathrm{total}}}$ where $(x_{k, l},\, y_{k, l})=(g_{k, x}(s_{k, l}),\, g_{k, y}(s_{k, l}))$. By repeating this process for each $k$, we end up with a hierarchical dataset\footnote{A hierarchical dataset is a collection of observations from many functions sharing some underlying characteristics.} $\mathcal{E} = \{\mathcal{E}_k\}_{k=1}^{n_b}$ \citep{garnelo2018conditional, garnelo2018neural}. The variable $s$ is discarded after sampling $\mathcal{E}_k$. In the course of training, the total number of data points $N_{\mathrm{total}}$ is randomly chosen for each batch such that $N_{\mathrm{total}} \sim \mathcal{U}[6, 100]$. Lastly, each $\mathcal{E}_k$ is split into context and target sets by choosing a random index $N_{\mathrm{context}} \sim \mathcal{U}[3, N_{\mathrm{total}}-3]$ and setting $\mathcal{E}_{k, {\mathrm{context}}} = \{(x_{k, l},\, y_{k, l})\}_{l=1}^{N_{\mathrm{context}}}$ and $\mathcal{E}_{k, {\mathrm{target}}} = \{(x_{k, l},\, y_{k, l})\}_{l=N_{\mathrm{context}}+1}^{N_{\mathrm{total}}}$.
During testing, however, we fix $N_{\mathrm{total}}=500$ and select $N_{\mathrm{context}} \sim \mathcal{U}[3, 100]$. Each method is trained and tested over $10^5$ and $10^3$ batches with $n_b=128$ and $n_b=16$, respectively. Note that, unlike training data which is generated in each iteration of training, we fix the testing data across different models by generating them in advance. We consider data arising from the three stochastic processes described in table \ref{table: 1d-benchmark-generative-processes} with the following choice of parameters:
\begin{itemize}
  \item Double-Sine: $s \sim \mathcal{U}[-2,\,2)$, $\alpha_{1}, \alpha_{2} \sim \mathcal{U}[0.5, 1.5)$, $\omega_{1}, \omega_{2} \sim \mathcal{U}[1, 3)$ and $p(s) \sim \mathcal{U}[0, 1)$
  \item Circle: $s \sim \mathcal{U}[-\pi,\,\pi)$, $\alpha \sim \mathcal{U}[0.5, 1.5)$, $\delta \sim \mathcal{U}[-0.5, 0.5)$
  \item Lissajous: $s \sim \mathcal{U}[-\pi,\,\pi)$, $\alpha_{1}, \alpha_{2} \sim \mathcal{U}[1, 2)$, $\omega \sim [0.5, 2)$, $\delta \sim \mathcal{U}[0, 2)$
\end{itemize}
Table \ref{tabel: 1d-benchmark-synthetic-results} summarizes the predictive log-likelihood of different methods over testing data. We see that A/CQNP constantly outperforms baselines despite using the vanilla deterministic encoder in CNP which: 1) does not incorporate any latent variable for capturing the functional uncertainty and correlations as in BNP, and 2) does not enjoy the expressive context representations provided by the attention mechanism used in CANP. As depicted in figure \ref{fig: 1d-synthetic-benchmark}, CANP fails to capture multimodality and instead tends to make predictions that resemble the average of modes. This happens less severely with BNP. Nonetheless, the sampled curves bounce between the modes which results in unnecessarily wide prediction bands. CQNP, on the other hand, provides decent fits which are further polished by incorporating the quantile adaptation structure in ACQNP. 
Figure \ref{fig: 1d-benchmark-tau-distribution} illustrates the quantile levels $\tau$ that were adapted by ACQNP for predicting the quantiles shown in figure \ref{fig: 1d-synthetic-benchmark}b. We see that ACQNP behaves in line with our motivations behind adaptive quantile regression discussed in section \ref{subsec: adaptive-quantile-regression}. The distribution of $\tau$ imitates the distribution of the data, in the sense that $p(\ry \,|\, x, \mathcal{D})$ and $p(\tau \,|\,x, \mathcal{D})$ have the same number of modes almost everywhere. Additional experimental results on unimodal regression tasks are provided in the supplements.
\input{tables/synthetic.tex}

\subsection{Speed-Flow Data}\label{subsec: speed-flow-experiment}
\begin{figure*}[!h]
	\centering
        \scalebox{0.9}{
            \begingroup
            \setlength{\tabcolsep}{-1.5pt}
            \begin{tabular}{c@{\hskip 2pt}ccccc}
                \raisebox{4.4\normalbaselineskip}[0pt][0pt]{\rotatebox[origin=c]{90}{Lane 2}} &
                \includegraphics[width=0.22\textwidth]{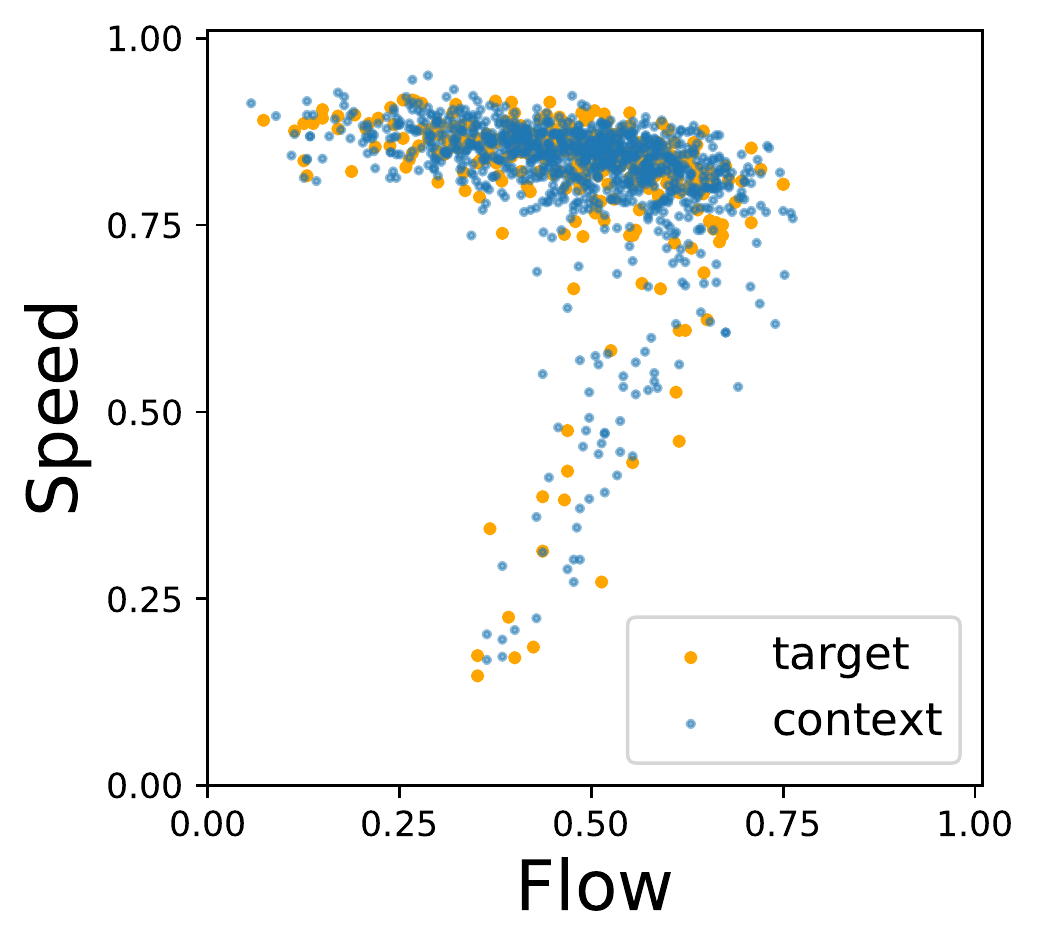} & 
	          \includegraphics[width=0.22\textwidth]{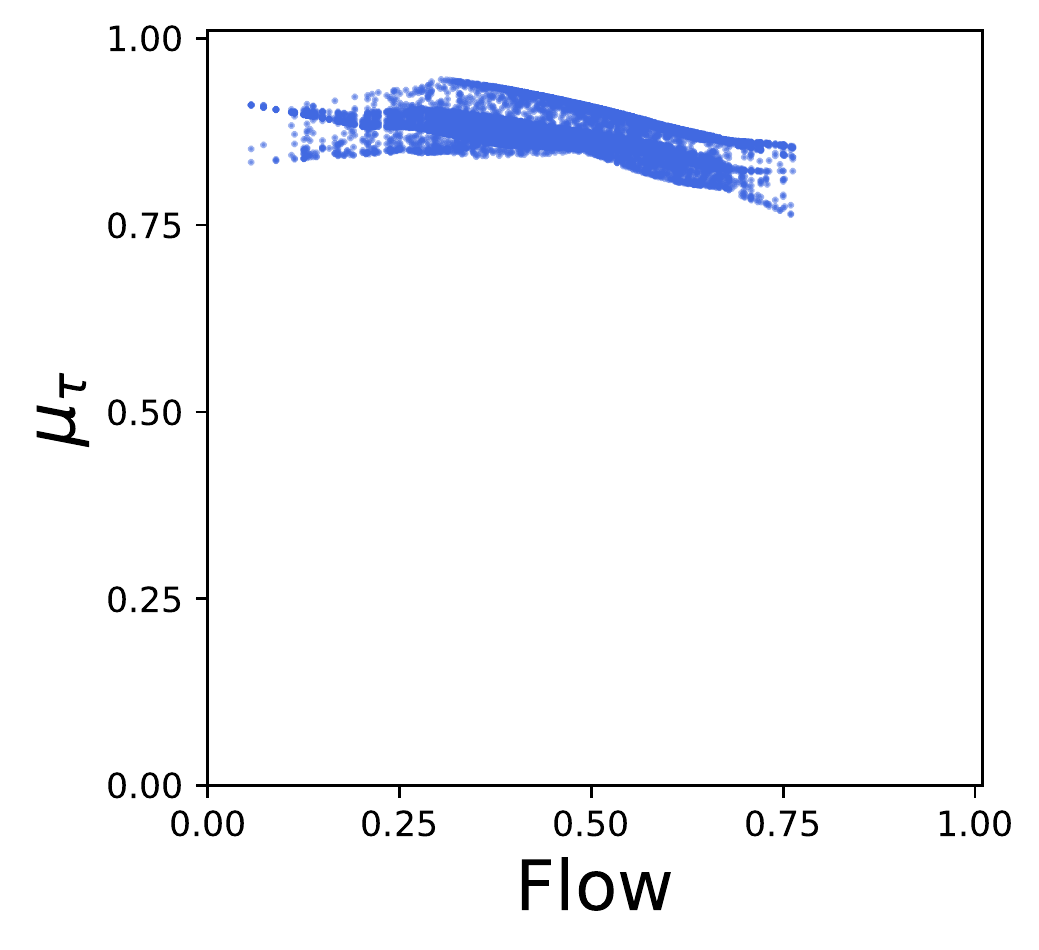} &
                \includegraphics[width=0.22\textwidth]{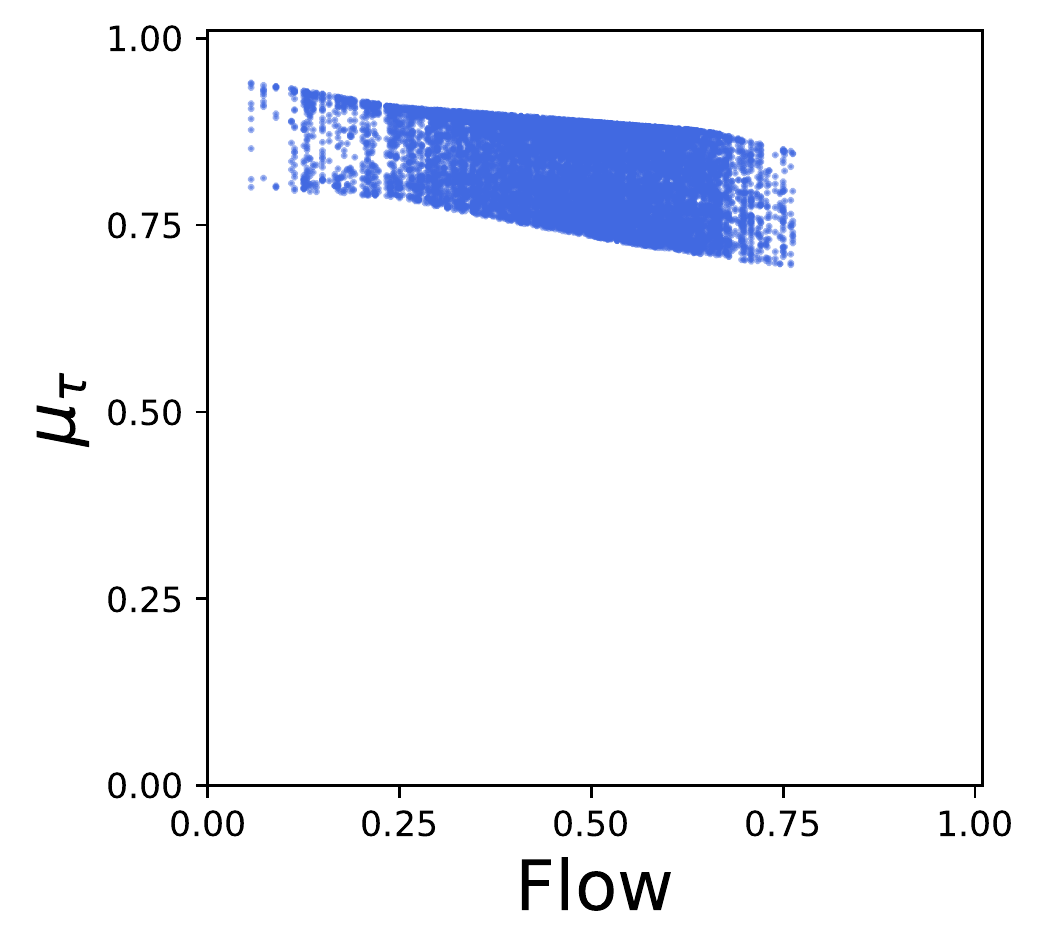} &
                \includegraphics[width=0.22\textwidth]{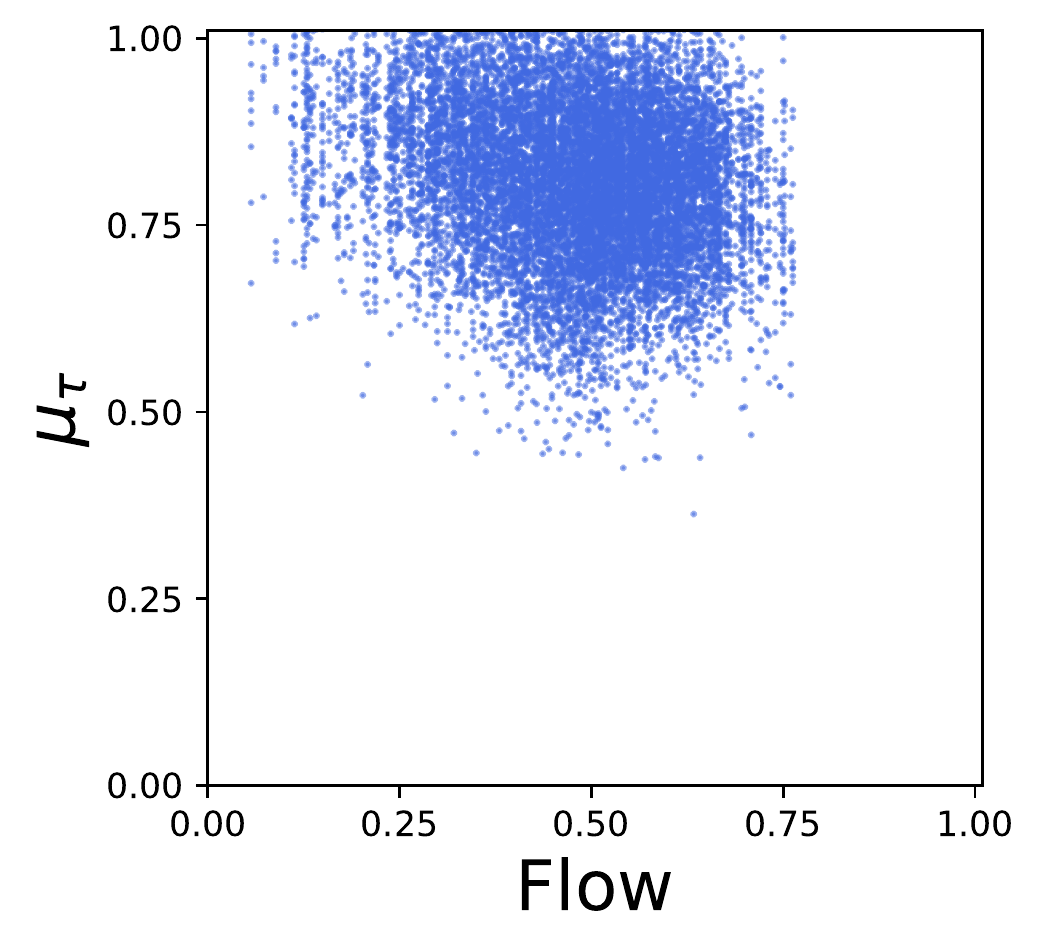} &
                \includegraphics[width=0.22\textwidth]{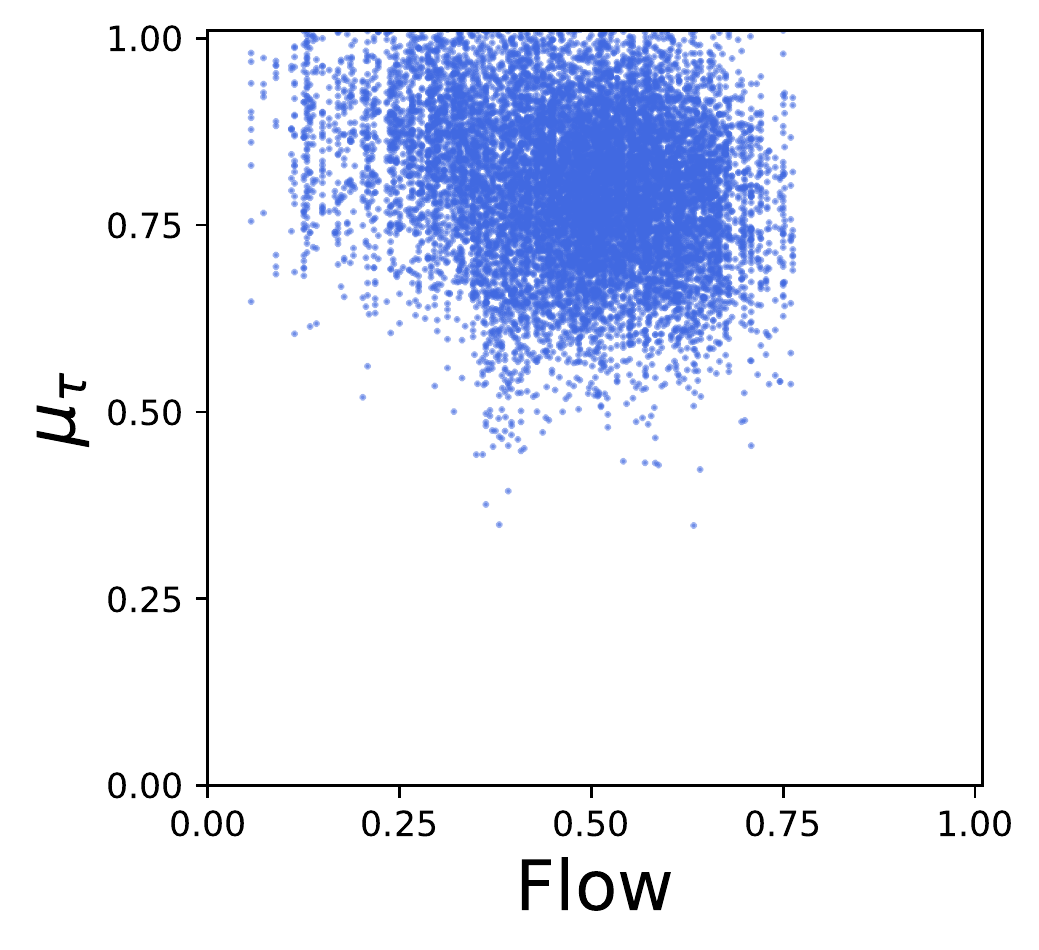} 
                \\
                \raisebox{4.4\normalbaselineskip}[0pt][0pt]{\rotatebox[origin=c]{90}{Lane 3}} &
                \includegraphics[width=0.22\textwidth]{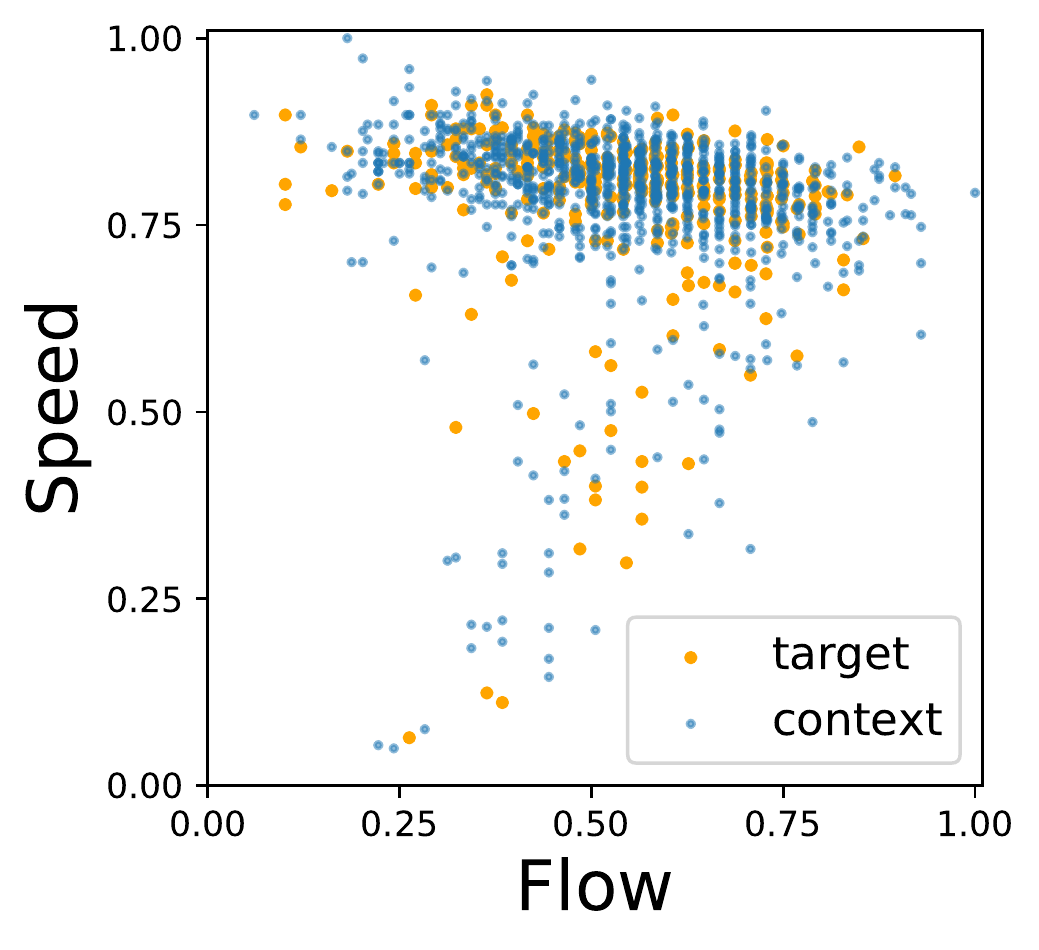} & 
	          \includegraphics[width=0.22\textwidth]{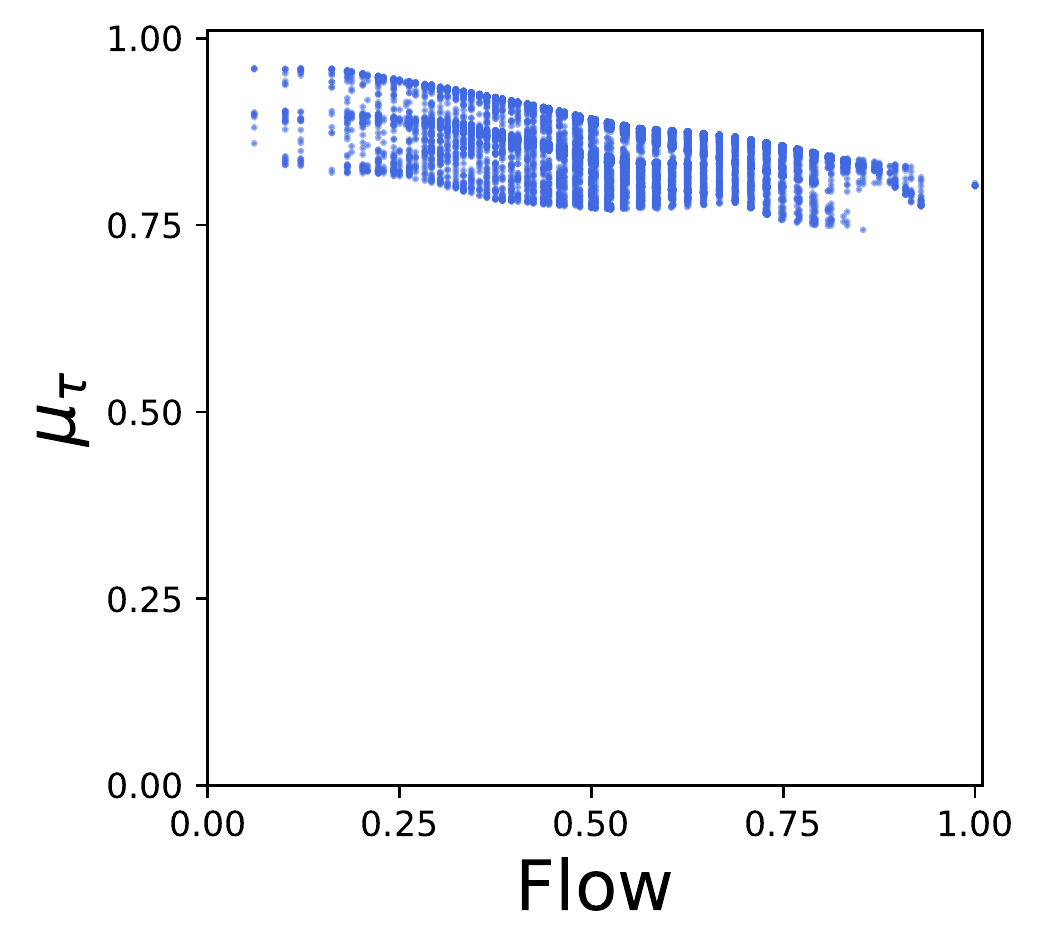} &
                \includegraphics[width=0.22\textwidth]{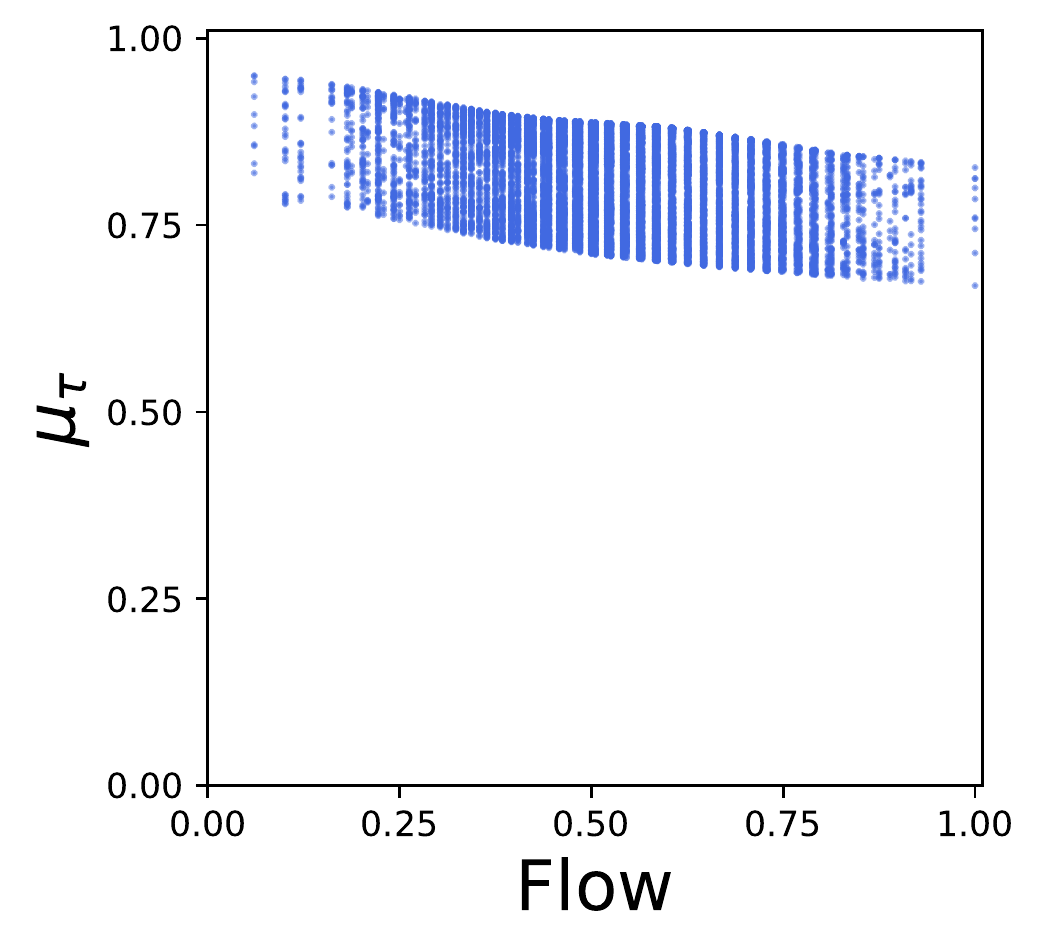} &
                \includegraphics[width=0.22\textwidth]{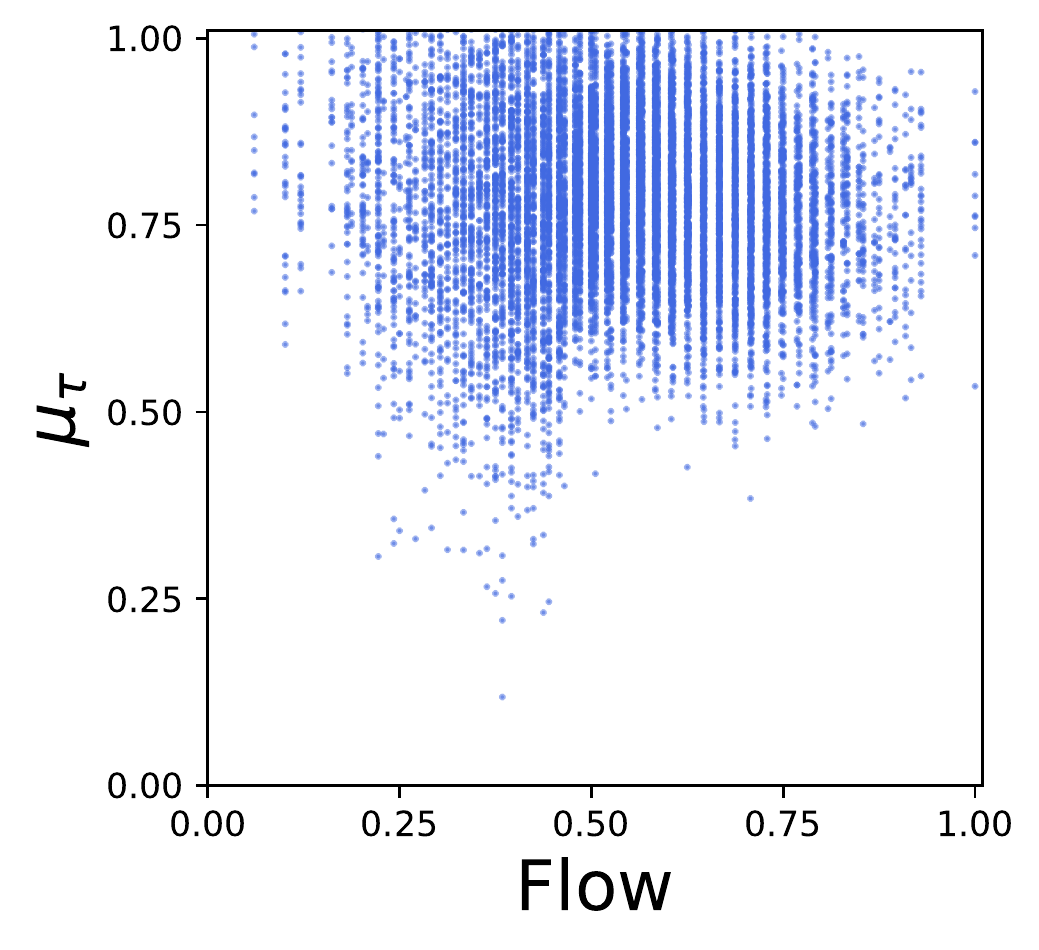} &
                \includegraphics[width=0.22\textwidth]{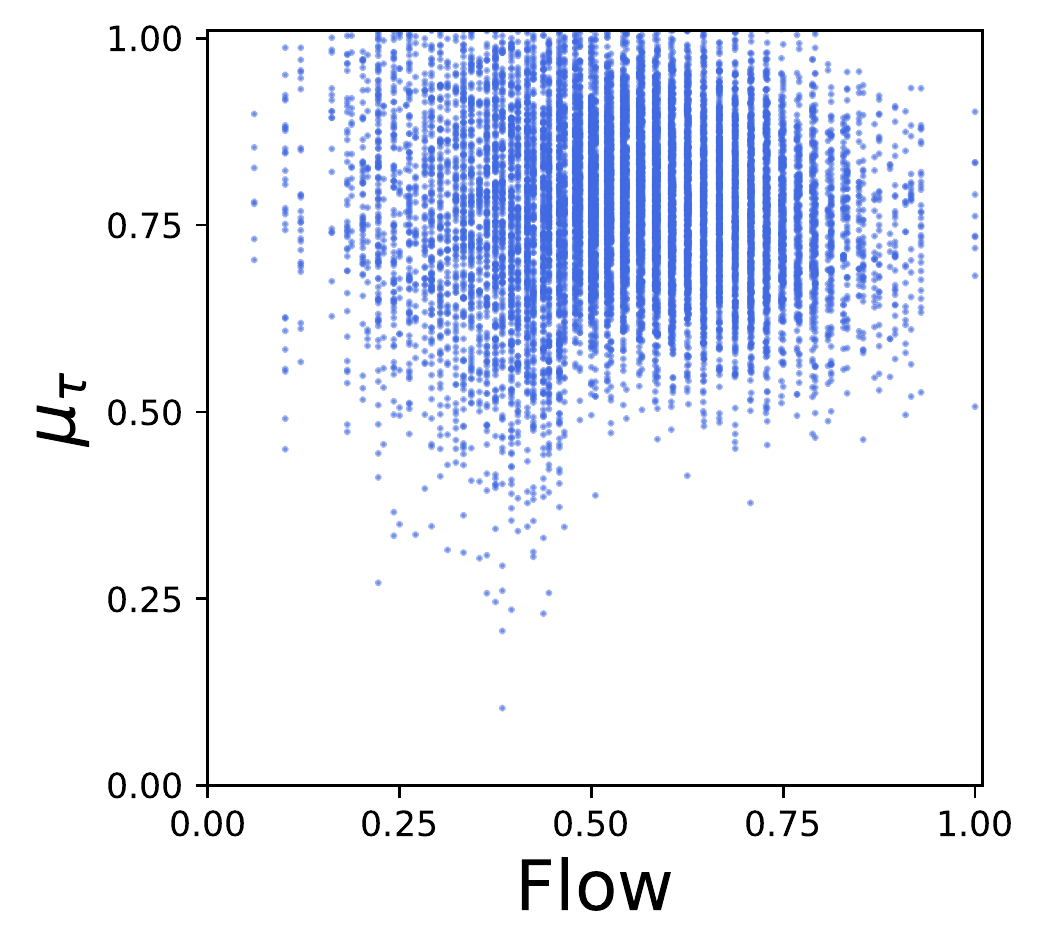} 
                \\
                &a) Data & b) ACQNP & c) CQNP & d) BNP & e) CANP
            \end{tabular}
            \endgroup
        }
	\caption{First column from left includes the ground truth data. Second and third columns show the quantiles of the predictive distribution of A/CQNP at 10 different levels. Fourth column shows the ensembled quantiles at the same levels for 5 bootstrap context sets. Last column shows the conditional quantiles of CANP at the same levels.}
        \label{fig: speed-flow}
\end{figure*}
The problem of traffic speed prediction has been widely investigated in transportation science with applications in approximating the expected arrival time \citep{einbeck2006modelling}. A key factor in such models is the traffic flow which is usually presented by speed-flow diagrams. As a case study, we consider the speed-flow data collected by \citet{petty1996freeway}. The dataset contains the speed-flow diagrams of lanes 2 and 3 of a four-lane Californian freeway with 1318 measurements for each lane (Figure \ref{fig: speed-flow}a) and is included in the R-package \texttt{hdrcde} \citep{hdrcde}. This collection can be viewed as an example of a hierarchical dataset where observations from different lanes correspond to different realizations of a random process. The hierarchical structure of this data makes it an ideal fit with NPs as they allow for sharing information among different lanes. 
We randomly select $75\%$ of each lane's observations ($\approx 988$) for training and use the rest for testing. The speed and flow values are both scaled to $[0, 1]$.
In each training iteration, we split the training data into context and target sets such that $N_{\mathrm{context}} \sim \mathcal{U}[500, 985]$. For testing, however, we take the context and target sets to be the training and testing data, respectively. Table \ref{tabel: speed-flow-benchmark-results} provides a quantitative comparison of different approaches, with A/CQNP being the best-performing model on this real-world task. As shown in figure \ref{fig: speed-flow}, the Gaussian predictive distributions in BNP and CANP, which resemble the conditional mean regression, lead to wide prediction bands compared to A/CQNP. This is due to the sensitivity of mean estimators in dealing with the less dense cloud of data points at the bottom which can be interpreted as outliers \citep{feng2020statistical}. 
In contrast to the mean which acts on a global level, the local nature of quantiles makes them more robust to the tail behavior.
\input{tables/speed-flow.tex}

\input{tables/image-completion.tex}
\subsection{Image Completion}\label{subsec: image-completion-experiment}
A collection of images can also be thought of as a hierarchical dataset where each image is a realization of some random process mapping 2D pixel coordinates to pixel intensities. Motivated by this observation, image inpainting can be framed as a regression problem where conditioned on a set of observed pixels, we are interested in filling the missing regions of the image.
MNIST \citep{lecun1998gradient}, Fashion-MNIST \citep{xiao2017fashion}, SVHN \citep{netzer2011reading}, Omniglot \citep{lake2015human} (resized to 32 $\times$32) and FreyFace \citep{roweis2001global} are the datasets that we consider here. Except for FreyFace, we use the default train/test split used by the publishers. For FreyFace, we randomly select $75\%$ of the images for training and keep the rest for testing. For all the benchmarks, the pixel values and pixel coordinates are rescaled to $[0, 1]$ and $[-1,\, 1]$ respectively.
In each case, we take $\mathcal{E}_{k}$ as the set of all image pixels. Similar to section \ref{subsec: synthetic-data-results}, the context and target sets in both training and testing are chosen such that $N_{\mathrm{context}} \sim \mathcal{U}[3, N_{\mathrm{total}}/2)$ where $N_{\mathrm{total}}$ is the number of image pixels. Table \ref{tabel: 2d-benchmark-image-completion-results} shows that A/CQNP substantially outperforms the baselines across all datasets in terms of predictive log-likelihood. This holds for both context and target sets revealing that A/CQNP yields better reconstruction of context data. Moreover, results from Omniglot experiments suggest that A/CQNP has better generalization capabilities as the default test split has distinct classes from training.

\subsection{Ablation Study}\label{subsec: ablation-study}
\begin{figure*}[h!]
	\centering
        \scalebox{0.75}{
            \begingroup
            \setlength{\tabcolsep}{0pt}
            \begin{tabular}{ccc}
                \includegraphics[width=0.5\textwidth]{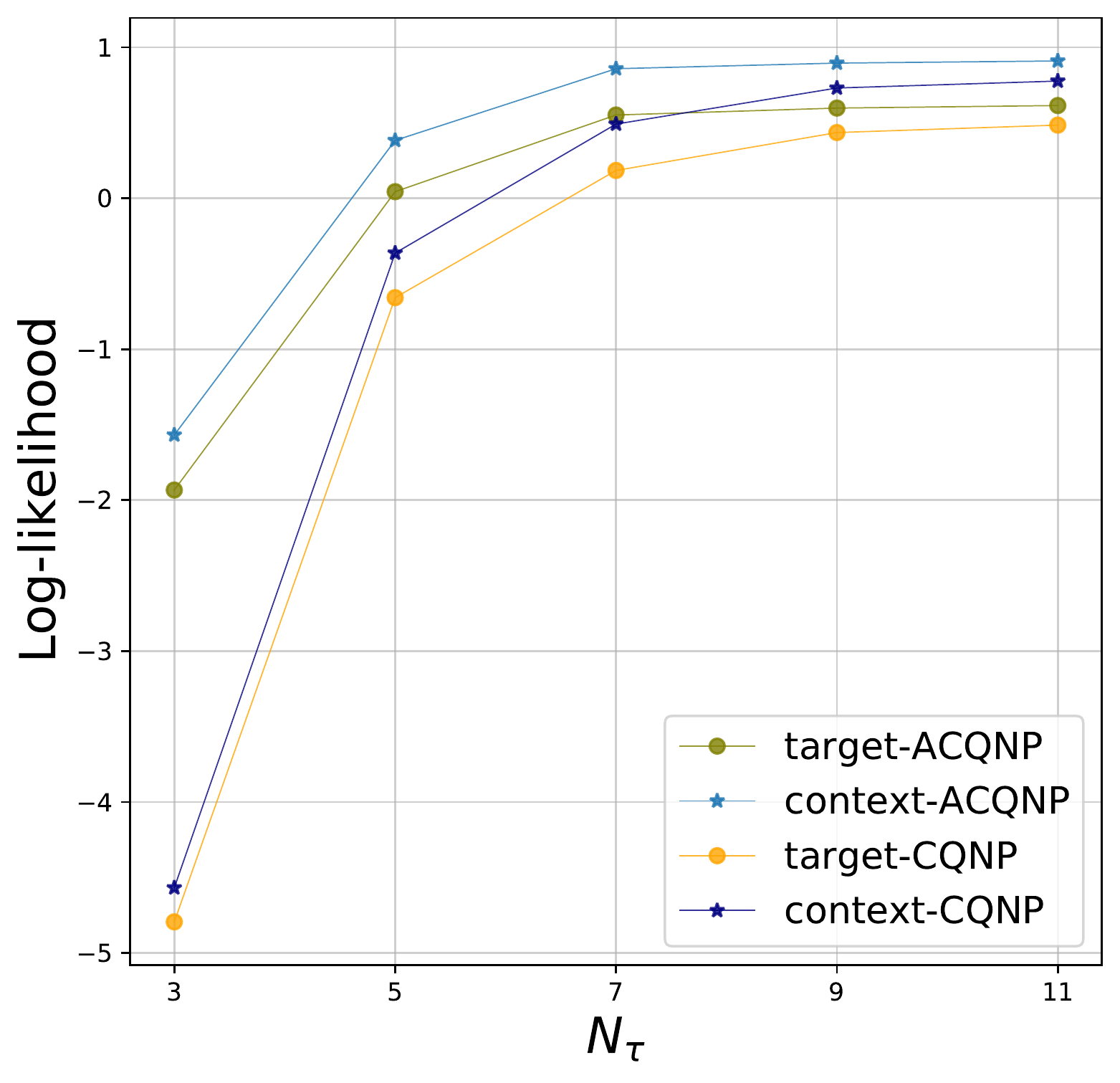} &
                \qquad &
                \includegraphics[width=0.5\textwidth]{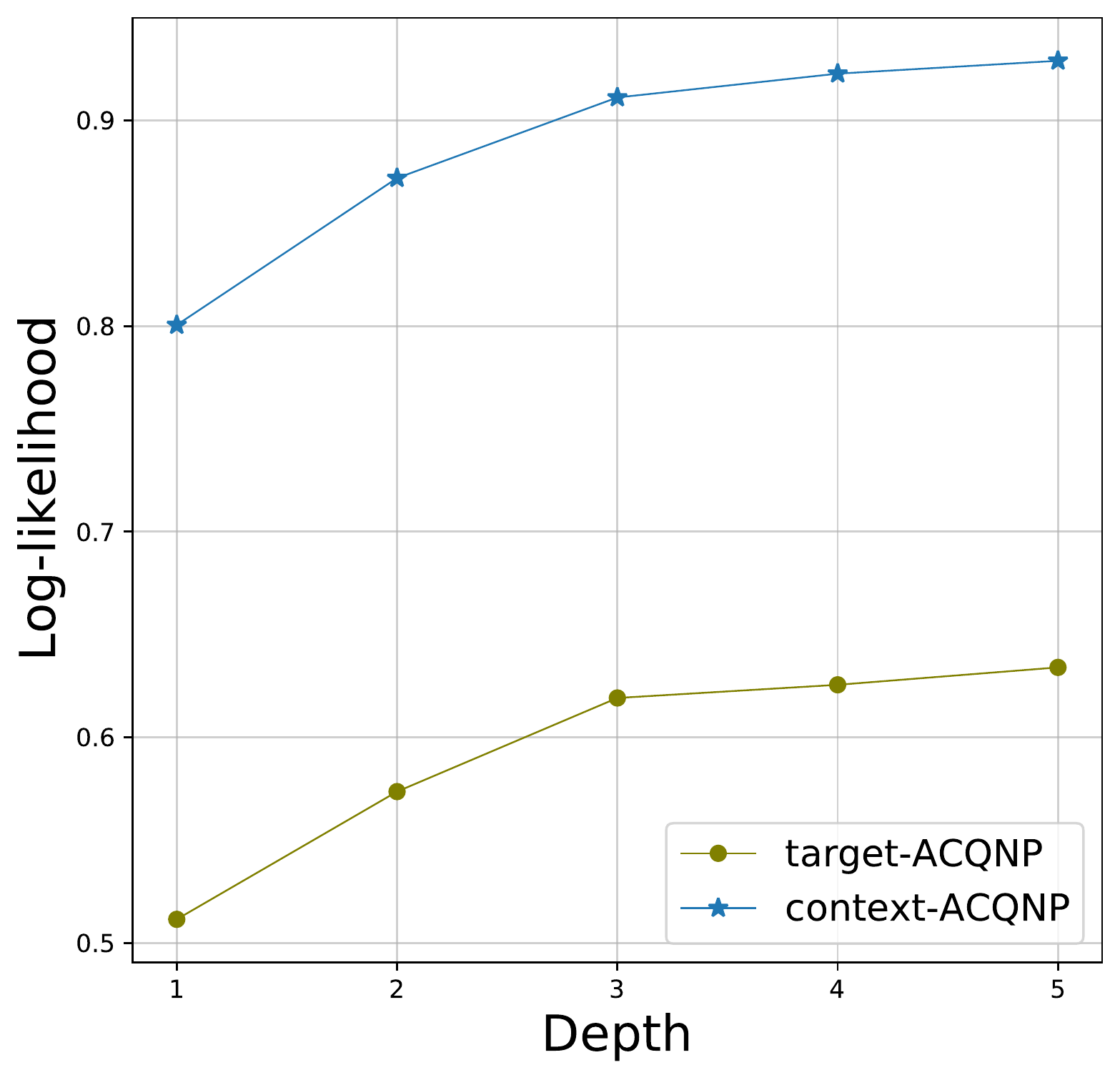}
            \end{tabular}
            \endgroup
        }
	\caption{\textbf{Left}: Comparison of the predictive log-likelihood over context and target points for A/CQNP with different values of $N_\tau$ during testing (6 seeds); \textbf{Right}: Predictive performance of ACQNP versus the depth of $\psi$ (6 seeds).}
        \label{fig: ablation-adaptive-quantile}
\end{figure*}
\paragraph{Number of quantiles $N_{\tau}$.}
As mentioned earlier in section \ref{subsec: CQNP}, we approximate the conditional likelihood by $N_{\tau}$ Monte Carlo samples. While using a larger sample size offers more precise approximation, it demands further computational resources which highlights the importance of a rather fine-grained scheme for sampling $\tau$ instead of random draws from a uniform distribution, especially when we are restricted to work with a small $N_{\tau}$. To check if the adaptive mechanism can alleviate this issue, we compare the predictive performance of ACQNP against CQNP on Lissajous curves studied in section \ref{subsec: synthetic-data-results}. 
Both models have the same architectural design for their encoder/decoder modules and are trained with $N_{\tau}=50$. During testing, however, we evaluate each method with $N_{\tau} \in \{3, 5, 7, 9, 11\}$ as depicted in figure \ref{fig: ablation-adaptive-quantile}a. Note that in addition to testing data, we also fix the input quantile levels $u$ to be $\{u_0, u_0+d, \dotsc, u_0+(N_{\tau}-2)d, u_1\}$ where $u_0=0.001$, $u_1=0.999$, and $d=\frac{u_1-u_0}{N_{\tau}-1}$.
It can be seen that the log-likelihood of CQNP decreases significantly with smaller sample sizes, whereas ACQNP suffers less as it can manage to use its few shots efficiently and locate informative quantiles.

\paragraph{Flexibility of $\psi$.}
The adaptive process that we followed in this paper works by transforming $u$ through some nonlinear mapping $\psi$ introduced in equation \ref{eq: al-mixture-with-adaptive-quantiles}; thus, the choice of this mapping is expected to affect the performance. Throughout this work, we modeled $\psi$ with a fully-connected neural network. We investigate the effect of the depth of the neural network as a measure of its expressive power. As illustrated in figure \ref{fig: ablation-adaptive-quantile}, a deeper neural network improves the overall performance of the model. However, the performance gain comes at the cost of additional memory usage and computational complexity which needs to be considered as a tradeoff. 
 
\section{Related Works}

\paragraph{Vector quantile regression}
Quantile regression (QR), introduced by \citet{koenker1978regression}, is a compelling statistical technique that can be used for studying the dependence between random variables by modeling the conditional quantiles of a target variable as a function of some explanatory variables. Unfortunately, the QR framework only considers scalar target variables as the notion of quantile is not well-defined in higher dimensions. We can apply QR to scalar components of a vector-valued response variable by assuming independence among them. This assumption, however, is usually violated. Recently, \citet{carlier2016vector, chernozhukov2017monge, carlier2017vector} introduced vector quantiles as extensions to the univariate quantiles which allows for vector quantile regression (VQR). \citet{rosenberg2022fast} propose nonlinear vector quantile regression (NL-VQR) which drops the restrictive specification of a linear conditional quantile function. They further introduce a solver allowing for applying their method to large datasets.

\paragraph{Function-space inference}
There is a growing line of research on using neural networks for direct parameterization of distributions over functions. NPs and their variants are well-known byproducts of this viewpoint. \citet{ma2019variational} introduced implicit processes (IPs) as priors over functions by placing an implicit joint distribution over any finite collection of random variables. However, they use GPs for approximating the intractable posterior of IPs which is: 1) computationally expensive, and 2) limited by the Gaussian likelihood assumption. Sparse implicit processes (SIPs, \citet{rodríguez2022function}) try to address these issues by relying on inducing points \citep{snelson2005sparse} and using a mixture of Gaussians as the predictive distribution. \citet{yang2020energy} use energy-based models (EBMs) to construct a family of expressive stochastic processes for exchangeable data. The additional flexibility, however, comes at the cost of complicated training and inference schemes requiring several approximations. Inspired by contrastive methods \citep{durkan2020contrastive, gutmann2010noise, gondal2021function}, \citet{mathieu2021contrastive} drop the explicit likelihood requirement used in NPs which necessitates exact reconstruction of observations. Despite higher tolerance in facing noisy high-dimensional data, their method is incapable of uncertainty quantification.

\section{Conclusions}
In this paper, we proposed Conditional Quantile Neural Processes (CQNPs), a new member of the CNP family that leverages advances in quantile regression to 
increase the expressive power of CNPs in modeling heterogeneous distributions.
Furthermore, we introduced an extended framework for quantile regression, named adaptive quantile regression, where instead of fixing the quantile levels, the model gets to choose which quantiles to estimate. 
Our experiments with several synthetic and real-world datasets showed that A/CQNPs improve the predictive performance of CNPs across regression tasks in terms of log-likelihood, and faithfully model multimodality in predictive distributions.

\begin{acknowledgements}
This work was supported in part by the National Science Foundation under award 1848596. We also thank Texas A\&M High Performance Research Computing for providing computational resources to perform experiments in this work.
\end{acknowledgements}
% References
{\bibliography{bibfile}}

\onecolumn %% Turn this off if single column is desired for the supplement
\appendix
\input{supp}

\end{document}

%% file: math_commands.tex
\usepackage{amsmath,amsfonts,bm}

\def\eqref#1{equation~\ref{#1}}
\def\1{\bm{1}}
\newcommand{\train}{\mathcal{D}}

\def\rf{{\textnormal{f}}}

\def\ry{{\textnormal{y}}}

\def\rvx{{\mathbf{x}}}
\def\rvy{{\mathbf{y}}}

\DeclareMathAlphabet{\mathsfit}{\encodingdefault}{\sfdefault}{m}{sl}
\SetMathAlphabet{\mathsfit}{bold}{\encodingdefault}{\sfdefault}{bx}{n}

\def\gA{{\mathcal{A}}}

\def\gL{{\mathcal{L}}}

\def\sR{{\mathbb{R}}}

\def\sX{{\mathbb{X}}}
\def\sY{{\mathbb{Y}}}

%% file: tables/synthetic.tex
\begin{table*}[!tp]\centering
    \caption{Comparison of predictive log-likelihood obtained by different methods over synthetically generated tasks ($6$ Seeds).}
    \label{tabel: 1d-benchmark-synthetic-results}
    \scalebox{0.9}{
        \begin{tabular}{@{}l cc cc cc@{}}
        \toprule
            & \multicolumn{2}{c}{Double Sine}      & \multicolumn{2}{c}{Circle}      & \multicolumn{2}{c}{Lissajous} \\
            \cmidrule[0.2pt]{2-7}  
            & context & target      & context & target      & context & target \\
            \midrule
            CNP         & ${-0.195_{\pm0.009}}$ & ${-0.520_{\pm0.019}}$       & ${-2.086_{\pm0.204}}$ & ${-2.387_{\pm0.216}}$       & ${-2.212_{\pm0.146}}$ & ${-2.962_{\pm0.165}}$ \vspace*{2pt}\\
            CANP        & ${0.436_{\pm0.236}}$ & ${-1.742_{\pm0.222}}$       & ${-0.272_{\pm0.043}}$ & ${-1.685_{\pm0.082}}$       & ${-1.112_{\pm0.499}}$ & ${-2.151_{\pm0.248}}$ \vspace*{2pt}\\
            BNP         & ${0.330_{\pm0.010}}$ & ${0.134_{\pm0.017}}$       & ${0.150_{\pm0.010}}$ & ${0.065_{\pm0.009}}$       & ${-0.314_{\pm0.011}}$ & ${-0.434_{\pm0.010}}$ \vspace*{2pt}\\
            CQNP(ours)  & ${\bm{1.448}_{\pm0.042}}$ & ${\bm{1.244}_{\pm0.049}}$       & ${\bm{2.047}_{\pm0.076}}$ & ${\bm{1.932}_{\pm0.080}}$       & ${\bm{0.798}_{\pm0.020}}$ & ${\bm{0.508}_{\pm0.021}}$ \vspace*{2pt}\\
            ACQNP(ours) & ${\bm{1.582}_{\pm0.108}}$ & ${\bm{1.349}_{\pm0.098}}$       & ${\bm{2.118}_{\pm0.059}}$ & ${\bm{2.028}_{\pm0.057}}$       & ${\bm{0.929}_{\pm0.038}}$ & ${\bm{0.634}_{\pm0.034}}$ \\
        \bottomrule 
    \end{tabular}}
\end{table*}

%% file: tables/speed-flow.tex
\begin{table}[!t]\centering
    \caption{Context and target log-likelihoods from experiments on speed-flow data ($6$ Seeds).}
    \label{tabel: speed-flow-benchmark-results}
    \scalebox{0.9}{
        \begin{tabular}{@{}l cc@{}}
        \toprule
            & \multicolumn{2}{c}{Speed-Flow} \\
            \cmidrule[0.2pt]{2-3}  
            & context & target \\
            \midrule
            CNP         & ${0.845_{\pm0.010}}$ & ${0.719_{\pm0.002}}$ \vspace*{2pt}\\
            CANP        & ${0.887_{\pm0.010}}$ & ${0.741_{\pm0.014}}$  \vspace*{2pt}\\
            BNP         & ${0.879_{\pm0.005}}$ & ${0.720_{\pm0.015}}$   \vspace*{2pt}\\
            CQNP(ours)  & ${\bm{1.518}_{\pm0.013}}$ & ${\bm{1.495}_{\pm0.007}}$  \vspace*{2pt}\\
            ACQNP(ours) & ${\bm{1.544}_{\pm0.001}}$ & ${\bm{1.507}_{\pm0.006}}$  \\
        \bottomrule 
    \end{tabular}}
\end{table}

%% file: tables/image-completion.tex
\begin{table*}[!bp]\centering
    \caption{Context and target log-likelihoods on 2D regression tasks ($6$ Seeds).}
    \label{tabel: 2d-benchmark-image-completion-results}
    \scalebox{0.9}{
    \begin{tabular}{@{}l cc cc cc cc cc@{}}
        \toprule
        & \multicolumn{2}{c}{MNIST}      & \multicolumn{2}{c}{FashionMNIST}      & \multicolumn{2}{c}{SVHN}      & \multicolumn{2}{c}{Omniglot}      & \multicolumn{2}{c}{FreyFace} \\
        \cmidrule[0.2pt]{2-11}
        & context & target      & context & target       & context & target      & context & target      & context & target \\
        \midrule
        CNP         & $\underset{(\pm0.006)}{1.061}$ & $\underset{(\pm0.001)}{0.938}$       & $\underset{(\pm0.005)}{0.963}$ & $\underset{(\pm0.004)}{0.872}$       & $\underset{(\pm0.014)}{3.554}$ & $\underset{(\pm0.013)}{3.388}$       & $\underset{(\pm0.004)}{0.978}$ & $\underset{(\pm0.009)}{0.874}$       & $\underset{(\pm0.083)}{0.970}$ & $\underset{(\pm0.088)}{0.941}$ \vspace*{2pt}\\
        CANP        & $\underset{(\pm0.003)}{1.350}$ & $\underset{(\pm0.006)}{0.913}$       & $\underset{(\pm0.007)}{1.226}$ & $\underset{(\pm0.024)}{0.857}$       & $\underset{(\pm0.002)}{4.112}$ & $\underset{(\pm0.016)}{3.715}$       & $\underset{(\pm0.005)}{1.366}$ & $\underset{(\pm0.004)}{0.974}$       & $\underset{(\pm0.053)}{1.062}$ & $\underset{(\pm0.023)}{1.015}$ \vspace*{2pt}\\
        BNP         & $\underset{(\pm0.013)}{1.128}$ & $\underset{(\pm0.009)}{1.061}$       & $\underset{(\pm0.002)}{1.039}$ & $\underset{(\pm0.001)}{0.971}$       & $\underset{(\pm0.008)}{3.679}$ & $\underset{(\pm0.007)}{3.580}$       & $\underset{(\pm0.006)}{0.983}$ & $\underset{(\pm0.004)}{0.950}$       & $\underset{(\pm0.015)}{1.073}$ & $\underset{(\pm0.014)}{1.052}$ \vspace*{2pt}\\
        CQNP(ours)  & $\underset{(\pm0.006)}{\bm{2.683}}$ & $\underset{(\pm0.006)}{\bm{2.609}}$       & $\underset{(\pm0.004)}{\bm{2.012}}$ & $\underset{(\pm0.002)}{\bm{1.932}}$       & $\underset{(\pm0.033)}{\bm{4.725}}$ & $\underset{(\pm0.029)}{\bm{4.447}}$       & $\underset{(\pm0.034)}{\bm{2.529}}$ & $\underset{(\pm0.038)}{\bm{2.471}}$       & $\underset{(\pm0.062)}{\bm{1.433}}$ & $\underset{(\pm0.061)}{\bm{1.392}}$ \vspace*{2pt}\\
        ACQNP(ours) & $\underset{(\pm0.011)}{\bm{2.681}}$ & $\underset{(\pm0.013)}{\bm{2.616}}$       & $\underset{(\pm0.016)}{\bm{2.040}}$ & $\underset{(\pm0.015)}{\bm{1.954}}$       & $\underset{(\pm0.042)}{\bm{4.959}}$ & $\underset{(\pm0.034)}{\bm{4.651}}$       & $\underset{(\pm0.009)}{\bm{2.516}}$ & $\underset{(\pm0.007)}{\bm{2.461}}$       & $\underset{(\pm0.089)}{\bm{1.522}}$ & $\underset{(\pm0.086)}{\bm{1.475}}$ \\
        \bottomrule 
    \end{tabular}}
\end{table*}

%% file: supp.tex
\section{Additional Results}\label{sec: supp-additional-result}
\begin{figure*}[!h]
	\centering
        \scalebox{0.9}{
            \begingroup
            \setlength{\tabcolsep}{-0.5pt}
            \begin{tabular}{c@{\hskip -0.2pt}ccccc}
                \raisebox{4.8\normalbaselineskip}[0pt][0pt]{\rotatebox[origin=c]{90}{Sawtooth}} & \includegraphics[width=0.22\textwidth]{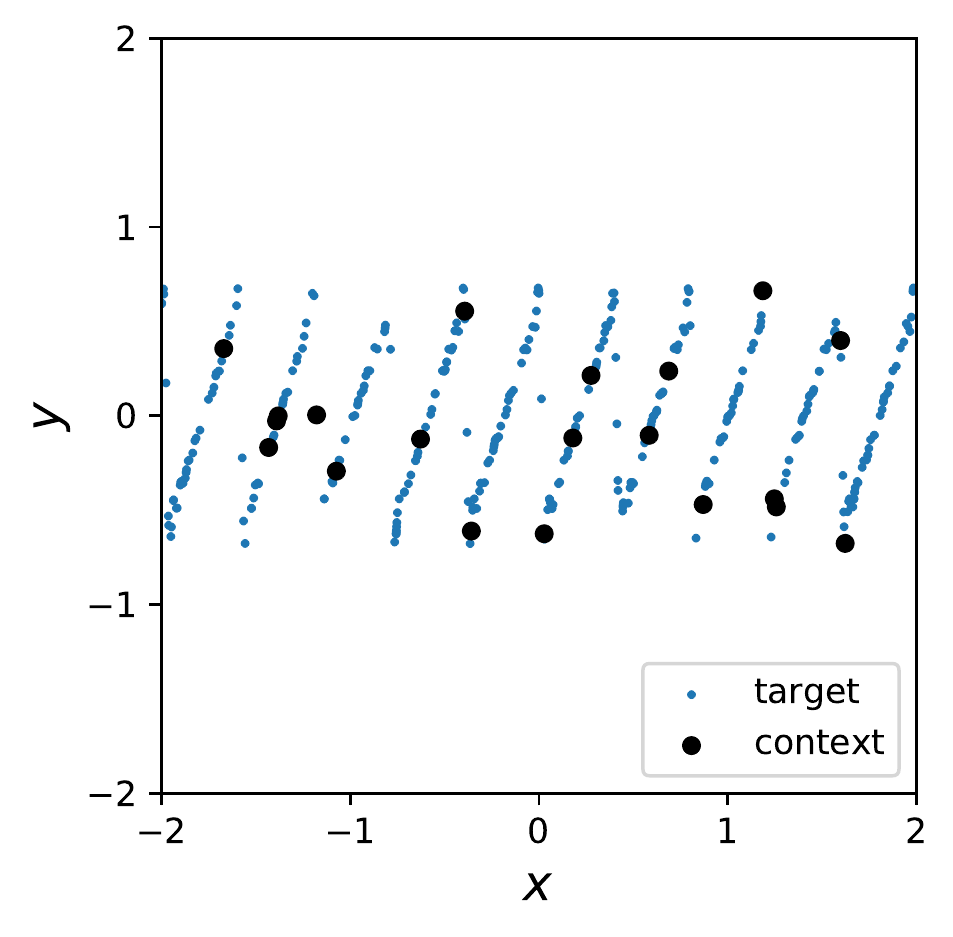} & 
	          \includegraphics[width=0.22\textwidth]{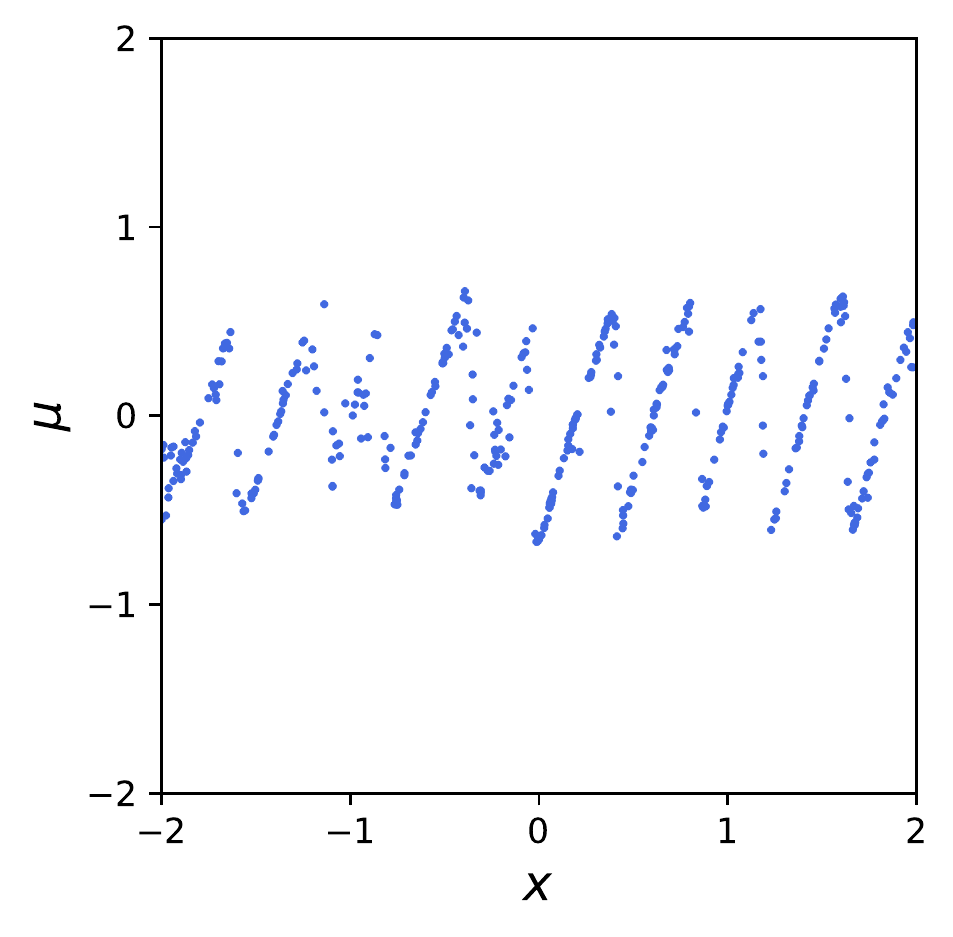} &
                \includegraphics[width=0.22\textwidth]{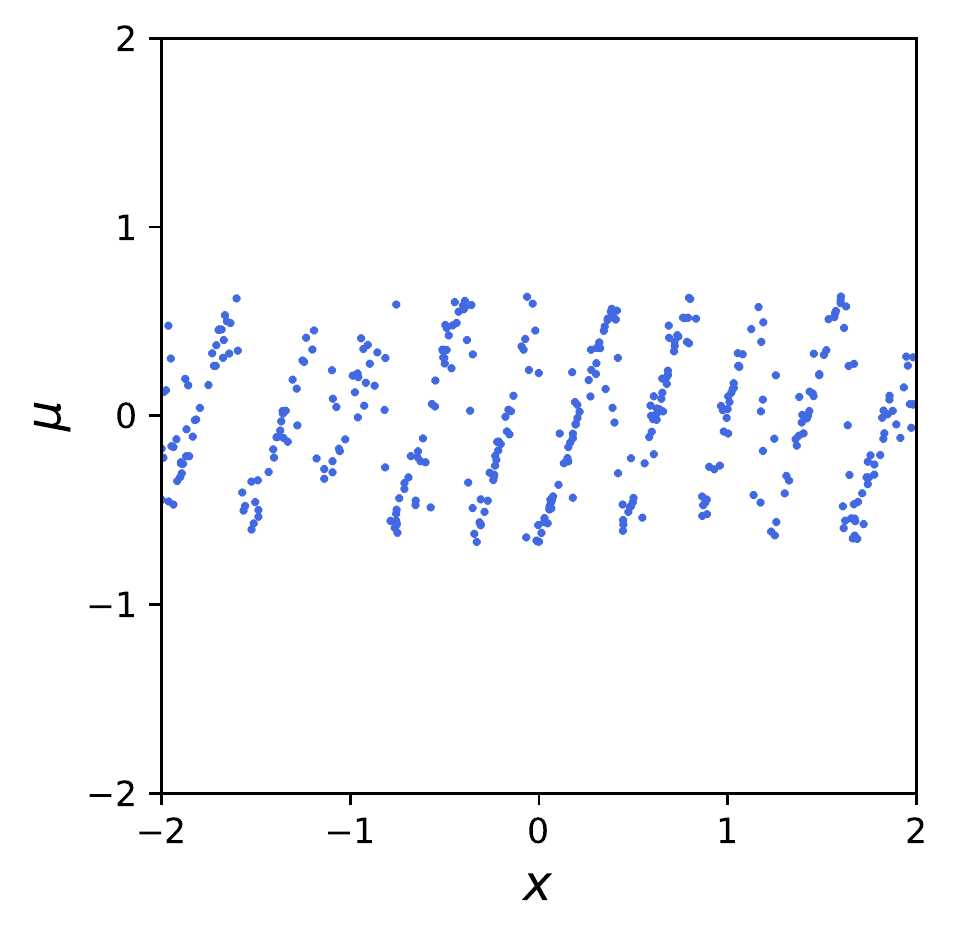} &
                \includegraphics[width=0.22\textwidth]{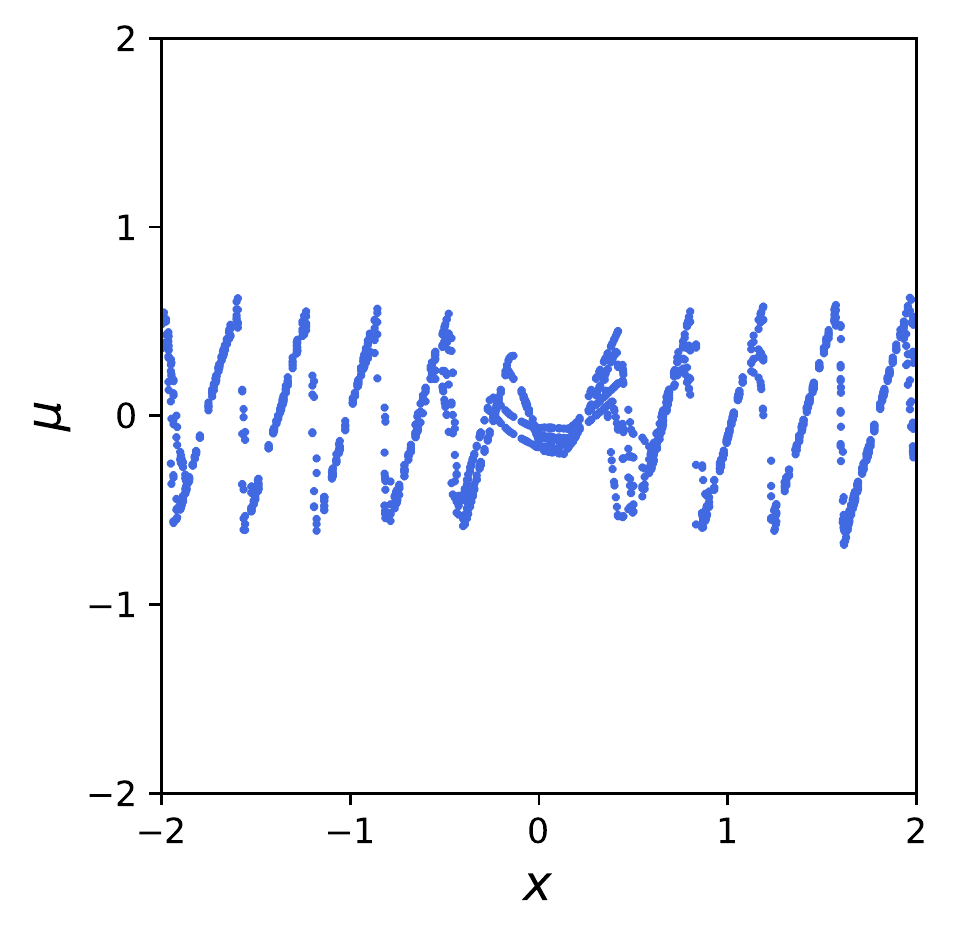} &
                \includegraphics[width=0.22\textwidth]{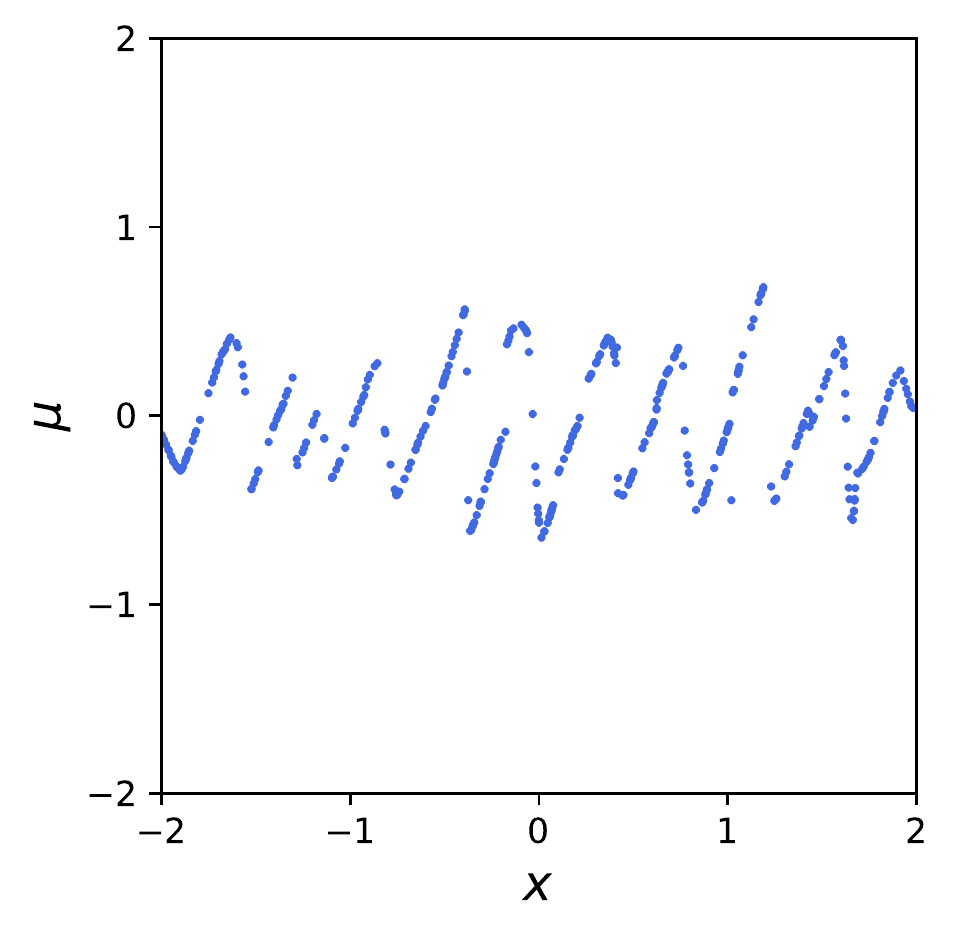} \\
                \raisebox{4.8\normalbaselineskip}[0pt][0pt]{\rotatebox[origin=c]{90}{RBF}} & \includegraphics[width=0.22\textwidth]{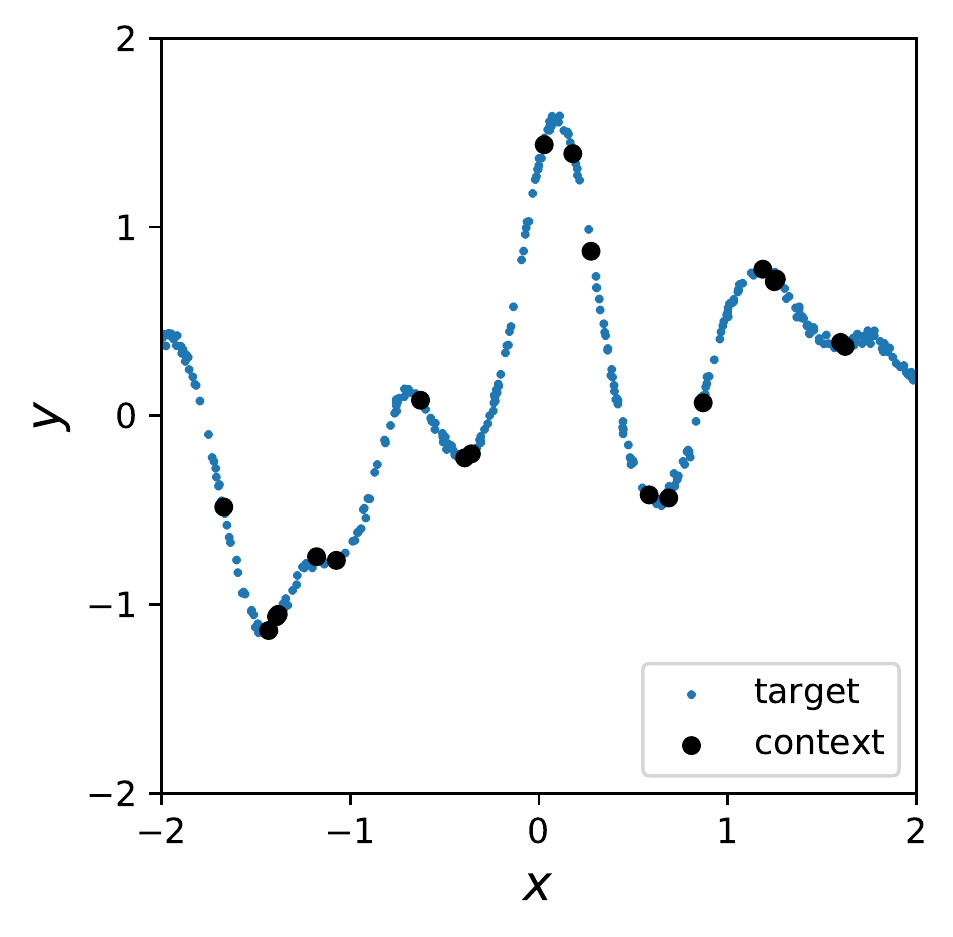} & 
	          \includegraphics[width=0.22\textwidth]{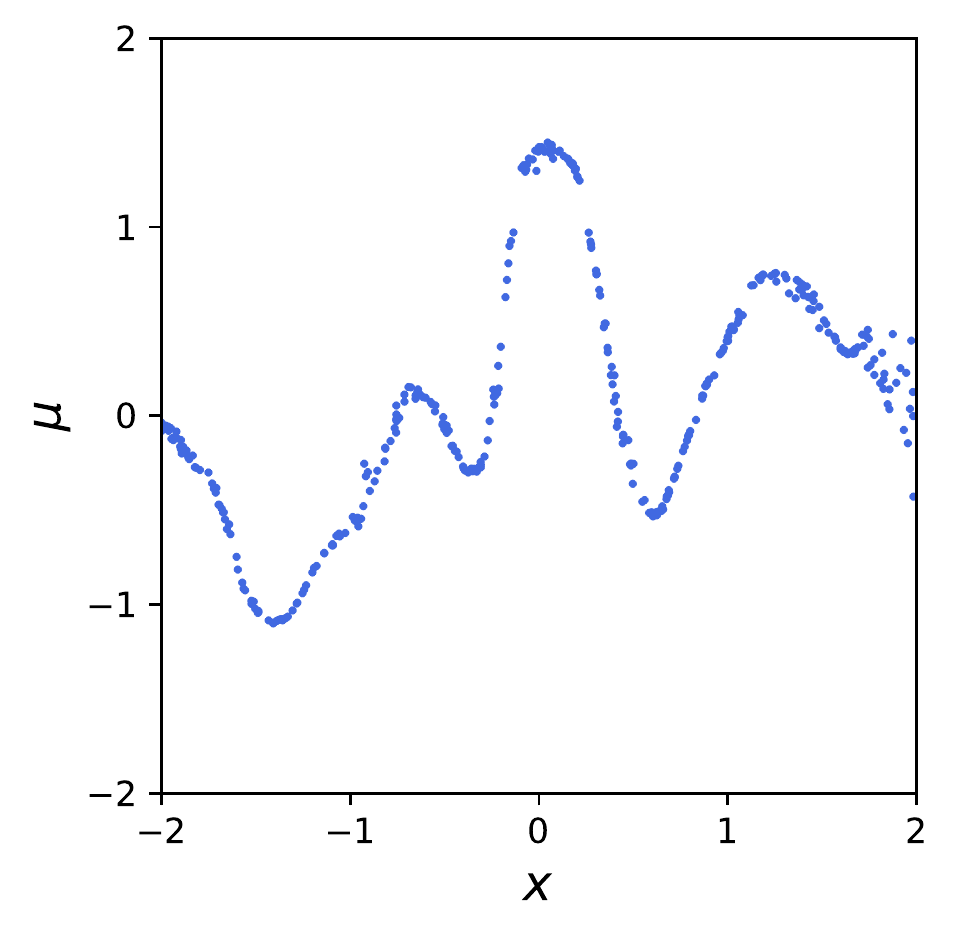} &
                \includegraphics[width=0.22\textwidth]{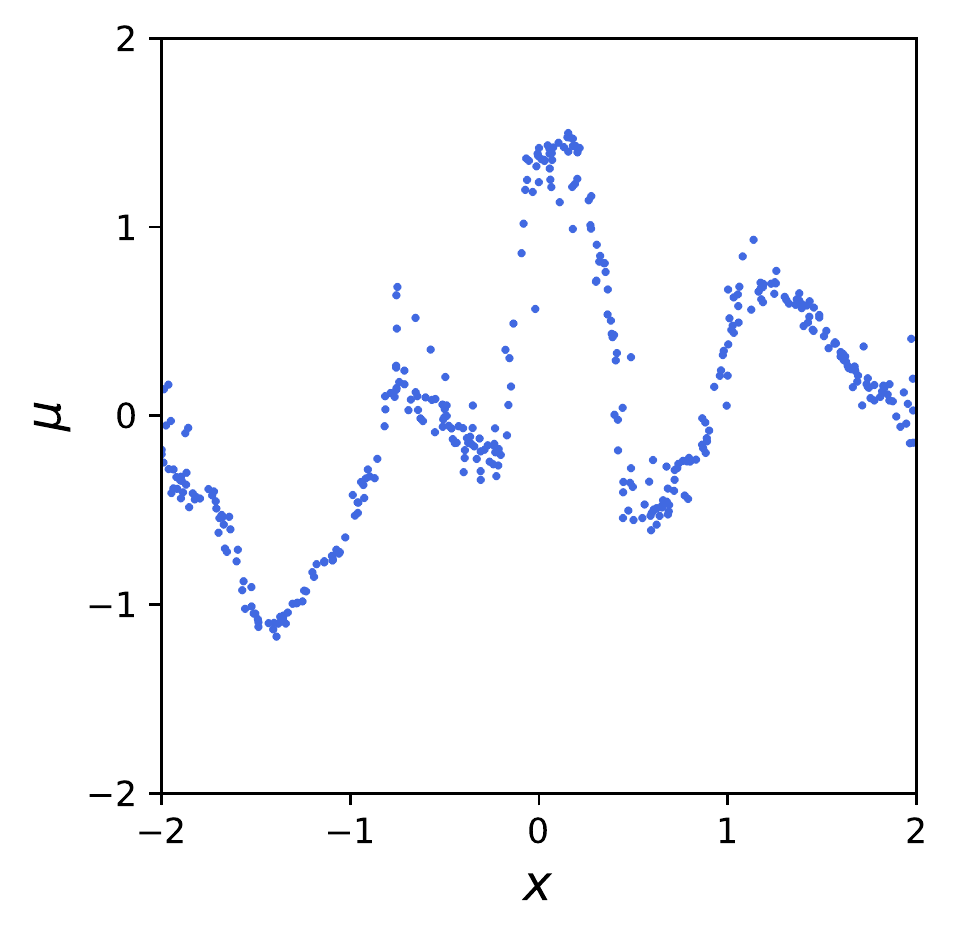} &
                \includegraphics[width=0.22\textwidth]{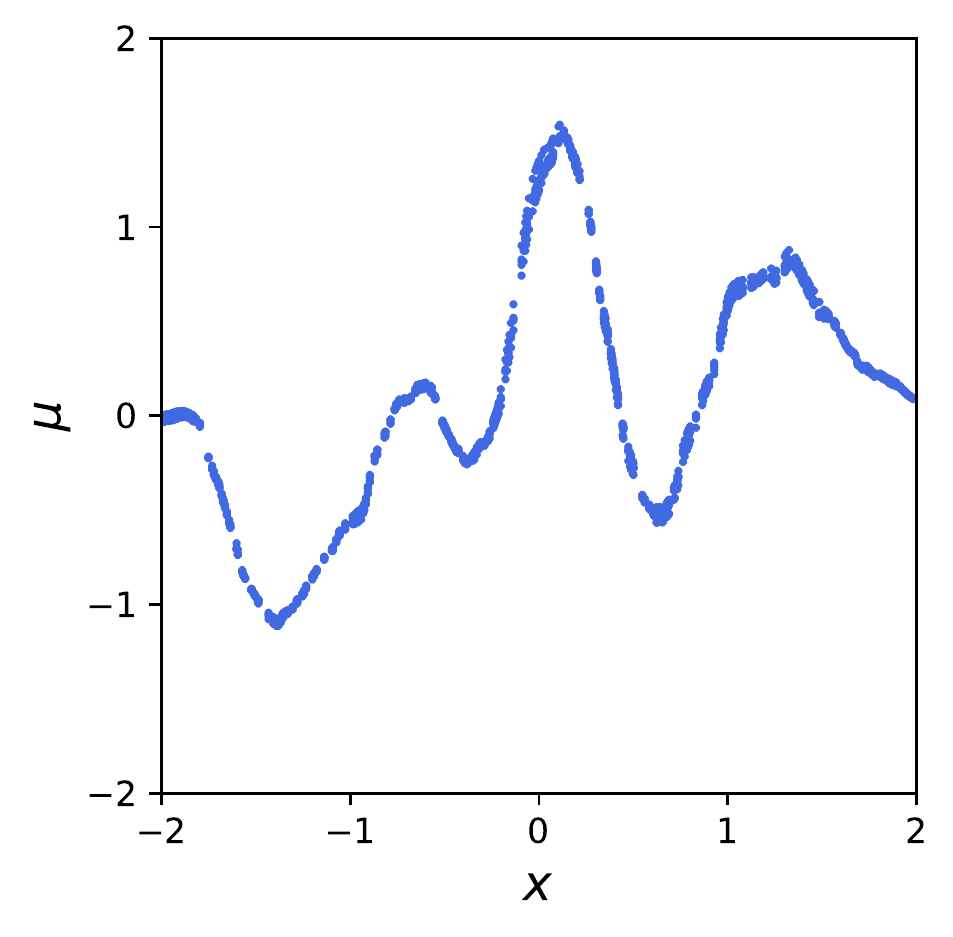} &
                \includegraphics[width=0.22\textwidth]{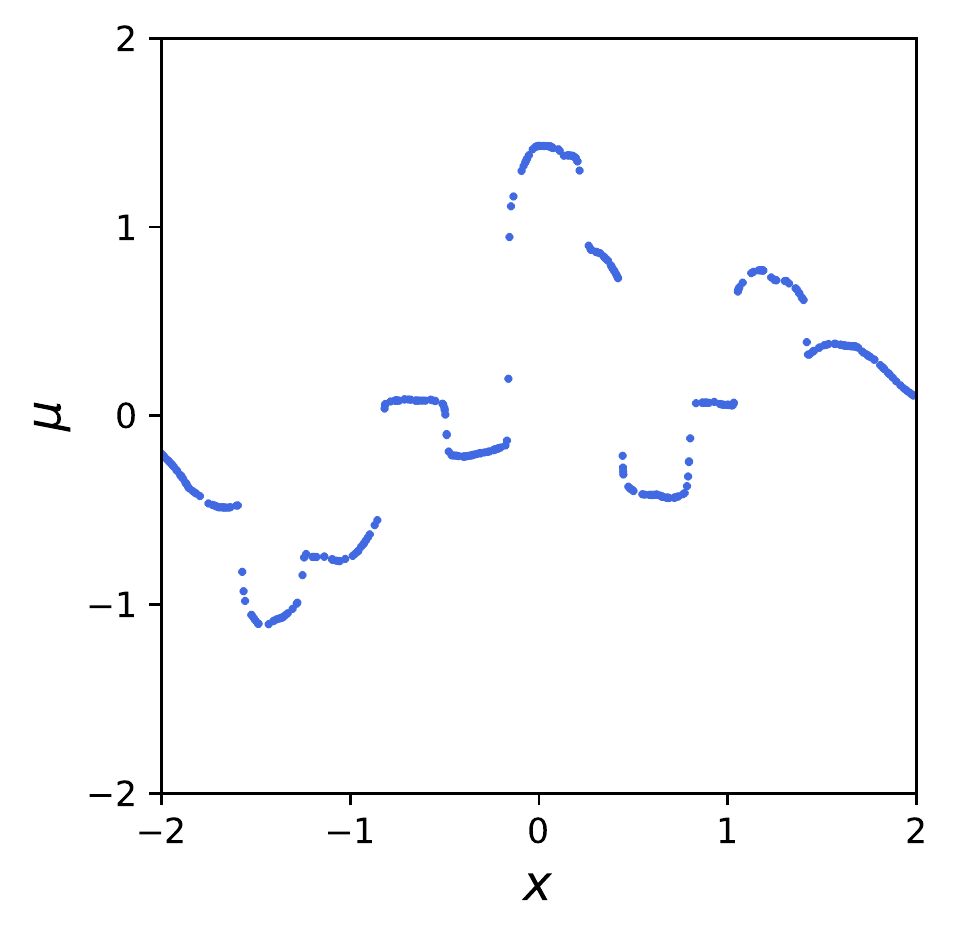} \\
                \raisebox{4.8\normalbaselineskip}[0pt][0pt]{\rotatebox[origin=c]{90}{Mat\'ern 5/2}} & \includegraphics[width=0.22\textwidth]{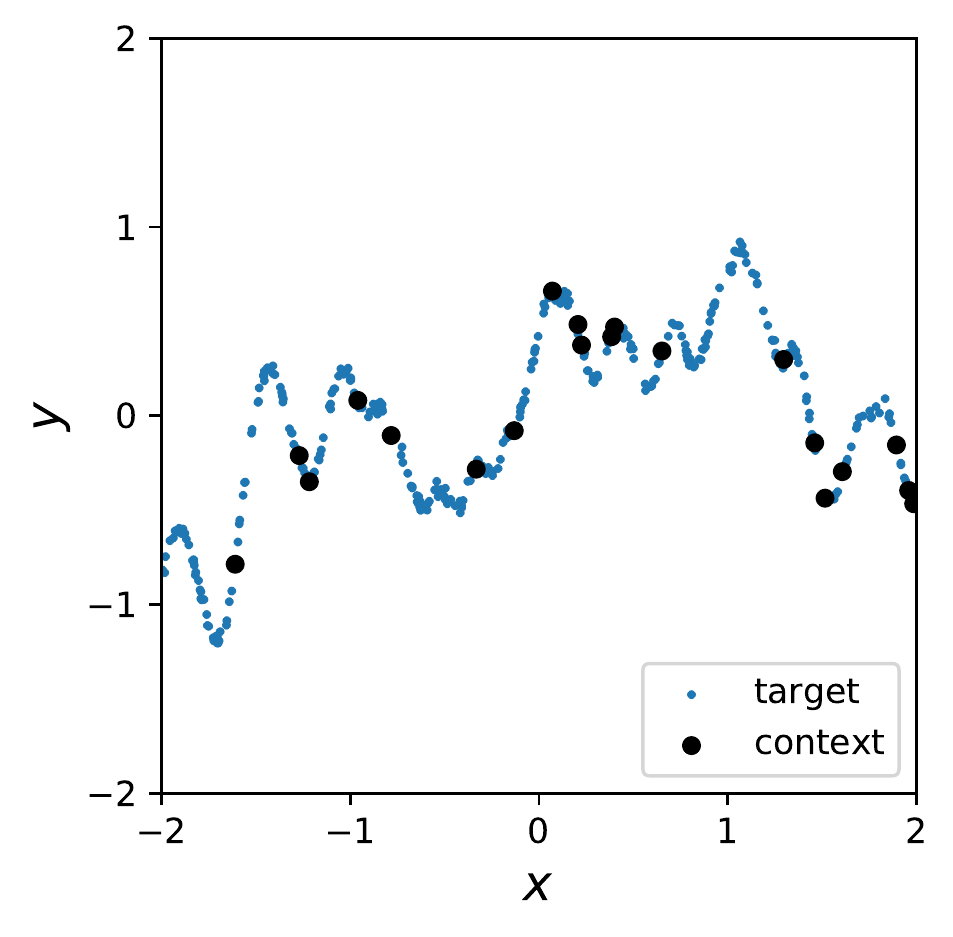} & 
	          \includegraphics[width=0.22\textwidth]{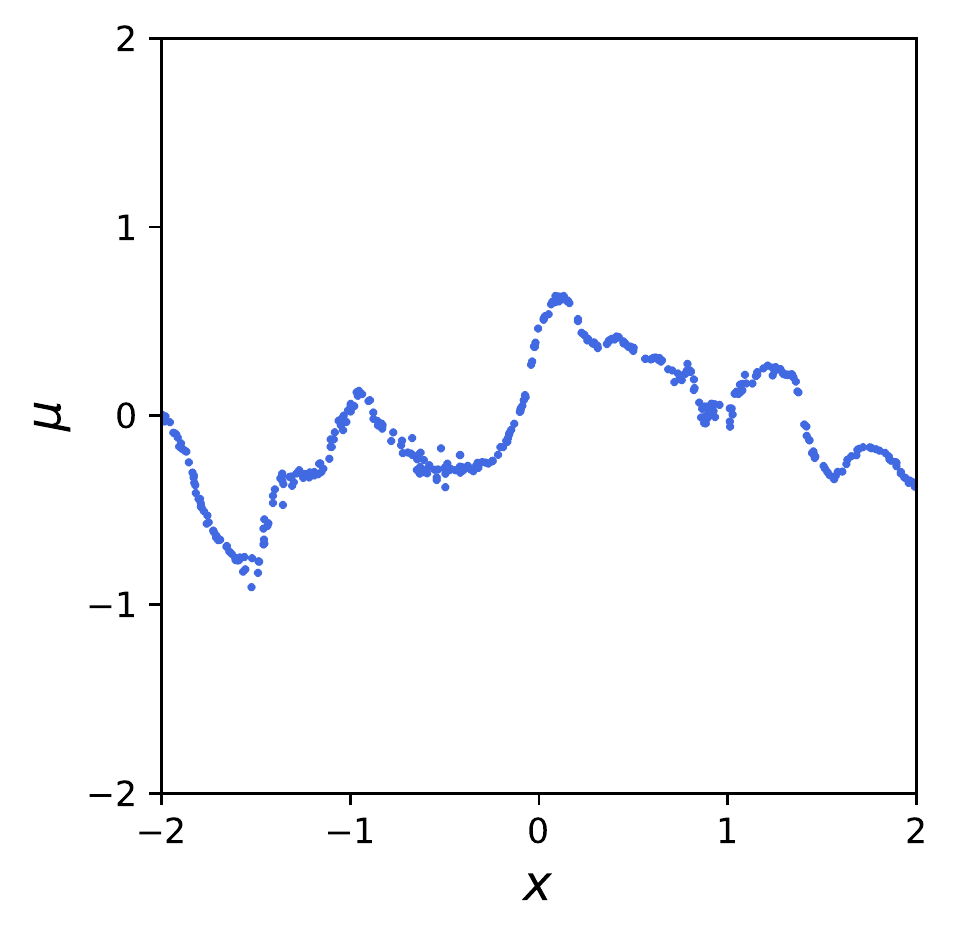} &
                \includegraphics[width=0.22\textwidth]{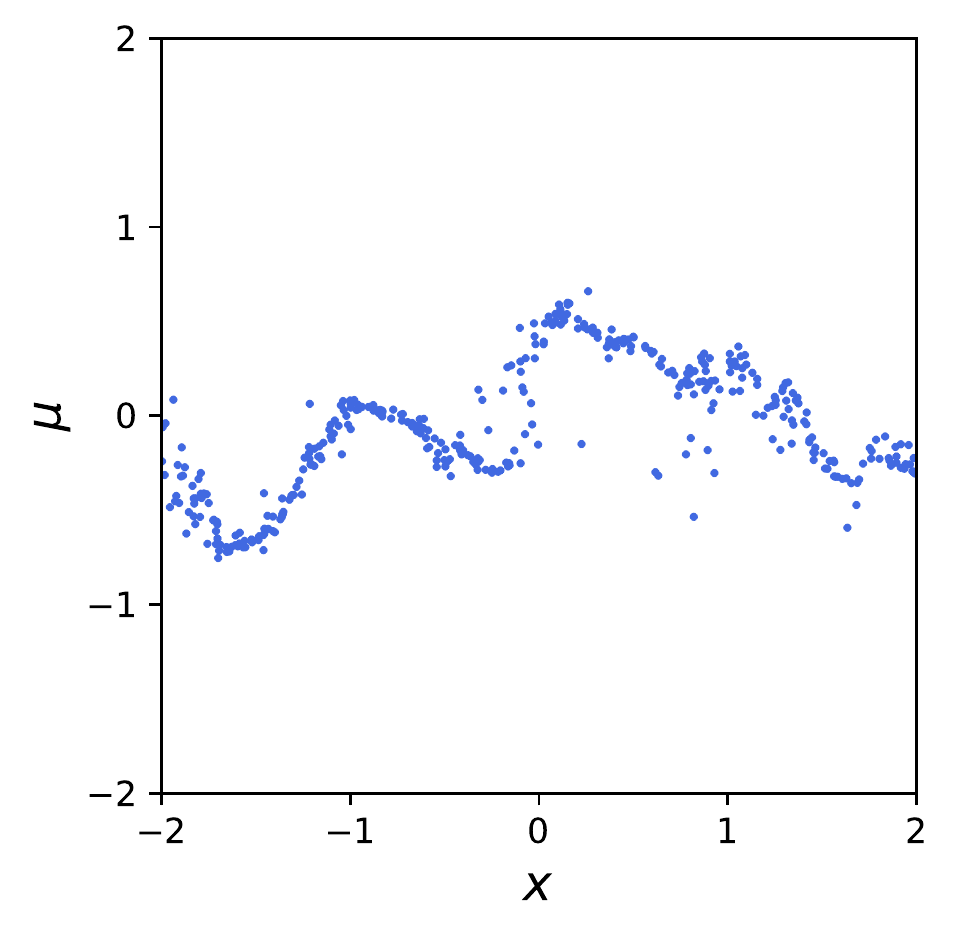} &
                \includegraphics[width=0.22\textwidth]{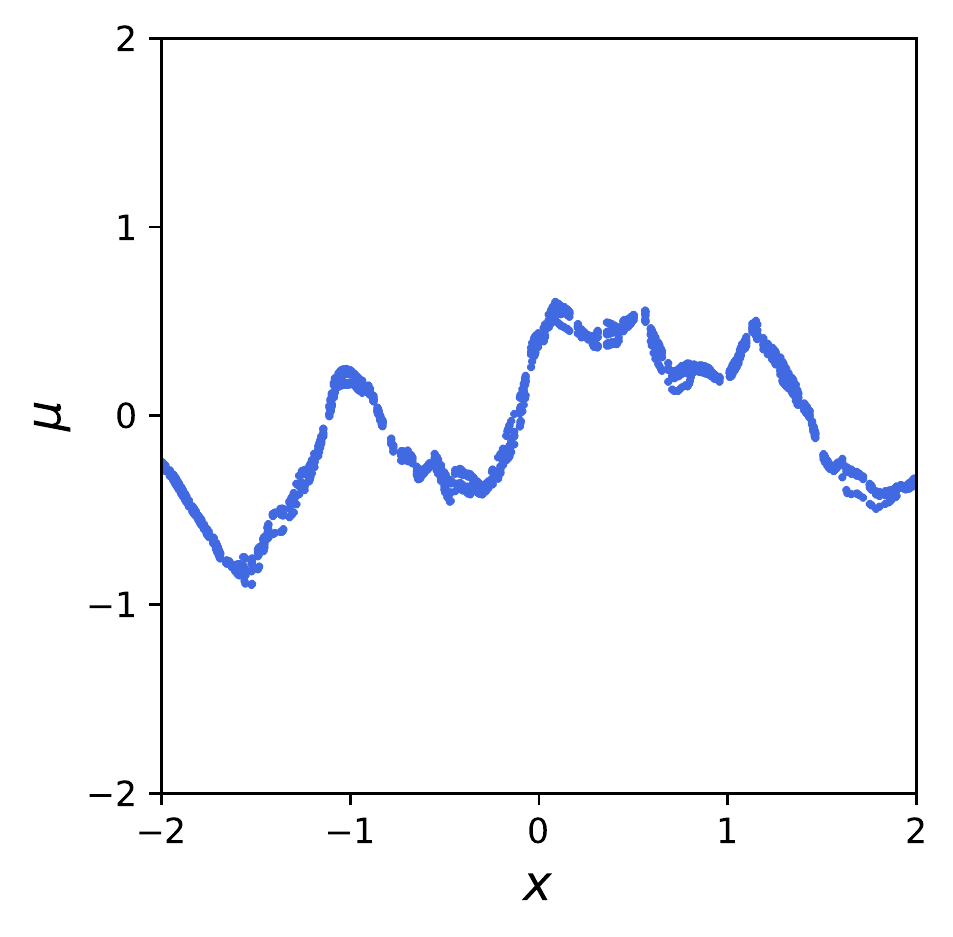} &
                \includegraphics[width=0.22\textwidth]{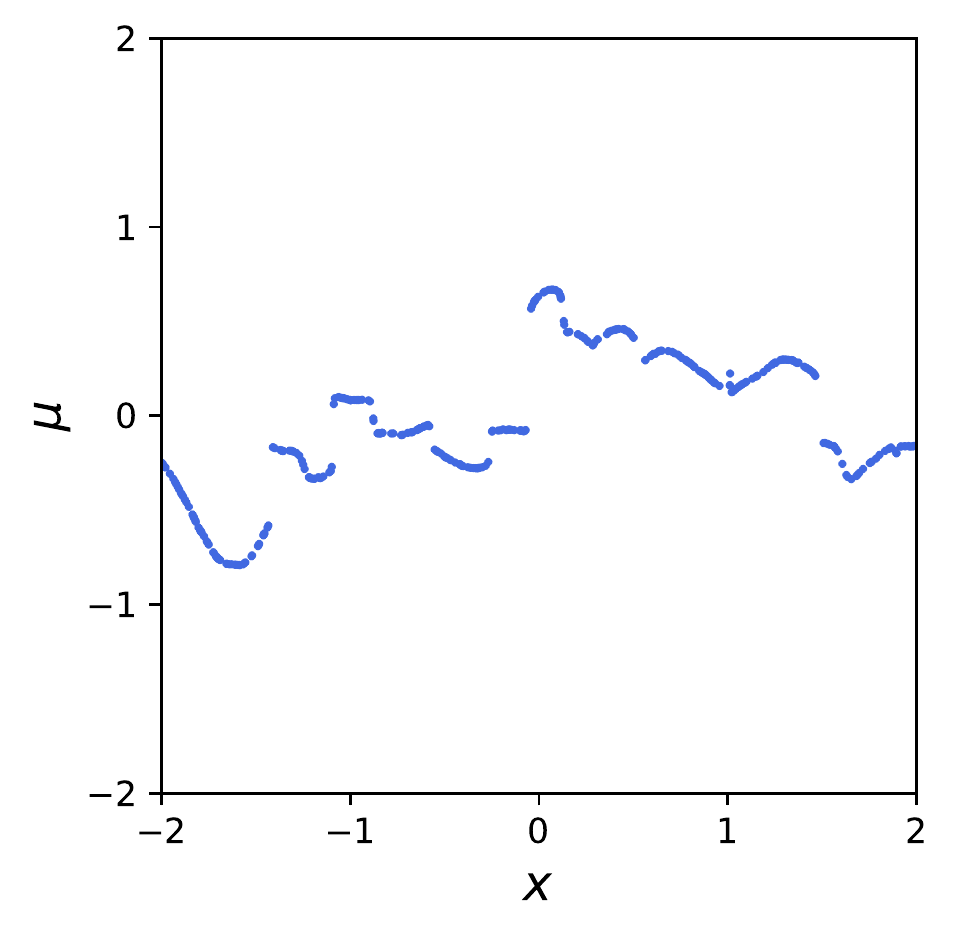} \\
                & a) Data & b) ACQNP & c) CQNP & d) BNP & e) CANP
            \end{tabular}
            \endgroup
        }
	\caption{Examples of predictions made by different methods. For A/CQNP, the mean of the compound predictive distribution, approximated with $N_{\tau}=10$ samples, is plotted. For BNP and CANP, we plot the mean of the Gaussian predictive distribution as the predictions. For BNP, we plot the predictions obtained from 20 different sets of bootstrap contexts.}
        \label{fig: 1d-synthetic-benchmark-supp}
\end{figure*}

We compare A/CQNPs and baselines on data generated from three additional processes described in table \ref{tab: 1d-benchmark-generative-processes-supp} with the following choice of parameters:
\begin{itemize}
    \item Sawtooth\citep{gordon2019convolutional}: $s \sim \mathcal{U}[-2,\,2)$, $\alpha \sim \mathcal{U}[1, 2)$, $\omega \sim [1, 3)$, $\delta \sim \mathcal{U}[-2, 2)$, $K \sim \mathcal{U}[10, 20)$
    \item RBF: $s \sim \mathcal{U}[-2,\,2)$, $\ell = 0.25$, $\sigma=0.75$, $\delta=0.02$
    \item Mat\'ern 5/2: $s \sim \mathcal{U}[-2,\,2)$, $\ell = 0.25$, $\sigma=0.75$, $\delta=0.02$
\end{itemize}
The results are provided in table \ref{tab: 1d-benchmark-synthetic-unimodal-supp}. Figure \ref{fig: 1d-synthetic-benchmark-supp} illustrates examples of predictions made by each method. Note that for A/CQNP, the predictions correspond to the mean of the conditional distribution $p(y\,|\,x)$, not the quantiles. The mean of the uncountable mixture of $\gA L$ distributions can be computed as the following:
\begin{equation}
\begin{split}
    \mathbb{E}_{\ry}[p(\ry \,|\, x)] & = \mathbb{E}_{\ry}\left[\mathbb{E}_{\tau \,\sim\, \mathcal{U}(0,1)}\left[\alpha_{\tau}(x) \, \mathcal{A}L\left(\ry \,|\, \mu_{\tau}(x), \sigma_{\tau}(x), \tau\right)\right]\right] \\
    & = \mathbb{E}_{\tau \,\sim\, \mathcal{U}(0,1)}\left[\mathbb{E}_{\ry}\left[\alpha_{\tau}(x) \, \mathcal{A}L\left(\ry \,|\, \mu_{\tau}(x), \sigma_{\tau}(x), \tau\right)\right]\right] \\
    & = \mathbb{E}_{\tau \,\sim\, \mathcal{U}(0,1)}\left[\alpha_{\tau}(x) \, \mathbb{E}_{\ry}\left[\mathcal{A}L\left(\ry \,|\, \mu_{\tau}(x), \sigma_{\tau}(x), \tau\right)\right]\right] \\
    & = \mathbb{E}_{\tau \,\sim\, \mathcal{U}(0,1)}\left[\alpha_{\tau}(x) \, (\mu_{\tau}(x) + \frac{1-2\tau}{\tau(1-\tau)}\sigma_{\tau}(x))
    \right].
\end{split}
\end{equation}
Similar to section 3.2, we use Monte Carlo to approximate this expectation. For sawtooth, RBF, and Mat\'ern 5/2, we used 100, 50, and 50 maximum context points, respectively.
\begin{table*}[!h]
	% \vspace*{5pt}
	\centering
        \caption{Synthetic processes used in unimodal 1D regression experiments.}
        \label{tab: 1d-benchmark-generative-processes-supp}
        \scalebox{1}{
	{\setlength{\tabcolsep}{1.2pt}
	\begin{tabular}{lc} \toprule
		Process          & $g(s)=(g_x(s),\, g_y(s))$ \\ \midrule
		Sawtooth    & $\quad g_x(s)=s\, , \, g_{y}(s) = \frac{\alpha}{2} - \frac{\alpha}{\pi}\sum_{k=1}^{K}(-1)^k\frac{\sin{(2 \pi k \omega (s+\delta)})}{k}$ \vspace*{5pt} \\ 
		GP (RBF)        & $\quad g_x(s)=s\, , \, g_{y}\sim\mathcal{GP}(0, C)\, , \, C(x, x')=\sigma^2\exp{(-\frac{{\lVert x-x'\rVert}^2}{2\mathcal{\ell}})}+\delta$ \vspace*{5pt} \\
		 GP (Mat\'ern 5/2)   & $\quad g_x(s)=s\, , \, g_{y}\sim\mathcal{GP}(0, C)\, , \, C(x, x')=\sigma^2(1+\frac{\sqrt{5}d}{\ell}+\frac{5d^2}{3\ell^2})+\delta, d = \lVert x-x'\rVert $ \vspace*{1pt}\\ \bottomrule
	\end{tabular}}
 }
\end{table*}

\begin{table*}[h!]\centering
    \caption{Context and target log-likelihoods on synthetic 1D regression tasks ($6$ Seeds).}
    \label{tab: 1d-benchmark-synthetic-unimodal-supp}
    \scalebox{0.9}{
        \begin{tabular}{@{}l cc cc cc@{}}
        \toprule
            & \multicolumn{2}{c}{Sawtooth}      & \multicolumn{2}{c}{RBF}      & \multicolumn{2}{c}{Mat\'ern 5/2} \\
            \cmidrule[0.2pt]{2-7}  
            & context & target      & context & target      & context & target \\
            \midrule
            CNP         & ${0.937}_{\pm0.023}$ & ${0.586}_{\pm0.038}$       & ${0.837}_{\pm0.058}$ & ${0.100}_{\pm0.023}$       & ${0.626}_{\pm0.056}$ & ${-0.183}_{\pm0.013}$ \vspace*{2pt}\\
            CANP        & ${1.191}_{\pm0.190}$ & ${0.341}_{\pm0.085}$      & ${\bm{1.269}}_{\pm0.083}$ & ${0.225}_{\pm0.052}$       & ${\bm{1.058}}_{\pm0.382}$ & ${\bm{-0.015}}_{\pm0.198}$ \vspace*{2pt}\\
            BNP         & ${0.884}_{\pm0.038}$ & ${0.769}_{\pm0.039}$       & ${1.121}_{\pm0.008}$ & ${\bm{0.339}}_{\pm0.009}$       & ${0.879}_{\pm0.018}$ & ${\bm{-0.048}}_{\pm0.018}$ \vspace*{2pt}\\
            CQNP(ours)  & ${\bm{1.229}}_{\pm0.031}$ & ${\bm{0.833}}_{\pm0.035}$       & ${0.947}_{\pm0.042}$ & ${0.083}_{\pm0.037}$       & ${0.515}_{\pm0.039}$ & ${-0.373}_{\pm0.041}$ \vspace*{2pt}\\
            ACQNP(ours) & ${\bm{1.386}}_{\pm0.042}$ & ${\bm{1.026}}_{\pm0.039}$      & ${\bm{1.215}}_{\pm0.027}$ & ${\bm{0.254}}_{\pm0.020}$       & ${\bm{0.912}}_{\pm0.042}$ & ${-0.117}_{\pm0.025}$ \\
        \bottomrule 
    \end{tabular}}
\end{table*}

\newpage
\section{Implementation details}\label{sec: supp-imp-detailes}
This section provides a detailed description of different methods' implementation mentioned throughout the paper. All the implementations are based on PyTorch \citep{paszke2019pytorch}. For CNP and CANP, we closely followed the official implementation\footnote{\url{https://github.com/deepmind/neural-processes}}. Note that the implementation of CANP is identical to ANP, but without the latent path shown in figure 2 of \cite{kim2019attentive}. For BNP, however, we borrowed the implementation provided by the authors \footnote{\url{https://github.com/juho-lee/bnp}}, and thus use their terminology in describing the network architecture.
Nonetheless, neural networks used in all models are instances of multi-layer perceptrons (MLPs) with ReLU activations and the only differences are regarding their depth and width. To specify the architecture of MLPs, we use the following notation:
$$[d_{h_0}] \times [d_{h_1}, \dots, d_{h_{n-1}}] \times [d_{h_n}],$$
where $d_{h_0}$ and $d_{h_n}$ denote the dimension of the network's input and output respectively. Furthermore, $[d_{h_1}, \dots, d_{h_{n-1}}]$ shows that the network has $n-1$ hidden layers with $d_{h_i}$ as the width of i-th hidden layer. In all the experiments, Adam \citep{kingma2014adam} is used for optimizing the objective function. Other than the learning rate and the $\ell_2$ regularizer, rest of the hyper-parameters used with Adam are set to the default values in Pytorch. 

\subsection{Synthetic Data}\label{sec: supp-imp-detailes-synth-data}
For synthetic data, each model is trained for $10^5$ iterations with 128 sampled functions per batch. During training, the tasks are generated at the moment, i.e. the training data is not fixed across different models and seeds. However, for evaluation, we generate and save $5\times10^3$ batches, each containing 16 curves. This data is later used to evaluate all the models. Note that in our implementation, $N_{\mathrm{total}}$ is the same across all observations in each batch. More precisely, for a batch $\mathcal{E}=\{\mathcal{E}_k\}_{k=1}^{n_b}$ with $n_b$ as the batch size, all $\mathcal{E}_k$s contain the same number of data points. However, $N_{\mathrm{total}}$ is not necessarily the same between two different batches as explained in section 4.1. The same setup holds for $N_{\mathrm{context}}$.
%----------------------------------------------------------------
\subsubsection{CNP}\label{sec: supp-CNP-imp-synth-data}
Table \ref{tab: supp-CNP-imp-detail-synth-data} shows the encoder and decoder architectures used in the implementation of CNPs for experiments on synthetic data. Table \ref{tab: supp-CNP-hyperparams-detail-synth-data} summarizes the choice of optimization hyperparameters along with the GPU devices used for training and testing.
\input{tables/supplement/CNP/synthetic-data/CNP-architecture-synthetic-data.tex}
\input{tables/supplement/CNP/synthetic-data/CNP-hyperparam-synthetic-data.tex}
%----------------------------------------------------------------
\subsubsection{CANP}\label{sec: supp-CANP-imp-synth-data}
Table \ref{tab: supp-CANP-imp-detail-synth-data} contains details on the MLP architectures used for modeling the encoder and decoder modules in CANPs. Note that instead of passing raw context and target inputs as keys and queries to the attention modules, we first pass them through separate MLPs, namely key encoder and query encoder, and then apply the attention to the obtained embeddings. Here we work with the same 8-headed attention \citep{vaswani2017attention} mechanism used in the official implementation. Table \ref{tab: supp-CANP-hyperparams-detail-synth-data} summarizes the choice of optimization hyperparameters along with the GPU devices used for training and testing.
\input{tables/supplement/CANP/synthetic-data/CANP-architecture-synthetic-data.tex}
\input{tables/supplement/CANP/synthetic-data/CANP-hyperparam-synthetic-data.tex}
%----------------------------------------------------------------
\subsubsection{BNP}\label{sec: supp-BNP-imp-synth-data}
The architecture details for different components of BNPs including the encoder, adaptation layer, and decoder are provided in table \ref{tab: supp-BNP-imp-detail-synth-data}. For all the benchmarks, we use $k=4$ and $k=50$ bootstrap contexts for training and testing, respectively. The choice of optimization hyperparameters along with the GPU devices used for training and testing are included in table \ref{tab: supp-BNP-hyperparams-detail-synth-data}.
\input{tables/supplement/BNP/synthetic-data/BNP-architecture-synthetic-data.tex}
\input{tables/supplement/BNP/synthetic-data/BNP-hyperparam-synthetic-data.tex}
%----------------------------------------------------------------
\subsubsection{CQNP}\label{sec: supp-CQNP-imp-synth-data}
The encoder and decoder architectures used for implementing CQNPs in different benchmarks are shown in table \ref{tab: supp-CQNP-imp-detail-synth-data}. Table \ref{tab: supp-CQNP-hyperparams-detail-synth-data} summarizes the choice of hyperparameters along with the GPU devices used for training and testing.
\input{tables/supplement/CQNP/synthetic-data/CQNP-architecture-synthetic-data.tex}
\input{tables/supplement/CQNP/synthetic-data/CQNP-hyperparam-synthetic-data.tex}
%----------------------------------------------------------------
\subsubsection{ACQNP}\label{sec: supp-ACQNP-imp-synth-data}
Compared to CQNP, ACQNP has an additional component named the adaptation layer which takes in the raw sample $u$ together with context representation and target input and maps them to a new set of quantile levels $\tau$ that we eventually approximate. Note that this is different from the adaptation layer used in BNPs. Also, we apply a sigmoid function to the final outputs of the adaptation layer to make sure that they correspond to valid quantile levels. The depth and width of the MLPs used for parameterizing the encoder, adaptation layer, and decoder in ACQNPs are presented in table \ref{tab: supp-ACQNP-imp-detail-synth-data}. We summarize the choice of hyperparameters along with the GPU models used for training and testing in table \ref{tab: supp-ACQNP-hyperparams-detail-synth-data}.
\input{tables/supplement/ACQNP/synthetic-data/ACQNP-architecture-synthetic-data.tex}
\input{tables/supplement/ACQNP/synthetic-data/ACQNP-hyperparam-synthetic-data.tex}
%----------------------------------------------------------------
%----------------------------------------------------------------
\subsection{Speed-Flow}\label{sec: supp-imp-detailes-speed-flow}
For the speed-flow data, $75\%$ of each lane's observations ($\approx 988$) are randomly selected for training and the rest are held out for testing. The batch size for both training and testing is 2 since we have data from 2 lanes. For the final evaluation of each method, we take the context and target sets to be the training and testing data, respectively. For training, we generate and save $10^4$ copies of the training data with random partitioning to context and target sets. This means that we fix the training curves as well as the evaluation data across all models as the dataset is quite small and does not require a lot of memory for storage.
%----------------------------------------------------------------
\subsubsection{CNP}\label{sec: supp-CNP-imp-speed-flow}
Table \ref{tab: supp-CNP-imp-detail-speed-flow} shows the encoder and decoder architectures used for implementing CNPs in different benchmarks. Table \ref{tab: supp-CNP-hyperparams-detail-speed-flow} summarizes the choice of optimization hyperparameters along with the GPU devices used for training and testing.
\input{tables/supplement/CNP/speed-flow/CNP-architecture-speed-flow.tex}
\input{tables/supplement/CNP/speed-flow/CNP-hyperparam-speed-flow.tex}
%----------------------------------------------------------------
\subsubsection{CANP}\label{sec: supp-CANP-imp-speed-flow}
Table \ref{tab: supp-CANP-imp-detail-speed-flow} contains details on the MLP architectures used for modeling the encoder and decoder modules in CANPs. Note that instead of passing raw context and target inputs as keys and queries to the attention modules, we first pass them through separate MLPs, namely the key encoder and query encoder, and then apply the attention mechanism to the obtained embedding. Here we work with the same 8-headed attention mechanism used in the official implementation. Table \ref{tab: supp-CANP-hyperparams-detail-speed-flow} summarizes the choice of optimization hyperparameters along with the GPU devices used for training and testing.
\input{tables/supplement/CANP/speed-flow/CANP-architecture-speed-flow.tex}
\input{tables/supplement/CANP/speed-flow/CANP-hyperparam-speed-flow.tex}
%----------------------------------------------------------------
\subsubsection{BNP}\label{sec: supp-BNP-imp-speed-flow}
The architecture details for different components of the BNPs including the encoder, adaptation layer, and decoder are provided in table \ref{tab: supp-BNP-imp-detail-speed-flow}. For all the benchmarks, we use $k=4$ and $k=50$ bootstrap contexts for training and testing, respectively. The choice of optimization hyperparameters along with the GPU devices used for training and testing are included in table \ref{tab: supp-BNP-hyperparams-detail-speed-flow}.
\input{tables/supplement/BNP/speed-flow/BNP-architecture-speed-flow.tex}
\input{tables/supplement/BNP/speed-flow/BNP-hyperparam-speed-flow.tex}
%----------------------------------------------------------------
\subsubsection{CQNP}\label{sec: supp-CQNP-imp-speed-flow}
The encoder and decoder architectures used for implementing CQNPs in different benchmarks are shown in table \ref{tab: supp-CQNP-imp-detail-speed-flow}. Table \ref{tab: supp-CQNP-hyperparams-detail-speed-flow} summarizes the choice of hyperparameters along with the GPU devices used for training and testing.
\input{tables/supplement/CQNP/speed-flow/CQNP-architecture-speed-flow.tex}
\input{tables/supplement/CQNP/speed-flow/CQNP-hyperparam-speed-flow.tex}
%----------------------------------------------------------------
\subsubsection{ACQNP}\label{sec: supp-ACQNP-imp-speed-flow}
Compared to CQNP, ACQNP has an additional component named the adaptation layer which takes in the raw sample of $u$ together with context representation and target inputs and maps them to a new set of quantile levels $\tau$ that we eventually approximate. Note that this is different from the adaptation layer used in BNPs. Also, we apply a sigmoid function to the outputs of the adaptation layer to make sure that they correspond to valid quantile levels. The depth and width of the MLPs used for parameterizing the encoder, adaptation layer, and decoder in ACQNPs are presented in table \ref{tab: supp-ACQNP-imp-detail-speed-flow}. We summarize the choice of hyperparameters along with the GPU models used in training and testing in table \ref{tab: supp-ACQNP-hyperparams-detail-speed-flow}.
\input{tables/supplement/ACQNP/speed-flow/ACQNP-architecture-speed-flow.tex}
\input{tables/supplement/ACQNP/speed-flow/ACQNP-hyperparam-speed-flow.tex}
%----------------------------------------------------------------
%----------------------------------------------------------------
\subsection{Image completion}\label{sec: supp-imp-detailes-image-comp}
In image completion experiments on MNIST, Fashion-MNIST, SVHN, and Omniglot, we use the default train/test split of the data. For FreyFace, however, we randomly select $75\%$ of the images for training and keep the rest for testing. Similar to the experiments on synthetic data, the partitioning of image pixels to context and target sets is done randomly and during training, i.e. context and target sets are not fixed across different models and seeds. For evaluation, however, we saved the generated batches from test images. In the case of FreyFace, we repeat this process 4 more times so that each test image has 5 copies with different context/target splits in the stored evaluation batches.
Note that the number of context points across different tasks in a batch is the same, but might change from one batch to another. Obviously, the union of context and target sets which comprises all pixels of an image is the same in all cases as the image size is fixed in each dataset. All the models were trained for 100 epochs with 16 images per batch. The same batch size is used for testing.
%----------------------------------------------------------------
\subsubsection{CNP}\label{sec: supp-CNP-imp-image-comp}
Table \ref{tab: supp-CNP-imp-detail-image-comp} shows the encoder and decoder architectures used for implementing CNPs in different benchmarks. Table \ref{tab: supp-CNP-hyperparams-detail-image-comp} summarizes the choice of optimization hyperparameters along with the GPU devices used for training and testing.
\input{tables/supplement/CNP/image-completion/CNP-architecture-image-completion.tex}
\input{tables/supplement/CNP/image-completion/CNP-hyperparam-image-completion.tex}
%----------------------------------------------------------------
\subsubsection{CANP}\label{sec: supp-CANP-imp-image-comp}
Table \ref{tab: supp-CANP-imp-detail-image-comp} contains details on the MLP architectures used for modeling the encoder and decoder modules in CANPs. Note that instead of passing raw context and target inputs as keys and queries to the attention modules, we first pass them through separate MLPs, namely the key encoder and query encoder, and then apply attention to the obtained embedding. Here we work with the same 8-headed attention mechanism used in the official implementation. Table \ref{tab: supp-CANP-hyperparams-detail-image-comp} summarizes the choice of optimization hyperparameters along with the GPU devices used for training and testing.
\input{tables/supplement/CANP/image-completion/CANP-architecture-image-completion.tex}
\input{tables/supplement/CANP/image-completion/CANP-hyperparam-image-completion.tex}
%----------------------------------------------------------------
\subsubsection{BNP}\label{sec: supp-BNP-imp-image-comp}
The architecture details for different components of the BNPs including the encoder, adaptation layer, and decoder are provided in table \ref{tab: supp-BNP-imp-detail-image-comp}. For all the benchmarks, we use $k=4$ and $k=50$ bootstrap contexts for training and testing, respectively. The choice of optimization hyperparameters along with the GPU devices used for training and testing are included in table \ref{tab: supp-BNP-hyperparams-detail-image-comp}.
\input{tables/supplement/BNP/image-completion/BNP-architecture-image-completion.tex}
\input{tables/supplement/BNP/image-completion/BNP-hyperparam-image-completion.tex}
%----------------------------------------------------------------
\subsubsection{CQNP}\label{sec: supp-CQNP-imp-image-comp}
The encoder and decoder architectures used for implementing CQNPs in different benchmarks are shown in table \ref{tab: supp-CQNP-imp-detail-image-comp}. Table \ref{tab: supp-CQNP-hyperparams-detail-image-comp} summarizes the choice of hyperparameters along with the GPU devices used for training and testing.
\input{tables/supplement/CQNP/image-completion/CQNP-architecture-image-completion.tex}
\input{tables/supplement/CQNP/image-completion/CQNP-hyperparam-image-completion.tex}
%----------------------------------------------------------------
\subsubsection{ACQNP}\label{sec: supp-ACQNP-imp-image-comp}
Compared to CQNP, ACQNP has an additional component named the adaptation layer which takes in the raw sample of $u$ together with context representation and target inputs and maps them to a new set of quantile levels $\tau$ that we eventually approximate. Note that this is different from the adaptation layer used in BNPs. Also, we apply a sigmoid function to the outputs of the adaptation layer to make sure that they correspond to valid quantile levels. The depth and width of the MLPs used for parameterizing the encoder, adaptation layer, and decoder in ACQNPs are presented in table \ref{tab: supp-ACQNP-imp-detail-image-comp}.  
We summarize the choice of hyperparameters along with the GPU models used in training and testing in table \ref{tab: supp-ACQNP-hyperparams-detail-image-comp}.
\input{tables/supplement/ACQNP/image-completion/ACQNP-architecture-image-completion.tex}
\input{tables/supplement/ACQNP/image-completion/ACQNP-hyperparam-image-completion.tex}
%----------------------------------------------------------------

%% file: tables/supplement/CNP/synthetic-data/CNP-architecture-synthetic-data.tex
\begin{table*}[!h]
    \centering
    \caption{Architectural details of CNPs for experiments on synthetic data.} \label{tab: supp-CNP-imp-detail-synth-data}
    \scalebox{0.9}{
    \begin{tabular}{c|c|c}
      \toprule 
      Benchmark   & Encoder   & Decoder\\
      \midrule
        \begin{tabular}{c} Sawtooth\\ RBF\\ Mat\'ern 5/2 \\ Double Sine\\ Circle \\ Lissajous \end{tabular}  & $[2]\times[128, 128, 128]\times[128]$   & $[129]\times[128, 128, 128]\times[2]$ \\
      \bottomrule
    \end{tabular}}
\end{table*}

%% file: tables/supplement/CNP/synthetic-data/CNP-hyperparam-synthetic-data.tex
\begin{table*}[!h]
    \centering
    \caption{Hyper-parameters and GPU devices used for training and testing CNPs on synthetic data.} \label{tab: supp-CNP-hyperparams-detail-synth-data}
    \scalebox{0.9}{
    \begin{tabular}{c|c|c|c|c}
      \toprule 
      Benchmark   & Learning rate   & L2 regularizer  & GPU (Training)   & GPU (Testing)\\
      \midrule 
      Sawtooth      & $5\times10^{-4}$  & $0$   & Quadro RTX 6000  & NVIDIA A100\\
      RBF           & $5\times10^{-4}$  & $0$   & NVIDIA A100  & Quadro RTX 6000\\
      Mat\'ern 5/2  & $5\times10^{-4}$  & $0$   & NVIDIA A100  & Quadro RTX 6000\\
      Double Sine   & $5\times10^{-4}$  & $0$   & NVIDIA A100  & NVIDIA A100\\ 
      Circle        & $5\times10^{-4}$  & $10^{-5}$   & Quadro RTX 6000  & NVIDIA A100\\
      Lissajous     & $5\times10^{-4}$  & $0$   & Quadro RTX 6000  & NVIDIA A100\\
      \bottomrule
    \end{tabular}}
\end{table*}

%% file: tables/supplement/CANP/synthetic-data/CANP-architecture-synthetic-data.tex
\begin{table*}[!h]
    \centering
    \caption{Architectural details of CANPs for experiments on synthetic data.} \label{tab: supp-CANP-imp-detail-synth-data}
    \scalebox{0.9}{
    \begin{tabular}{c|c|c|c|c}
      \toprule 
      Benchmark   & Context Encoder   & Key Encoder   & Query Encoder   & Decoder\\
      \midrule
        \begin{tabular}{c} Sawtooth\\ RBF\\ Mat\'ern 5/2 \\ Double Sine\\ Circle \\ Lissajous \end{tabular}  & $[2]\times[128, 128, 128]\times[128]$   & $[1]\times[128]\times[128]$   & $[1]\times[128]\times[128]$   & $[129]\times[128, 128, 128]\times[2]$ \\
      \bottomrule
    \end{tabular}}
\end{table*}

%% file: tables/supplement/CANP/synthetic-data/CANP-hyperparam-synthetic-data.tex
\begin{table*}[!h]
    \centering
    \caption{Hyper-parameters and GPU devices used for training and testing CANPs on synthetic data.} \label{tab: supp-CANP-hyperparams-detail-synth-data}
    \scalebox{0.9}{
    \begin{tabular}{c|c|c|c|c}
      \toprule 
      Benchmark   & Learning rate   & L2 regularizer  & GPU (Training)   & GPU (Testing)\\
      \midrule 
      Sawtooth      & $10^{-4}$  & $0$   & Quadro RTX 6000  & NVIDIA A100\\ 
      RBF           & $10^{-4}$  & $0$   & Quadro RTX 6000  & Quadro RTX 6000\\
      Mat\'ern 5/2  & $10^{-4}$  & $0$   & Tesla T4  & NVIDIA A100\\
      Double Sine   & $10^{-4}$  & $10^{-5}$   & NVIDIA A100  & NVIDIA A100\\ 
      Circle        & $10^{-4}$  & $10^{-5}$   & Quadro RTX 6000  & NVIDIA A100\\
      Lissajous     & $10^{-4}$  & $10^{-5}$   & Quadro RTX 6000  & NVIDIA A100
      \\
      \bottomrule
    \end{tabular}}
\end{table*}

%% file: tables/supplement/BNP/synthetic-data/BNP-architecture-synthetic-data.tex
\begin{table*}[!h]
    \centering
    \caption{Architectural details of BNPs for experiments on synthetic data.} \label{tab: supp-BNP-imp-detail-synth-data}
    \scalebox{0.9}{
    \begin{tabular}{c|c|c|c|c|c|c}
      \toprule 
      Benchmark   & $d_x$   & $d_y$   & $d_h$   & $l_{pre}$   & $l_{post}$   & $l_{dec}$\\
      \midrule
        \begin{tabular}{c} Sawtooth\\ RBF\\ Mat\'ern 5/2 \\ Double Sine\\ Circle \\ Lissajous \end{tabular}  & $1$   & $1$   & $128$   & $5$   & $3$   & $5$ \\
      \bottomrule
    \end{tabular}}
\end{table*}

%% file: tables/supplement/BNP/synthetic-data/BNP-hyperparam-synthetic-data.tex
\begin{table*}[!h]
    \centering
    \caption{Hyper-parameters and GPU devices used for training and testing BNPs on synthetic data.} \label{tab: supp-BNP-hyperparams-detail-synth-data}
    \scalebox{0.9}{
    \begin{tabular}{c|c|c|c|c|c}
      \toprule 
      Benchmark   & Learning rate   & L2 regularizer  & Scheduler  & GPU (Training)   & GPU (Testing)\\
      \midrule
      Sawtooth      & $5\times10^{-4}$  & $0$   & cosine annealing   & Quadro RTX 6000  & Quadro RTX 6000\\ 
      RBF           & $5\times10^{-4}$  & $0$   & cosine annealing   & Quadro RTX 6000  & Quadro RTX 6000\\ 
      Mat\'ern 5/2  & $5\times10^{-4}$  & $0$   & cosine annealing   & Tesla T4  & NVIDIA A100\\
      Double Sine   & $5\times10^{-4}$  & $0$   & cosine annealing   & Quadro RTX 6000  & Quadro RTX 6000\\ 
      Circle        & $5\times10^{-4}$  & $0$   & cosine annealing  & Quadro RTX 6000  & Quadro RTX 6000\\
      Lissajous     & $5\times10^{-4}$  & $0$   & cosine annealing  & Quadro RTX 6000  & Quadro RTX 6000\\
      \bottomrule
    \end{tabular}}
\end{table*}

%% file: tables/supplement/CQNP/synthetic-data/CQNP-architecture-synthetic-data.tex
\begin{table*}[!h]
    \centering
    \caption{Architectural details of CQNPs for experiments on synthetic data.} \label{tab: supp-CQNP-imp-detail-synth-data}
    \scalebox{0.9}{
    \begin{tabular}{c|c|c}
      \toprule 
      Benchmark   & Context Encoder   & Decoder\\
      \midrule
        \begin{tabular}{c} Sawtooth\\ RBF\\ Mat\'ern 5/2 \\ Double Sine\\ Circle \\ Lissajous \end{tabular}  & $[2]\times[128, 128, 128]\times[128]$   & $[130]\times[128, 128, 128]\times[3]$ \\
      \bottomrule
    \end{tabular}}
\end{table*}

%% file: tables/supplement/CQNP/synthetic-data/CQNP-hyperparam-synthetic-data.tex
\begin{table*}[!h]
    \centering
    \caption{Hyper-parameters and GPU devices used for training and testing CQNPs on synthetic data.} \label{tab: supp-CQNP-hyperparams-detail-synth-data}
    \scalebox{0.9}{
    \begin{tabular}{c|c|c|c|c|c|c}
      \toprule 
      Benchmark   & Learning rate   & L2 regularizer   & $N_{\tau}$(Training)   & $N_{\tau}$(Testing)  & GPU (Training)   & GPU (Testing)\\
      \midrule
      Sawtooth      & $5\times10^{-4}$  & $10^{-5}$  & 50  & 100   & Quadro RTX 6000  & NVIDIA A100\\ 
      RBF           & $10^{-3}$  & $10^{-5}$  & 50  & 100   & Quadro RTX 6000  & Quadro RTX 6000\\ 
      Mat\'ern 5/2  & $5\times10^{-3}$  & $10^{-5}$  & 50  & 100   & Tesla T4  & Tesla T4\\
      Double Sine   & $10^{-3}$  & $10^{-5}$  & 50  & 100   & NVIDIA A100  & Quadro RTX 6000\\ 
      Circle        & $10^{-3}$  & $0$  & 50  & 100   & NVIDIA A100  & NVIDIA A100\\
      Lissajous     & $10^{-3}$  & $10^{-5}$  & 50  & 100   & NVIDIA A100  & NVIDIA A100
      \\
      \bottomrule
    \end{tabular}}
\end{table*}

%% file: tables/supplement/ACQNP/synthetic-data/ACQNP-architecture-synthetic-data.tex
\begin{table*}[!h]
    \centering
    \caption{Architectural details of ACQNPs for experiments on synthetic data.} \label{tab: supp-ACQNP-imp-detail-synth-data}
    \scalebox{0.9}{
    \begin{tabular}{c|c|c|c}
      \toprule 
      Benchmark   & Context Encoder   & Adaptor    & Decoder\\
      \midrule
      Sawtooth    & $[2]\times[128, 128, 128]\times[128]$   & $[129]\times[128, 128, 128]\times[1]$   & $[130]\times[128, 128, 128]\times[3]$\\
      \midrule
      RBF         & $[2]\times[128, 128, 128]\times[128]$   & $[129]\times[128, 128, 128, 128, 128]\times[1]$   & $[130]\times[128, 128, 128]\times[3]$\\
      \midrule
      Mat\'ern 5/2 & $[2]\times[128, 128, 128]\times[128]$   & $[129]\times[128, 128, 128, 128]\times[1]$   & $[130]\times[128, 128, 128]\times[3]$\\
      \midrule
        \begin{tabular}{c} Double Sine\\ Circle \\ Lissajous \end{tabular}  & $[2]\times[128, 128, 128]\times[128]$   & $[129]\times[128, 128, 128, 128, 128]\times[1]$   & $[130]\times[128, 128, 128]\times[3]$ \\
      \bottomrule
    \end{tabular}}
\end{table*}

%% file: tables/supplement/ACQNP/synthetic-data/ACQNP-hyperparam-synthetic-data.tex
\begin{table*}[!h]
    \centering
    \caption{Hyper-parameters and GPU devices used for training and testing ACQNPs on synthetic data.} \label{tab: supp-ACQNP-hyperparams-detail-synth-data}
    \scalebox{0.9}{
    \begin{tabular}{c|c|c|c|c|c|c}
      \toprule 
      Benchmark   & Learning rate   & L2 regularizer   & $N_{\tau}$(Training)   & $N_{\tau}$(Testing)  & GPU (Training)   & GPU (Testing)\\
      \midrule 
      Sawtooth      & $5\times10^{-4}$  & $0$  & 50  & 100   & Quadro RTX 6000  & NVIDIA A100\\ 
      RBF           & $10^{-3}$  & $0$  & 50  & 100   & NVIDIA A100  & Quadro RTX 6000\\ 
      Mat\'ern 5/2  & $10^{-3}$  & $10^{-5}$  & 50  & 100   & Tesla T4  & NVIDIA A100\\
      Double Sine   & $10^{-3}$  & $10^{-5}$  & 50  & 100   & NVIDIA A100  & NVIDIA A100\\ 
      Circle        & $10^{-3}$  & $10^{-5}$  & 50  & 100   & Quadro RTX 6000  & NVIDIA A100\\
      Lissajous     & $10^{-3}$  & $10^{-5}$  & 50  & 100   & NVIDIA A100  & NVIDIA A100\\
      \bottomrule
    \end{tabular}}
\end{table*}

%% file: tables/supplement/CNP/speed-flow/CNP-architecture-speed-flow.tex
\begin{table*}[!h]
    \centering
    \caption{Architectural details of CNPs for experiments on speed-flow data.} \label{tab: supp-CNP-imp-detail-speed-flow}
    \scalebox{0.9}{
    \begin{tabular}{c|c|c}
      \toprule 
      Benchmark   & Encoder   & Decoder\\
      \midrule
        Speed-Flow   & $[2]\times[64, 64]\times[64]$   & $[65]\times[64, 64]\times[2]$ \\
      \bottomrule
    \end{tabular}}
\end{table*}

%% file: tables/supplement/CNP/speed-flow/CNP-hyperparam-speed-flow.tex
\begin{table*}[!h]
    \centering
    \caption{Hyper-parameters and GPU devices used for training and testing CNPs on speed-flow data.} \label{tab: supp-CNP-hyperparams-detail-speed-flow}
    \scalebox{0.9}{
    \begin{tabular}{c|c|c|c|c}
      \toprule 
      Benchmark   & Learning rate   & L2 regularizer  & GPU (Training)   & GPU (Testing)\\
      \midrule
      Speed-Flow    & $10^{-4}$  & $10^{-5}$   & Tesla T4  & Tesla T4\\
      \bottomrule
    \end{tabular}}
\end{table*}

%% file: tables/supplement/CANP/speed-flow/CANP-architecture-speed-flow.tex
\begin{table*}[!h]
    \centering
    \caption{Architectural details of CANPs for experiments on speed-flow data.} \label{tab: supp-CANP-imp-detail-speed-flow}
    \scalebox{0.9}{
    \begin{tabular}{c|c|c|c|c}
      \toprule 
      Benchmark   & Context Encoder   & Key Encoder   & Query Encoder   & Decoder\\
      \midrule
        Speed-Flow   & $[2]\times[64, 64]\times[64]$   & $[1]\times[64]\times[64]$   & $[1]\times[64]\times[64]$   & $[65]\times[64, 64]\times[2]$ \\
      \bottomrule
    \end{tabular}}
\end{table*}

%% file: tables/supplement/CANP/speed-flow/CANP-hyperparam-speed-flow.tex
\begin{table*}[!h]
    \centering
    \caption{Hyper-parameters and GPU devices used for training and testing CANPs on speed-flow data.} \label{tab: supp-CANP-hyperparams-detail-speed-flow}
    \scalebox{0.9}{
    \begin{tabular}{c|c|c|c|c}
      \toprule 
      Benchmark   & Learning rate   & L2 regularizer  & GPU (Training)   & GPU (Testing)\\
      \midrule
      Speed-Flow    & $10^{-4}$  & $10^{-5}$   & Tesla T4  & Tesla T4
      \\
      \bottomrule
    \end{tabular}}
\end{table*}

%% file: tables/supplement/BNP/speed-flow/BNP-architecture-speed-flow.tex
\begin{table*}[!h]
    \centering
    \caption{Architectural details of BNPs for experiments on speed-flow data.} \label{tab: supp-BNP-imp-detail-speed-flow}
    \scalebox{0.9}{
    \begin{tabular}{c|c|c|c|c|c|c}
      \toprule 
      Benchmark   & $d_x$   & $d_y$   & $d_h$   & $l_{pre}$   & $l_{post}$   & $l_{dec}$\\
      \midrule
        Speed-Flow   & $1$   & $1$   & $64$   & $4$   & $3$   & $4$ \\
      \bottomrule
    \end{tabular}}
\end{table*}

%% file: tables/supplement/BNP/speed-flow/BNP-hyperparam-speed-flow.tex
\begin{table*}[!h]
    \centering
    \caption{Hyper-parameters and GPU devices used for training and testing BNPs on speed-flow data.} \label{tab: supp-BNP-hyperparams-detail-speed-flow}
    \scalebox{0.9}{
    \begin{tabular}{c|c|c|c|c|c}
      \toprule 
      Benchmark   & Learning rate   & L2 regularizer  & Scheduler  & GPU (Training)   & GPU (Testing)\\
      \midrule 
      Speed-Flow    & $5\times10^{-4}$  & $10^{-5}$   & None  & Tesla T4  & Tesla T4
      \\
      \bottomrule
    \end{tabular}}
\end{table*}

%% file: tables/supplement/CQNP/speed-flow/CQNP-architecture-speed-flow.tex
\begin{table*}[!h]
    \centering
    \caption{Architectural details of CQNPs for experiments on speed-flow data.} \label{tab: supp-CQNP-imp-detail-speed-flow}
    \scalebox{0.9}{
    \begin{tabular}{c|c|c}
      \toprule 
      Benchmark   & Context Encoder   & Decoder\\
      \midrule
        Speed-Flow   & $[2]\times[64, 64]\times[64]$   & $[66]\times[64, 64]\times[3]$ \\
      \bottomrule
    \end{tabular}}
\end{table*}

%% file: tables/supplement/CQNP/speed-flow/CQNP-hyperparam-speed-flow.tex
\begin{table*}[!h]
    \centering
    \caption{Hyper-parameters and GPU devices used for training and testing CQNPs on speed-flow data.} \label{tab: supp-CQNP-hyperparams-detail-speed-flow}
    \scalebox{0.9}{
    \begin{tabular}{c|c|c|c|c|c|c}
      \toprule 
      Benchmark   & Learning rate   & L2 regularizer   & $N_{\tau}$(Training)   & $N_{\tau}$(Testing)  & GPU (Training)   & GPU (Testing)\\
      \midrule
      Speed-Flow    & $5\times10^{-3}$  & $10^{-5}$  & 100  & 50   & Tesla T4  & Quadro RTX 6000
      \\
      \bottomrule
    \end{tabular}}
\end{table*}

%% file: tables/supplement/ACQNP/speed-flow/ACQNP-architecture-speed-flow.tex
\begin{table*}[!h]
    \centering
    \caption{Architectural details of ACQNPs for experiments on speed-flow data.} \label{tab: supp-ACQNP-imp-detail-speed-flow}
    \scalebox{0.9}{
    \begin{tabular}{c|c|c|c}
      \toprule 
      Benchmark   & Context Encoder   & Adaptor    & Decoder\\
      \midrule
        Speed-Flow   & $[2]\times[64, 64]\times[64]$   & $[65]\times[64, 64]\times[1]$    & $[66]\times[64, 64]\times[3]$ \\
      \bottomrule
    \end{tabular}}
\end{table*}

%% file: tables/supplement/ACQNP/speed-flow/ACQNP-hyperparam-speed-flow.tex
\begin{table*}[!h]
    \centering
    \caption{Hyper-parameters and GPU devices used for training and testing ACQNPs on speed-flow data.} \label{tab: supp-ACQNP-hyperparams-detail-speed-flow}
    \scalebox{0.9}{
    \begin{tabular}{c|c|c|c|c|c|c}
      \toprule 
      Benchmark   & Learning rate   & L2 regularizer   & $N_{\tau}$(Training)   & $N_{\tau}$(Testing)  & GPU (Training)   & GPU (Testing)\\
      \midrule
      Speed-Flow    & $5\times10^{-3}$  & $10^{-5}$  & 100  & 50   & Tesla T4  & Quadro RTX 6000\\
      \bottomrule
    \end{tabular}}
\end{table*}

%% file: tables/supplement/CNP/image-completion/CNP-architecture-image-completion.tex
\begin{table*}[!h]
    \centering
    \caption{Architectural details of CNPs for image completion tasks.} \label{tab: supp-CNP-imp-detail-image-comp}
    \scalebox{0.9}{
    \begin{tabular}{c|c|c}
      \toprule 
      Benchmark   & Encoder   & Decoder\\
      \midrule
        \begin{tabular}{c} MNIST\\ Fashion MNIST \\ Omniglot \\ FreyFace  \end{tabular}   & $[3]\times[128, 128, 128]\times[128]$   & $[130]\times[128, 128, 128]\times[2]$ \\
      \midrule
        SVHN   & $[5]\times[128, 128, 128]\times[128]$   & $[130]\times[128, 128, 128]\times[6]$ \\
      \bottomrule
    \end{tabular}}
\end{table*}

%% file: tables/supplement/CNP/image-completion/CNP-hyperparam-image-completion.tex
\begin{table*}[!h]
    \centering
    \caption{Hyper-parameters and GPU devices used for training and testing CNPs on image completion tasks.} \label{tab: supp-CNP-hyperparams-detail-image-comp}
    \scalebox{0.9}{
    \begin{tabular}{c|c|c|c|c}
      \toprule 
      Benchmark   & Learning rate   & L2 regularizer  & GPU (Training)   & GPU (Testing)\\
      \midrule
      MNIST         & $5\times10^{-4}$  & $0$   & Quadro RTX 6000  & Quadro RTX 6000\\
      Fashion MNIST & $5\times10^{-4}$  & $0$   & Tesla T4  & Quadro RTX 6000\\
      Omniglot      & $5\times10^{-4}$  & $0$   & Quadro RTX 6000  & Quadro RTX 6000\\
      FreyFace      & $5\times10^{-4}$  & $0$   & Quadro RTX 6000  & NVIDIA A100\\
      SVHN          & $5\times10^{-4}$  & $0$   & Tesla T4  & Quadro RTX 6000
      \\
      \bottomrule
    \end{tabular}}
\end{table*}

%% file: tables/supplement/CANP/image-completion/CANP-architecture-image-completion.tex
\begin{table*}[!h]
    \centering
    \caption{Architectural details of CANPs for image completion tasks.} \label{tab: supp-CANP-imp-detail-image-comp}
    \scalebox{0.9}{
    \begin{tabular}{c|c|c|c|c}
      \toprule 
      Benchmark   & Context Encoder   & Key Encoder   & Query Encoder   & Decoder\\
      \midrule
        \begin{tabular}{c} MNIST\\ Fashion MNIST \\ Omniglot \\ FreyFace  \end{tabular}   & $[3]\times[128, 128, 128]\times[128]$   & $[1]\times[128]\times[128]$   & $[1]\times[128]\times[128]$   & $[130]\times[128, 128, 128]\times[2]$ \\
      \midrule
        SVHN   & $[5]\times[128, 128, 128]\times[128]$   & $[2]\times[128]\times[128]$   & $[2]\times[128]\times[128]$   & $[130]\times[128, 128, 128]\times[6]$ \\
      \bottomrule
    \end{tabular}}
\end{table*}

%% file: tables/supplement/CANP/image-completion/CANP-hyperparam-image-completion.tex
\begin{table*}[!h]
    \centering
    \caption{Hyper-parameters and GPU devices used for training and testing CANPs on image completion tasks.} \label{tab: supp-CANP-hyperparams-detail-image-comp}
    \scalebox{0.9}{
    \begin{tabular}{c|c|c|c|c}
      \toprule 
      Benchmark   & Learning rate   & L2 regularizer  & GPU (Training)   & GPU (Testing)\\
      \midrule
      MNIST         & $5\times10^{-4}$  & $0$   & Quadro RTX 6000  & Quadro RTX 6000\\
      Fashion MNIST & $5\times10^{-4}$  & $0$   & Quadro RTX 6000  & Quadro RTX 6000\\
      Omniglot      & $5\times10^{-4}$  & $0$   & Quadro RTX 6000  & Quadro RTX 6000\\
      FreyFace      & $5\times10^{-4}$  & $10^{-5}$   & Quadro RTX 6000  & NVIDIA A100\\
      SVHN          & $5\times10^{-4}$  & $0$   & Quadro RTX 6000  & NVIDIA A100
      \\
      \bottomrule
    \end{tabular}}
\end{table*}

%% file: tables/supplement/BNP/image-completion/BNP-architecture-image-completion.tex
\begin{table*}[!h]
    \centering
    \caption{Architectural details of BNPs for image completion tasks.} \label{tab: supp-BNP-imp-detail-image-comp}
    \scalebox{0.9}{
    \begin{tabular}{c|c|c|c|c|c|c}
      \toprule 
      Benchmark   & $d_x$   & $d_y$   & $d_h$   & $l_{pre}$   & $l_{post}$   & $l_{dec}$\\
      \midrule
        \begin{tabular}{c} MNIST\\ Fashion MNIST \\ Omniglot \\ FreyFace  \end{tabular}   & $2$   & $1$   & $128$   & $5$   & $3$   & $5$ \\
      \midrule
        SVHN   & $2$   & $3$   & $128$   & $5$   & $3$   & $5$ \\
      \bottomrule
    \end{tabular}}
\end{table*}

%% file: tables/supplement/BNP/image-completion/BNP-hyperparam-image-completion.tex
\begin{table*}[!h]
    \centering
    \caption{Hyper-parameters and GPU devices used for training and testing BNPs on image completion tasks.} \label{tab: supp-BNP-hyperparams-detail-image-comp}
    \scalebox{0.9}{
    \begin{tabular}{c|c|c|c|c|c}
      \toprule 
      Benchmark   & Learning rate   & L2 regularizer  & Scheduler  & GPU (Training)   & GPU (Testing)\\
      \midrule
      MNIST         & $5\times10^{-4}$  & $0$   & cosine annealing  & NVIDIA A100  & Quadro RTX 6000 \\
      Fashion MNIST & $5\times10^{-4}$  & $0$   & cosine annealing  & NVIDIA A100  & Quadro RTX 6000\\
      Omniglot      & $5\times10^{-4}$  & $0$   & cosine annealing  & Tesla T4  & Quadro RTX 6000\\
      FreyFace      & $5\times10^{-4}$  & $0$   & cosine annealing  & Quadro RTX 6000  & Quadro RTX 6000\\
      SVHN          & $5\times10^{-4}$  & $0$   & cosine annealing  & NVIDIA A100  & Quadro RTX 6000
      \\
      \bottomrule
    \end{tabular}}
\end{table*}

%% file: tables/supplement/CQNP/image-completion/CQNP-architecture-image-completion.tex
\begin{table*}[!h]
    \centering
    \caption{Architectural details of CQNPs for image completion tasks.} \label{tab: supp-CQNP-imp-detail-image-comp}
    \scalebox{0.9}{
    \begin{tabular}{c|c|c}
      \toprule 
      Benchmark   & Context Encoder   & Decoder\\
      \midrule
        \begin{tabular}{c} MNIST\\ Fashion MNIST \\ Omniglot \\ FreyFace  \end{tabular}   & $[3]\times[128, 128, 128]\times[128]$   & $[131]\times[128, 128, 128]\times[3]$ \\
      \midrule
        SVHN   & $[5]\times[128, 128, 128]\times[128]$   & $[131]\times[128, 128, 128]\times[9]$ \\
      \bottomrule
    \end{tabular}}
\end{table*}

%% file: tables/supplement/CQNP/image-completion/CQNP-hyperparam-image-completion.tex
\begin{table*}[!h]
    \centering
    \caption{Hyper-parameters and GPU devices used for training and testing CQNPs on image completion tasks.} \label{tab: supp-CQNP-hyperparams-detail-image-comp}
    \scalebox{0.9}{
    \begin{tabular}{c|c|c|c|c|c|c}
      \toprule 
      Benchmark   & Learning rate   & L2 regularizer   & $N_{\tau}$(Training)   & $N_{\tau}$(Testing)  & GPU (Training)   & GPU (Testing)\\
      \midrule
      MNIST         & $5\times10^{-4}$  & $0$  & 25  & 50   & NVIDIA A100  & NVIDIA A100\\
      Fashion MNIST & $10^{-3}$  & $10^{-5}$  & 25  & 50   & NVIDIA A100  & NVIDIA A100\\
      Omniglot      & $10^{-3}$  & $10^{-5}$  & 25  & 50   & Quadro RTX 6000  & NVIDIA A100\\
      FreyFace      & $10^{-3}$  & $0$  & 25  & 50   & Quadro RTX 6000  & NVIDIA A100\\
      SVHN          & $10^{-3}$  & $10^{-5}$  & 25  & 50   & Quadro RTX 6000  & NVIDIA A100
      \\
      \bottomrule
    \end{tabular}}
\end{table*}

%% file: tables/supplement/ACQNP/image-completion/ACQNP-architecture-image-completion.tex
\begin{table*}[!h]
    \centering
    \caption{Architectural details of ACQNPs for image completion tasks.} \label{tab: supp-ACQNP-imp-detail-image-comp}
    \scalebox{0.9}{
    \begin{tabular}{c|c|c|c}
      \toprule 
      Benchmark   & Context Encoder   & Adaptor    & Decoder\\
      \midrule
        \begin{tabular}{c} MNIST\\ Fashion MNIST \\ Omniglot \\ FreyFace  \end{tabular}   & $[3]\times[128, 128, 128]\times[128]$   & $[129]\times[128, 128, 128, 128, 128]\times[1]$   & $[130]\times[128, 128, 128]\times[3]$ \\
      \midrule
        SVHN   & $[5]\times[128, 128, 128]\times[128]$   & $[129]\times[128, 128, 128, 128, 128]\times[1]$  & $[130]\times[128, 128, 128]\times[9]$ \\
      \bottomrule
    \end{tabular}}
\end{table*}

%% file: tables/supplement/ACQNP/image-completion/ACQNP-hyperparam-image-completion.tex
\begin{table*}[!h]
    \centering
    \caption{Hyper-parameters and GPU devices used for training and testing ACQNPs on image completion tasks.} \label{tab: supp-ACQNP-hyperparams-detail-image-comp}
    \scalebox{0.9}{
    \begin{tabular}{c|c|c|c|c|c|c}
      \toprule 
      Benchmark   & Learning rate   & L2 regularizer   & $N_{\tau}$(Training)   & $N_{\tau}$(Testing)  & GPU (Training)   & GPU (Testing)\\
      \midrule 
      MNIST         & $10^{-3}$  & $10^{-5}$  & 25  & 50   & NVIDIA A100  & NVIDIA A100\\
      Fashion MNIST & $10^{-3}$  & $10^{-5}$  & 25  & 50   & NVIDIA A100  & NVIDIA A100\\
      Omniglot      & $10^{-3}$  & $10^{-5}$  & 25  & 50   & NVIDIA A100  & NVIDIA A100\\
      FreyFace      & $10^{-3}$  & $10^{-5}$  & 25  & 50   & NVIDIA A100  & Tesla T4\\
      SVHN          & $10^{-3}$  & $10^{-5}$  & 25  & 50   & NVIDIA A100  & NVIDIA A100
      \\
      \bottomrule
    \end{tabular}}
\end{table*}